\documentclass[aoas]{imsart}
\RequirePackage{amsthm,amsmath,amsfonts,amssymb}
\RequirePackage[authoryear]{natbib}
\RequirePackage{imsart}  % Ensure imsart package is loaded

\usepackage{tikz}
\usepackage{lineno}
\usepackage{fmtcount} % for \ordinalnum
\usepackage{array,multirow}
\usepackage{longtable}
\usepackage{afterpage}
\usepackage{lscape}
\usepackage[export]{adjustbox}
\modulolinenumbers[5]
\usepackage{listings}
\usepackage{xcolor}
\usepackage{todonotes}
\usepackage{cancel}
\usepackage[authoryear]{natbib}
\usepackage{mathtools}
\usepackage{appendix}
\usepackage{verbatim}
\usepackage{amsmath,bm}
\usepackage{amssymb}
\usepackage{blkarray}
\usepackage[para,online,flushleft]{threeparttable}
\usepackage{comment}
\usepackage{todonotes}
\usepackage{array}    % for better table alignment
\usepackage{booktabs} % for improved table aesthetics
\usepackage{caption}  % for custom captions and footnotes
\RequirePackage{fontenc,inputenc,subcaption,graphicx,sectsty,titlesec,caption}
\RequirePackage{amsthm,amsmath,amssymb,animate,nccmath}
\RequirePackage{algorithm,paralist}
\RequirePackage[noend]{algpseudocode}
\definecolor{blue}{HTML}{40B4E6}
\definecolor{magenta}{HTML}{BCBCF8}
\usepackage[colorlinks=true,linktoc=all,linkcolor=blue,citecolor=blue]{hyperref}
\usepackage[T1]{fontenc}
\usepackage{ae}
\usepackage{newtxtext,newtxmath}% added  % More complete math font coverage
\usepackage[final]{microtype}
\usepackage{booktabs}
\usepackage{cleveref}
\usepackage{makecell} % Add this in your preamble
\newcolumntype{C}[1]{>{\centering\arraybackslash}m{#1}}

\def\b1{\mbox{$\mathbf{1}$}}

\def\by{\mbox{\boldmath$y$}}

\def\by{\mbox{$\mathbf{y}$}}
\def\bY{\mbox{$\mathbf{Y}$}}

\def\bY{\mbox{$\mathbf{Y}$}}

\newcommand{\Yvec}{\mathbf{Y}}

\newcommand{\yvec}{\mathbf{y}}
\newcommand{\thetavec}{\boldsymbol{\theta}}

\newcommand{\psivec}{\boldsymbol{\psi}}

\DeclareMathOperator*{\argmax}{argmax}
\begin{document}

\begin{frontmatter}
%%%%%%%%%%%%%%%%%%%%%%%%%%%%%%%%%%%%%%%%%%%%%%
%%                                          %%
%% Enter the title of your article here     %%
%%                                          %%
%%%%%%%%%%%%%%%%%%%%%%%%%%%%%%%%%%%%%%%%%%%%%%
%\title{Point process modeling of accidentals in forensic shoeprint data}
\title{Scalable Spatial Point Process Models for Forensic Footwear Analysis}

%\title{Spatial Point Processes for Forensic Footwear Analysis: Scaling to Large Footwear Databases}
%\title{A sample article title with some additional note\thanksref{T1}}
\runtitle{Scalable Models for Forensic Footwear Images}
%\thankstext{T1}{A sample of additional note to the title.}

\begin{aug}
\author{Alokesh Manna\ead[label=e1]{alokesh.manna@uconn.edu}},
\author{Neil Spencer\ead[label=e2]{neil.spencer@uconn.edu}}
\and
\author{Dipak K. Dey\ead[label=e3]{dipak.dey@uconn.edu}}

\address{University of Connecticut, Department of Statistics, 
Storrs, CT 06269, United States. \printead{e1}, \printead{e2}, \printead{e3}}
\end{aug}
%%%%%%%%%%%%%%%%%%%%%%%%%%%%%%%%%%%%%%%%%%%%%%%
%% Only one address is permitted per author. %%
%% Only division, organization and e-mail is %%
%% included in the address.                  %%
%% Additional information can be included in %%
%% the Acknowledgments section if necessary. %%
%% ORCID can be inserted by command:         %%
%% \orcid{0000-0000-0000-0000}               %%
%%%%%%%%%%%%%%%%%%%%%%%%%%%%%%%%%%%%%%%%%%%%%%%
%\author[A]{\fnms{???}~\snm{???}\ead[label=e1]{???@???}},
%\author[B]{\fnms{???}~\snm{???}\ead[label=e2]{???@???}}
%\and
%\author[B]{\fnms{???}~\snm{???}\ead[label=e3]{???@???}}
%%%%%%%%%%%%%%%%%%%%%%%%%%%%%%%%%%%%%%%%%%%%%%
%% Addresses                                %%
%%%%%%%%%%%%%%%%%%%%%%%%%%%%%%%%%%%%%%%%%%%%%%
%\address[A]{???\printead[presep={,\ }]d{e1}}

%\address[B]{???\printead[presep={,\ }]{e2,e3}}
%\end{aug}

\begin{abstract}
Shoe print evidence recovered from crime scenes plays a key role in forensic investigations. By examining shoe prints, investigators can determine details of the footwear worn by suspects. However, establishing that a suspect’s shoes match the make and model of a crime scene print may not be sufficient. Typically, thousands of shoes of the same size, make, and model are manufactured, any of which could be responsible for the print. Accordingly, a popular approach used by investigators is to examine the print for signs of ``accidentals'' i.e., cuts, scrapes, and other features that accumulate on shoe soles after purchase due to wear. While some patterns of accidentals are common on certain types of shoes, others are highly distinctive, potentially distinguishing the suspect’s shoe from all others. Quantifying the rarity of a pattern is thus essential to accurately measuring the strength of forensic evidence. In this study, we address this task by developing a hierarchical Bayesian model. Our improvement over existing methods primarily stems from two advancements. First, we frame our approach in terms of a latent Gaussian model, thus enabling inference to be efficiently scaled to large collections of annotated shoe prints via integrated nested Laplace approximations. Secondly, we incorporate spatially varying coefficients to model the relationship between shoes’ tread patterns and accidental locations. We demonstrate these improvements through superior performance on held-out data, which enhances accuracy and reliability in forensic shoe print analysis.

\end{abstract}


\end{frontmatter}
%%%%%%%%%%%%%%%%%%%%%%%%%%%%%%%%%%%%%%%%%%%%%%
%% Please use \tableofcontents for articles %%
%% with 50 pages and more                   %%
%%%%%%%%%%%%%%%%%%%%%%%%%%%%%%%%%%%%%%%%%%%%%%
%\tableofcontents

%%%%%%%%%%%%%%%%%%%%%%%%%%%%%%%%%%%%%%%%%%%%%%
%%%% Main text entry area:

\section{Introduction}

Footwear impressions are a common form of forensic evidence, useful for both suspect identification and crime scene reconstruction \citep{bodziak2017footwear}. Traditionally, analysis and courtroom testimony regarding footwear evidence has relied almost exclusively on the expertise of trained examiners. However, recent reports on forensic science have emphasized the need for methods that are data-driven and empirically validated \citep{national2009strengthening, holdren2016report}. In response, researchers are developing statistical methods for footwear analysis \citep{srihari2011analysis, srihari2014computational, shor2018inherent, spencer2020bayesian, venkatasubramanian2021comparing, shafique2023creating}, and curating databases to train and evaluate these methods \citep{chochol2012characteristics, yekutieli2012expert, kortylewski2014unsupervised, speir2016quantifying, lin2022simulated, pasquier2023footwear}

One key focus in footwear analysis is the study of accidentals, or randomly acquired characteristics (RACs). These are the cuts, holes, and embedded materials that accumulate on shoe outsoles through use. Because thousands of identical shoes are manufactured, a suspect's shoe typically cannot be definitively linked to a print based solely on a tread design match. Instead, examiners rely on accidentals to distinguish between shoes with matching tread design.

Nevertheless, accidentals are not perfectly distinctive. Certain shoes may be especially prone to accidentals in particular regions, increasing the likelihood that two shoes exhibit patterns of accidentals that coincidentally match. Consequently, it is important to quantify the probability of such coincidences when considering the correspondence of a suspect's shoe with a crime scene print. Following the calls of \citet{national2009strengthening, holdren2016report}, these probability estimates should be grounded in relevant data and empirically validated. 

\textit{How should one go about this?} Ideally, for a specific shoe print, one would analyze the accidental configurations of many shoes of that same make and model, estimating a probability distribution over all possible accidental configurations. However, such data are unavailable outside a few special cases (\cite{champod2000statistical}, \cite{fruchtenicht2002discrimination}, \cite{adair2007mount}, \cite{wilson2012comparison}). Indeed, datasets consisting of worn shoes with annotated accidentals are scarce in general. Existing resources consist largely of convenience samples that span a wide range of outsole designs with little replication. The two most prominent examples in the literature are the JESA database \citep{yekutieli2012expert} and the West Virginia University (WVU) database \citep{speir2016quantifying}. JESA was collected by the Israeli police, comprising 386 worn shoes from forensic casework. The WVU database includes an opportunistic sample of 1,300 shoes from the United States. 

Although not perfectly suited for estimating the shoe-specific distribution of accidentals, models based on these collections have nonetheless shown promising results. Notably, \citet{spencer2020bayesian} developed a Bayesian hierarchical point process regression framework to pool information across shoes in the JESA database. Their model, tailored to capture both spatial trends in accidental locations and the impact of the shape of the shoe sole's contact surface, substantially outperformed more traditional baselines (\cite{stone2006footwear}). It is worth noting that such improvements in model accuracy can translate into large differences in estimated random match probabilities, substantially impacting the strength of the conclusions drawn. 
Related contributions, such as those by \citet{speir2016quantifying, kaplan2022location}, further validate the promise of data-driven methods.

Yet these existing approaches face important limitations. The model of \citet{spencer2020bayesian} was constrained by the relatively small size of the JESA database, and the computational challenges of fitting bespoke Bayesian hierarchical models. Expanding datasets such as the WVU database demand methods that are both expressive and scalable. Simpler aggregate models, such as those of \citet{kaplan2022location} or \citet{speir2016quantifying}, scale to larger datasets, but lack the flexibility to explicitly tailor accidental distributions to specific outsoles. 

Accordingly, the goal of this study is to develop a model for accidental configurations that is both expressive and computationally scalable. Building on the point process problem formulation of \cite{spencer2020bayesian}, we re-frame this task as fitting a latent Gaussian spatial point process model. This approach enables fast approximate Bayesian inference via integrated nested Laplace approximations (INLA, \cite{rue2017bayesian}), allowing for more sophisticated models that scale easily to the larger WVU dataset and any future datasets collected. Our model incorporates spatially varying coefficients for outsole contact-surface features, which not only yield interpretable results but also improve performance over existing methods by capturing nonlinear relationships between features and accidentals.

We have structured the remainder of the paper as follows. In Section \ref{description}, we provide a detailed description of the problem, followed by some practical considerations in Section~\ref{practical}, and a literature survey in Section \ref{literature_survey}. The available data from WVU is described in Section \ref{data}. In Section \ref{model_description}, we present the model formulation as a latent Gaussian Cox process. The computational procedure for parameter inference using integrated nested Laplace approximation and the choice of the priors are elaborated in Section \ref{inference_procedure}. In Section \ref{model_fit}, we discuss the fit and results of the model, providing insight on where accidentals are likely to occur, given a shoe's specific tread pattern. To assess predictive performance and compare different methods, we introduce the evaluation metrics in Section \ref{model_evaluation}, and demonstrate the superior performance of the proposed model over competing methods.
Finally, in Section \ref{conclusion}, we summarize the key findings, highlighting the novelty and potential contributions of this work. 
Additional detailed calculations and descriptions are provided in the supplementary material.

\section{Background}
\label{Background}

\subsection{Description of Problem}
\label{description}

As detailed in \cite{spencer2020bayesian}, the probability that a shoe coincidentally matches the accidentals of a print can be quantified using accidental-based random match probabilities. Generally, random match probabilities represent the chance that a randomly selected shoe from a relevant population (e.g., all shoes of a given type within a city) would produce a print consistent with the considered features (e.g. accidentals).

In forensic investigations, accidentals are typically compared only after establishing a match in both class characteristics (the size, make, and model of the shoe), and any distinctive general wear patterns (e.g. deterioration of an entire region of the outsole's tread). Accordingly, accidental-based random match probabilities are defined as conditional on established agreement in both class characteristics and general wear.

To formalize this concept, consider the framework presented in \cite{spencer2020bayesian}, adapted from \citet{skerrett2011bayesian}. Let $y$ denote a shoe print, with $y_M$, $y_W$, and $y_V$, respectively denoting its class characteristics, general wear, and accidentals. Similarly, let $s_M$, $s_W$, $s_V$ denote these features for a shoe $s$. Finally, let $\mathbb{A}$ denote the relevant population of shoes, and let $s \sim \mathbb{A}$ denote a shoe drawn uniformly at random from the set $\mathbb{A}$.

Within this notational framework, the accidental-based random match probability $\text{rmp}_V$ of the shoe $s$ and print $y$ is given by:
\begin{align*}
\text{rmp}_V := \mathbb{P}\left(y_V \equiv s_V \mid s \sim  A_y \right), \text{ where }
A_y &:=\left\{s \in \mathbb{A}: y_M \equiv s_M, y_W \equiv s_W \right\} .
\end{align*}
Here, $y_i\equiv s_i$ denotes feature $i \in \left\{V, M, W \right\}$ on shoe $s$ being consistent with that of the print $y$. 

It is worth noting that our use of `consistent with' $(\equiv)$ language instead of the stricter `equal to' ($=$) stems from inherent challenges of footwear comparison. For class characteristics, two shoes of slightly different models or sizes (e.g., off by a half size) can produce prints with practically indistinguishable tread patterns. It is thus more practical to consider all shoes with consistent tread patterns (up to some tolerance measure) rather than shoes that match the class characteristics exactly. The same goes for general wear; wear tends to present slightly differently between prints, especially if time has elapsed between comparisons. Finally, for accidentals, their spatial coordinates cannot be measured with arbitrary precision (\cite{shor2018inherent}). Moreover, depending on the quality of the print $y$, not all accidentals present on the source shoe $s$ may be detectable. To account for each of these challenges, the tolerance measures used to define $\equiv$ for each feature should be context-specific, depending on the quality and medium of the crime scene print, as well as the precision of the measurement tools.

Bringing this together, $\text{rmp}_V$ quantifies the chance that the accidentals $s_V$ on a randomly chosen shoe are consistent (up to some tolerance measure) with those of print $y$, given that its class characteristics $s_M$ and general wear $s_W$ are consistent with $y$. In this work, we avoid a specific choice of tolerance measures, instead directly modeling $s_V \mid s_M, s_W$---the distribution of a shoe's spatial configuration of accidentals conditional on its class characteristics and wear. For any chosen tolerance measure, this distribution allows $\text{rmp}_V$ to be evaluated.

Because $s_M$ and $s_W$ are not directly available in existing databases, we follow \cite{spencer2020bayesian} and use the ``contact surface'' $\mathcal{C}_s$ of shoe $s$ as a practical proxy, thus re-framing the problem as modeling the distribution $s_V| \mathcal{C}_s$. Here, the contact surface refers to an image of a high-quality sample print obtained using the shoe's outsole (e.g. Figure~\ref{fig:original_coarsened_reflected}). Such images, along with corresponding accidental coordinates, are routinely collected for shoes in forensic footwear investigations, and are available for all shoes in both the JESA and WVU databases. We elaborate on this point in Section~\ref{shoe_print_image}.

\subsection{Practical Considerations}
\label{practical}

Recall that the JESA and WVU databases consist of convenience samples of shoes exhibiting a wide range of outsole designs, with few repeats of the same contact surface. Thus, it is not practical to infer the distribution $p(s_V \mid \mathcal{C}_s)$ based only on shoes with contact surface $\mathcal{C}_s$. 

Instead, it is helpful to combine information across a variety of accidental-contact surface pairs $(s_V, \mathcal{C}_s)$ within a given database, improving statistical performance by capturing commonalities between them. This essentially frames the $s_V \mid \mathcal{C}_s$ distribution estimation problem as a spatial point pattern regression task, where the accidental locations $s_V$ are the random spatial point pattern (the response), and the contact surfaces $\mathcal{C}_s$ play the role of the covariates.

\subsection{Previous Work}
\label{literature_survey}

As a preliminary step toward computing accidental-based random match probabilities, \citet{stone2006footwear} proposed a simple hypothetical probability model for accidentals, assuming a uniform distribution for their locations on the shoe’s outsole. Rather than being data-informed, these simple assumptions were made out of mathematical convenience, intended to illustrate how random match probabilities could, in principle, be quantified objectively. \citet{stone2006footwear} claims the uniform assumption is a conservative choice, but this is not the case. The uniform distribution has the maximum entropy among continuous distributions, so it will generally underestimate random match probabilities whenever the true distribution is non-uniform. This leads to overconfident conclusions.

Subsequent studies have empirically examined the validity of the uniformity assumption. Based on a preliminary analysis of the JESA database, \citet{wiesner2020dataset} suggested that uniformity might be reasonable. However, this conclusion was later rejected by \citet{kaplan2022location} via a formal hypothesis test on the same data. Concurrently, \citet{richetelli2022spatial} analyzed the WVU database and also found strong evidence of non-uniformity. Finally, \citet{spencer2020bayesian} demonstrated that data-driven, non-uniform models consistently outperform the uniform baseline when evaluated on held-out data from the JESA database.

Moving beyond uniform models, most recent work has focused on data-driven approaches that pool information across convenience samples of accidentally annotated contact surfaces, such as those in the JESA or WVU databases. With the exception of \citet{spencer2020bayesian}, these studies follow a broadly similar two-step strategy:
\begin{enumerate}
\item Registration: Align all shoes to a common coordinate system based on a registration criterion.
\item Aggregation: Combine the aligned accidental locations to estimate a shared spatial distribution across shoes.
\end{enumerate}

\citet{yekutieli2012expert}, \citet{wiesner2020dataset}, and \citet{kaplan2022location} each used the JESA database, and the following registration strategy rooted in Euclidean coordinates: (i) center the contact surface at the origin, (ii) rotate it so that its principal axis is vertical, (iii) scale to unit height, and (iv) reflect right shoes so they coincide with left shoes. These studies then differed in how they aggregated accidentals, with \citet{yekutieli2012expert} and \citet{wiesner2020dataset} applying two-dimensional kernel density estimation, whereas \citet{kaplan2022location} aggregated counts within 14 expert-defined subregions of the shoe.

For the WVU database, \citet{speir2016quantifying} and \citet{richetelli2022spatial} registered shoes using polar coordinates, standardizing each shoe's accidental and contact surface locations onto a common reference outsole. This registration approach ensures alignment of the contact surface boundary across shoes, at the cost of distorting interior Euclidean distances. After registration, both studies estimated spatial accidental densities using two-dimensional histograms. In addition, \citet{richetelli2022spatial} proposed a second model that included an aggregated contact surface term, allowing the accidental propensity at a location to depend on the number of standardized WVU database shoes with contact surface at that coordinate.

Despite differences in implementation, all of these studies yield similar outputs: a spatial ``heat map'' of accidental propensity, indicating regions of the outsole that are more or less prone to accidentals. In principle, this heat map can be transformed back into the coordinate system of any given shoe, providing an estimate of $s_V \mid \mathcal{C}_s$. However, a key limitation of these approaches is that, beyond being used in registration, the specific contact surface $\mathcal{C}_s$ of a shoe is not considered. That is, the resultant $s_V \mid \mathcal{C}_s$ does not specifically depend on $\mathcal{C}_s$. Even for \citet{richetelli2022spatial}, the impact of $\mathcal{C}_s$ is the same as the contact surface of all other shoes in the database. Hence, $s_V \mid \mathcal{C}_s$ depends only very weakly on $\mathcal{C}_s$, limiting the capacity to infer shoe-specific accidental distributions.

To address this limitation, \citet{spencer2020bayesian} proposed a semi-parametric hierarchical Bayesian model that directly links accidental locations to local properties of the contact surface. Their framework treats the accidentals on a shoe as a realization of a spatial Cox process, with the spatial log-intensity modeled as a sum of smooth spatial effects and covariate effects derived from the contact surface. To make inference computationally feasible, the contact surface image was coarsened and binarized, allowing the geometric configuration of tread (contact versus non-contact) of nearby pixels to enter the model as a categorical predictor. Posterior inference was conducted using a bespoke Markov chain Monte Carlo strategy, and model performance was evaluated through cross-validation on the JESA database. The results showed that direct incorporation of the contact surface yielded a substantially better fit than registration-and-aggregation approaches, underscoring the importance of modeling $\mathcal{C}_s$ directly to estimate $s_V \mid \mathcal{C}_s$. Nevertheless, the model’s expressiveness was constrained by the small size of the JESA dataset and by the computational cost of the Bayesian inference strategy. The goal of the present manuscript is to build on this work by presenting a richer contact surface model fit to more data using a more computationally efficient inference strategy.

Before proceeding to this new work, it is worth highlighting additional work---both foundational and more recent---focusing on broader statistical issues in forensic footwear analysis. \cite{evett1998bayesian} presented a Bayesian consideration of the forensic practice of individualization of evidence, along with an application to the likelihood ratio approach in forensic footwear. \cite{lund2016likelihood}, \cite{lund2017likelihood}, and \cite{lund2022bayesian} explore similar foundational questions in greater depth, including the objectivity of forensic evidence and interpretations of uncertainty in forensic likelihood ratios. 

Multiple recent works \citep{richetelli2017quantitative, damary2018dependence, richetelli2019empirically, smale2023estimate}
 have considered the related problem of empirically assessing the distinctiveness of an individual accidental based on features such as location, shape, and type. Notably, \cite{smale2024estimate1}, \cite{smale2024estimate2} considered this problem on realistic crime scene prints, considering prints left in blood and dust, respectively.

Another promising line of recent research has applied deep learning methods to the matching of shoeprint images, mainly addressing the problems of checking $y_M \equiv s_M$ and $y_W \equiv s_W$. One notable example is the quantitative workflow proposed by \cite{venkatasubramanian2021comparing} that computes multiple similarity scores based on outsole design, size, wear, and accidentals. Performance of such algorithms depends on accurate alignment of shoeprint images, and is therefore sensitive to registration errors. To side-step this issue, \cite{jang2023finely} proposed Shoe-MS, a deep learning technique that can successfully compare prints without the need for explicit registration, and has been shown to be effective even on degraded images \citep{jang2025enhancing}. A key driver of the success of this technique is the use of a ``Siamese'' neural network to derive suitable features for comparison, a technique first proposed for shoe comparison by \cite{kong2017cross}. Other pioneering data-driven work in this area includes \cite{kortylewski2014unsupervised} and \cite{srihari2014computational}.

\section{Description of Data}
\label{data}
\begin{figure}[h]
    \centering
    \begin{subfigure}[b]{0.2\textwidth}
        \centering
        \includegraphics[width=\textwidth]{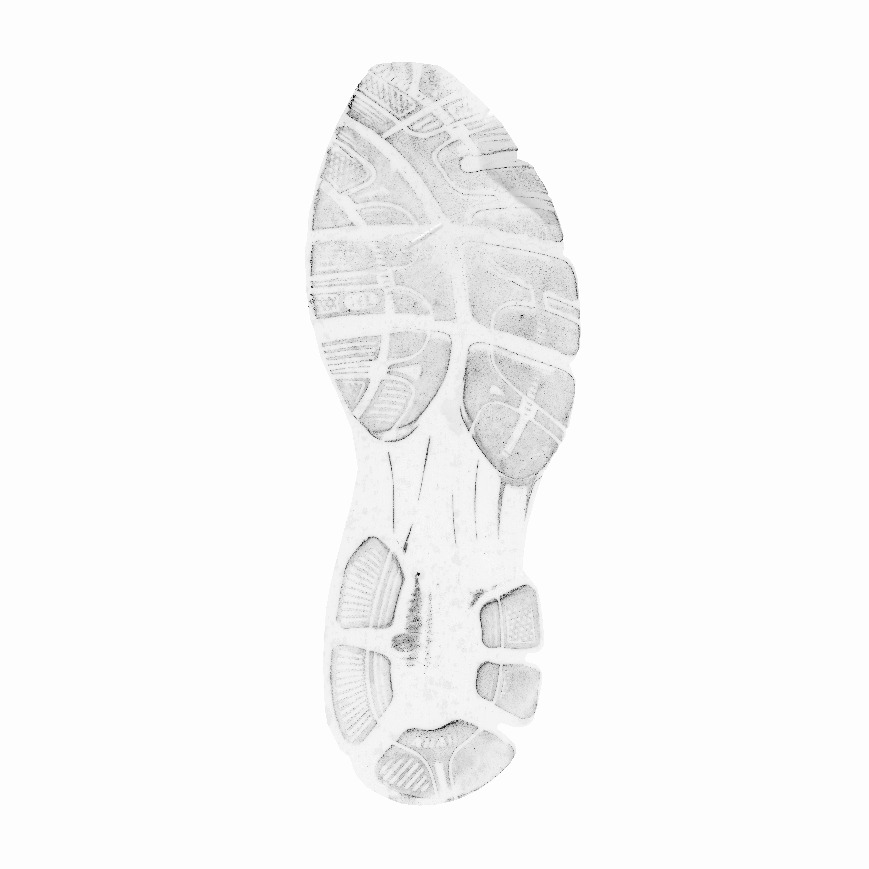}
        \caption{1 L}
        \label{fig:1L}
    \end{subfigure}
    \begin{subfigure}[b]{0.2\textwidth}
        \centering
        \includegraphics[width=\textwidth]{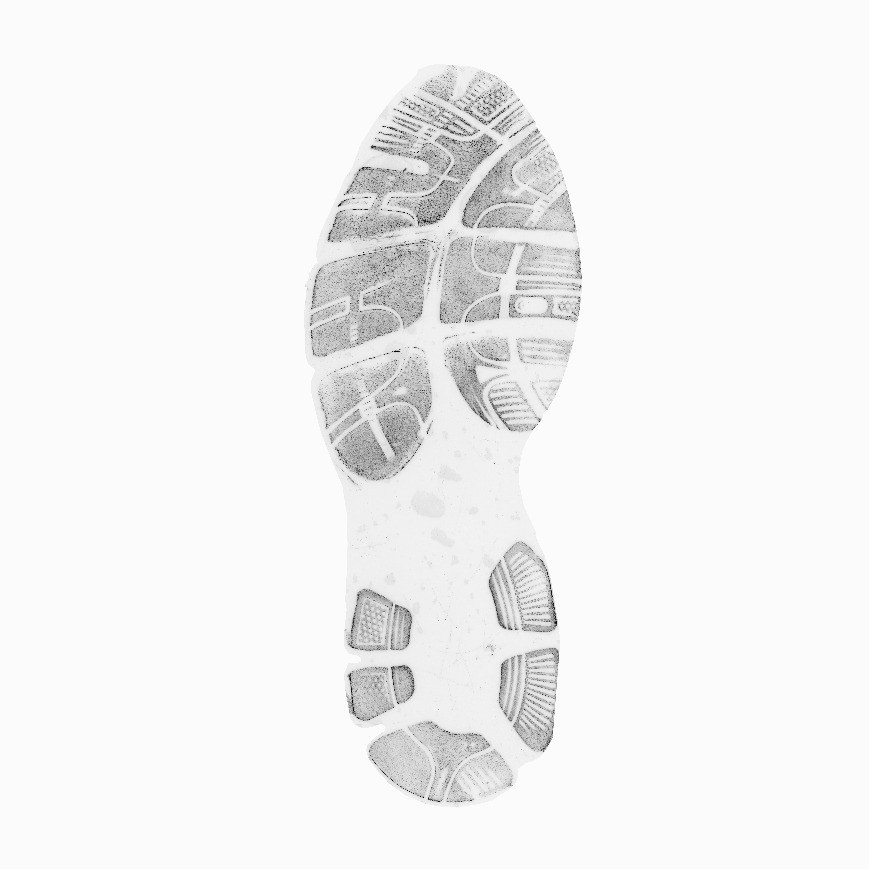}
        \caption{1 R}
        \label{fig:1R}
    \end{subfigure}
    \begin{subfigure}[b]{0.2\textwidth}
        \centering
        \includegraphics[width=\textwidth]{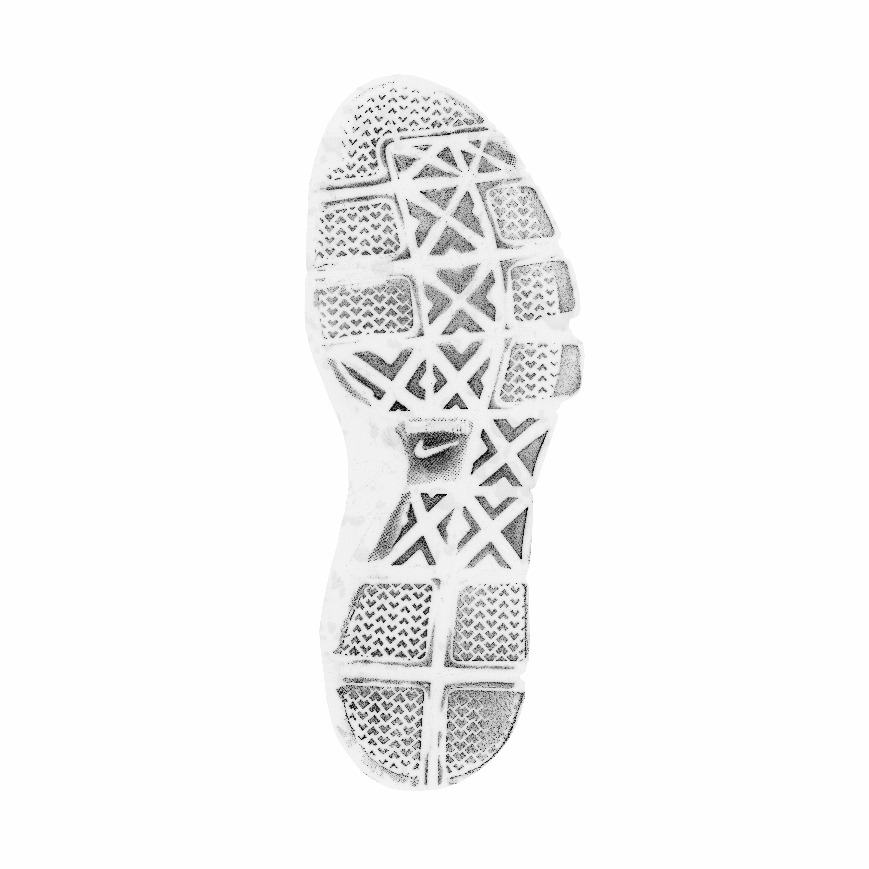}
        \caption{2 L}
        \label{fig:2L}
    \end{subfigure}
    \begin{subfigure}[b]{0.2\textwidth}
        \centering
        \includegraphics[width=\textwidth]{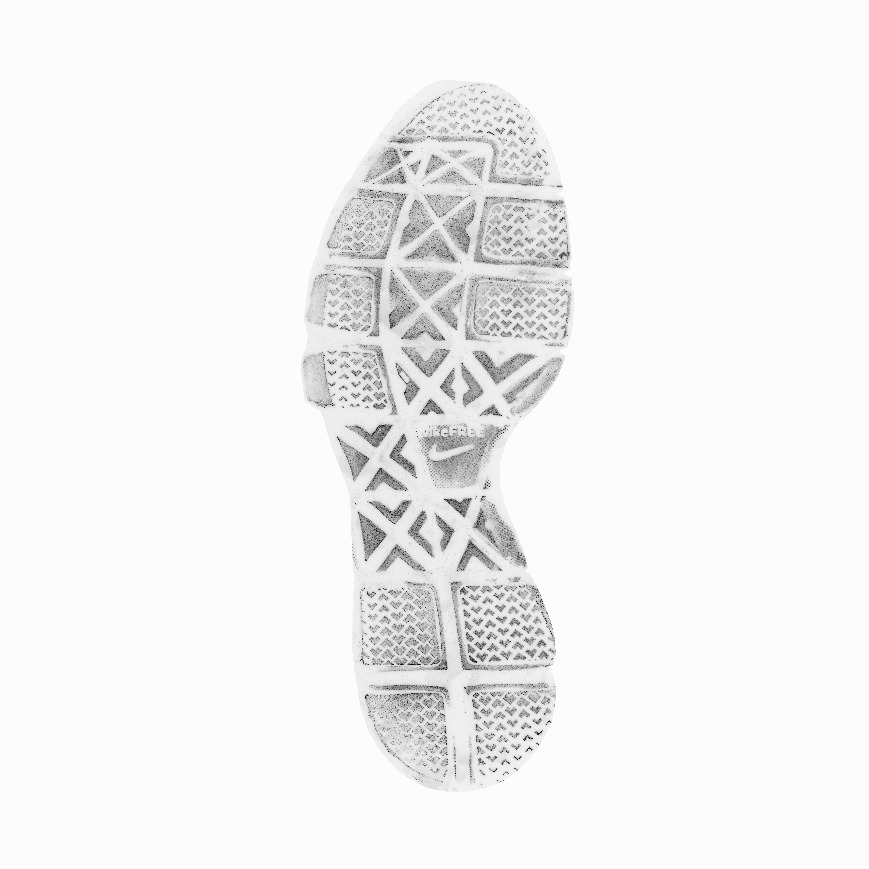}
        \caption{2 R}
        \label{fig:2R}
    \end{subfigure}
    
    \begin{subfigure}[b]{\textwidth}
        \centering
        \includegraphics[width=.7\textwidth, height=0.2\textheight]{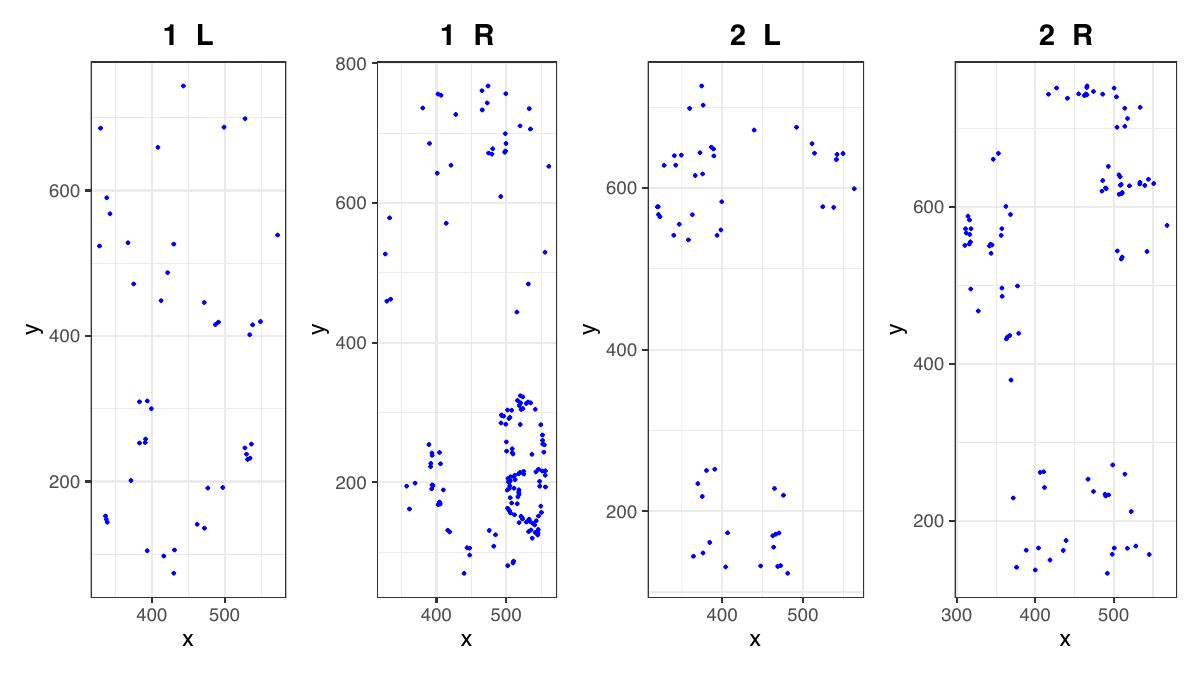}
        \caption{Accidentals locations}
        \label{fig:accidentals}
    \end{subfigure}
    \caption{Four original images of shoeprints and accidental locations}
    \label{fig:original_images}
\end{figure}

Our analysis will focus on a large footwear database of worn shoes as collected by Dr. Jacqueline Speir's research group at West Virginia University. This database pertains to 650 shoe pairs for a total of 1,300 distinct shoes. These shoes vary on a wide range of features, including size, manufacturer, amount of wear, and outsole design.
For each shoe, our investigation will primarily center on two distinct objects: 
\begin{enumerate}
\item The laboratory generated impression (e.g., Figures~\ref{fig:1L}--\ref{fig:2R}): a high-quality sample print obtained by dusting a shoe's outsole with black powder, pressing it onto a clear polyester sheet, then scanning the sheet to obtain a moderate to high resolution grayscale image.
\item The configuration of accidental locations (e.g., Figure~\ref{fig:accidentals}): the coordinates of each accidental occurrence on the shoe sole image as determined by trained researchers.
\end{enumerate}

Descriptions of these objects, as well as accompanying pre-processing steps and mathematical notation, follow in Sections~\ref{shoe_print_image} and \ref{Configuration_of_Accidental_Locations} below. For additional details concerning the data source and collection methodology, see \cite{speir2016quantifying}, \cite{richetelli2019empirically}, and \cite{richetelli2022spatial}.

\subsection{Shoe print Image}
\label{shoe_print_image}

\subsubsection{Raw Image Processing}
Figures~\ref{fig:1L}, \ref{fig:1R}, \ref{fig:2L}, \ref{fig:2R} depict four example shoe print images from the WVU database (\cite{speir2016quantifying}); Figures \ref{fig:1L} and \ref{fig:1R} show the left and right shoes comprising the first pair in the database, and \ref{fig:2L} and \ref{fig:2R} show the left and right shoes of the second pair. 

\begin{figure}[h!]
  \centering
  \begin{subfigure}[t]{.8\textwidth}
  \captionsetup{labelformat=empty}  % Disable (a), (b)
    \centering
    \includegraphics[height=0.42\textwidth]{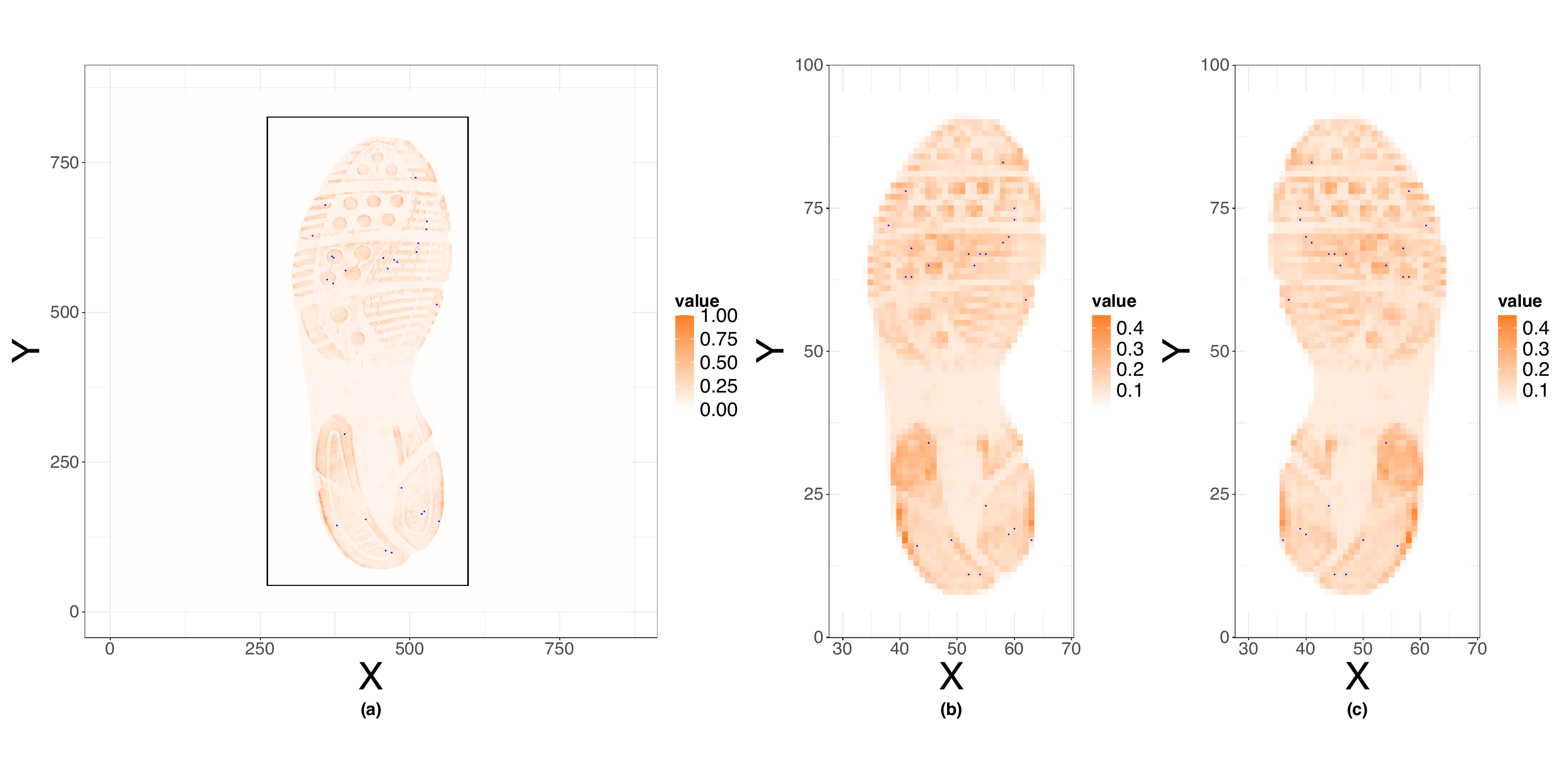}
    \caption{(2a) Shoe 3 R Images: Original, Coarsened, and Reflected.}
    \label{fig:original_coarsened_reflected}
  \end{subfigure}
  \hfill
  \begin{subfigure}[t]{.8\textwidth}
  \captionsetup{labelformat=empty}  % Disable (a), (b)
    \centering
    \includegraphics[height=0.45\textwidth]{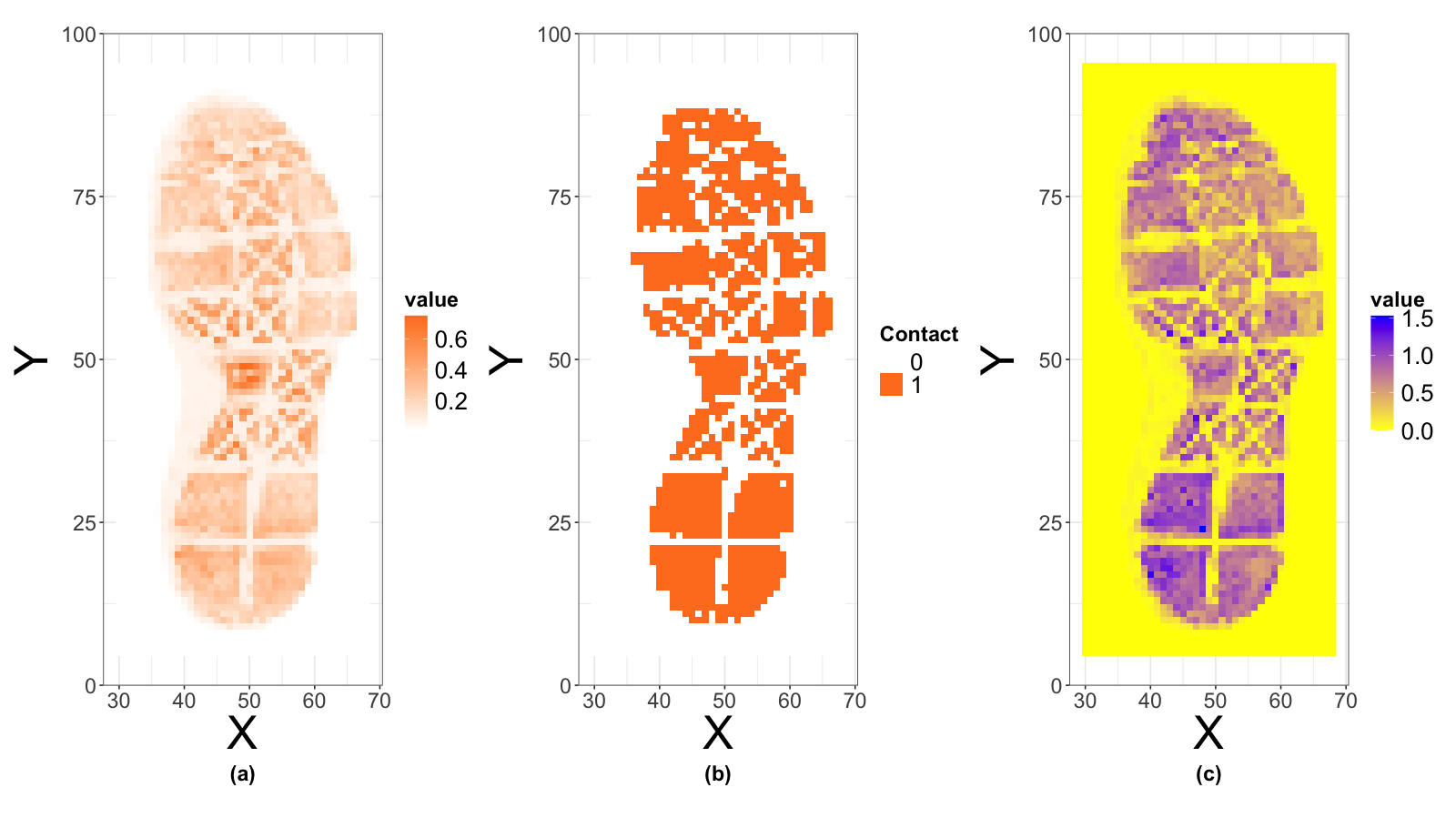}
    \caption{(2b) Shoe 2 Images: Coarsened, Binary, and Image Gradient.}
    \label{fig:coarsened_binary_sobel}
  \end{subfigure}
  \caption{Visual comparisons of contact surface transformations across two example shoes.}
  \label{fig:combined_surface_comparisons}
\end{figure}

We initially obtained these data in a moderate to high resolution format, with each image consisting of $869 \times 869$ grayscale pixels valued over $[0,1]$, with $0$ corresponding to white and 1 corresponding to black. The images were already pre-centered and aligned so that the toe-heel orientation generally aligns with the vertical axis, and the width of the shoe with the horizontal axis. Left shoe prints were oriented to have the big toe on the left-hand side, and right shoes to have it on the right-hand side.  

To render these images suitable for our modeling, additional processing steps were required. First, to allow for the direct pooling of spatial information across both left and right shoes, we followed conventions in the field \citep{yekutieli2012expert, richetelli2022spatial} and reflected right shoe images along their vertical axis to match the layout of left shoes.

Secondly, for computational efficiency, we eliminated the unnecessary white padding on the sides of the images created by their initial square shape $869 \times 869$, considering instead the rectangular region defined by the coordinates in $(262,597)$ for the horizontal axis and $(44,826)$ for the vertical axis. These boundaries retained all shoe print pixels, resulting in $336 \times 783$ images. This resolution is excessively fine-grained for both our modeling goals and computational budget. So, we also coarsened the images, averaging the intensities across 8.7 by 8.7 patches of pixels to reduce the dimension of our final images to a resolution of $39 \times 91$, depicted over the range $[30, 68] \times [5, 95]$. This coarser grid retains sufficient details of the tread patterns of shoes, as can be seen by comparing the original image to the coarsened version in Figure~\ref{fig:original_coarsened_reflected}. 
%\todo{This is not appropriately labeled.}

\subsubsection{Notation}
\label{notation}
The following mathematical notation helps describe how the shoe print images are incorporated into our models in Section \ref{model_description}. Let us first define the set of all 1300 sample shoes available as $\mathcal{S}$, with $|\mathcal{S}| = 1300$. We will use the notation $s \in \mathcal{S}$ when indexing over all shoes in the dataset. We will use two-dimensional Euclidean coordinates to describe the $39 \times 91$ grid defined by the processed shoe print images.

For our simplified notation, we define $\mathcal{A}:=\left\{(x,y):\;x=1,2,\dots,n_{x}, y=1,2,\dots,n_{y}\right\}$ as the collection of all gridpoints where $(x,y)\in \mathcal{A}$ denotes the area $(x-1,x]\times(y-1,y]$. For our processed shoes, $n_x = 39$ and $n_y = 91$. The contact surface of a shoe refers to the region of the sole that typically interacts with the ground during wear, and is therefore responsible for generating footwear impressions.

In our analysis, we will follow \cite{spencer2020bayesian} and use dark pixel regions on the processed shoe print image as indicative of contact surface. We denote this contact surface as $\mathcal{C}_s = (\mathcal{C}_{s,a})_{a \in \mathcal{A}} \in [0,1]^{\mathcal{A}}$ where, for instance, $\mathcal{C}_{5,(19,45)} = 0.4$ tells us that the central pixel located at $(19,45)$th coordinate of processed shoe print image of shoe 5 takes on the value 0.4 (indicating the presence of contact surface).

In addition to the $\mathcal{C}_{s}$ notation used for the shoe print image/contact surface, it will also be helpful to consider two related objects defined below:

\begin{itemize}
\item \textbf{ Binary thresholded contact surface} $\tilde{\mathcal{C}}_{s}$: Instead of considering continuous-valued contact surfaces, \cite{spencer2020bayesian} employed binary versions of the contact surface obtained by thresholding the pixels on the real-valued image. In this work, we use $\tilde{\mathcal{C}}_{s} \in \left\{0,1 \right\}^{|\mathcal{A}|}$ to refer to a binarized version of the contact surface, defined according to
\[
\tilde{\mathcal{C}}_{s,a} =
\begin{cases}
1 & \text{if } \mathcal{C}_{s,a} > c_{s} \\
0 & \text{otherwise},
\end{cases}
\]
where the threshold $c_{s} > 0$ is chosen for each shoe $s$ to accommodate subtle differences in background hue for each image. Figure~\ref{fig:coarsened_binary_sobel} depicts the contact surface $\mathcal{C}_s$ and corresponding binary thresholded version $\tilde{\mathcal{C}}_{s}$ for shoe 2L (originally presented in Figure \ref{fig:original_images}(c)).

\item \textbf{Image gradient} $I_s$: 
The image gradient (sometimes referred to as the Sobel transformation) quantifies the rate at which the pixel intensity is changing in space at a given pixel in an image. Computed using the intensities of neighboring pixels, it tends to be useful for highlighting edges for feature extraction and object recognition. We apply the Sobel method via the \texttt{EBImage} \citep{pau2010ebimage} package to each processed contact surface $\mathcal{C}_s$, yielding an image gradient object $I_s \in \mathbb{R}^{|\mathcal{A}|}$. An example of this object is depicted in Figure~\ref{fig:coarsened_binary_sobel}; mathematical details appear in the supplement.
\end{itemize}

\subsection{Configuration of Accidental Locations}
\label{Configuration_of_Accidental_Locations}

Forensic investigators examine latent shoeprints found at crime scenes by focusing on unique damage patterns that develop through wear. These distinctive features—including scratches, holes, cuts, and perforations—are referred to as accidentals or randomly acquired characteristics \citep{bodziak2017footwear}. Unlike the original manufacturing design, these identifying marks accumulate as the shoe sole encounters various surfaces and debris during normal use. Forensic examiners detect accidentals on footwear by evaluating both the outsole itself and reference impressions, which are high-quality prints produced from the shoe under controlled laboratory conditions.

For each shoe print, the WVU database includes the locations of accidentals on the shoe outsole image, as determined by trained researchers. Figure~\ref{fig:original_images}(e) depicts raw accidental positions as points for the example shoes shown in Figures~\ref{fig:original_images}(a-d). Figure~\ref{fig:original_coarsened_reflected} %\todo{This index is incorrect} 
depicts accidental positions superimposed on an original and subsequent processed versions of a shoe print image.

To support our analysis, we summarized each WVU shoe's accidental locations by determining the total number of accidentals occurring in each coarsened grid coordinate. Note that this summary was performed after transforming the accidental coordinates into the processed space, thus accounting for coarsening and reflecting.
We define the total accidental numbers in $a\in\mathcal{A}$ as a random variable $Y_{s,a}$ and the corresponding realizations of accidentals as $y_{s,a}$ for each shoe, i.e., $ s \in \mathcal{S}$. So, we express the accidental information as a vector of counts $\bY_{s}:=\left(Y_{s,a}\right)_{a\in\mathcal{A}}$ and the corresponding realizations of accidentals are denoted by $\by_{s}:=\left(y_{s,a}\right)_{a\in\mathcal{A}}$. 

\section{Description of Model}
\label{model_description}
\subsection{General Formulation} 
\label{general_formulation}
Our primary goal is to develop and fit a tractable model for the distribution of a shoe's accidental configuration $Y_s$ given its contact surface $\mathcal{C}_s$. Recognizing that each $Y_s$ can be viewed as a spatial point pattern observed over $\mathcal{C}_s$, we approach this problem using tools from spatial statistics. Specifically, it is natural to frame each spatial configuration $Y_s$ as a realization of spatial Poisson point process on $\mathcal{C}_s$. Coarsening the resolution of the process to that of the $\mathcal{C}_s$ grid results in
$$
Y_{s,a}| \mathcal{C}_s \sim \text{Poisson}\left(\lambda_{s,a}\right) \text{  for all  } s \in \mathcal{S} \text{ and }a \in \mathcal{A}. 
$$
Here, $\lambda_{s,a}$ denotes the intensity of the spatial Poisson process for shoe $s$ at spatial coordinate $a$. Equivalently, $\lambda_{s,a}$ models the expected number of accidentals at position $a$ on shoe $s$.

For each shoe $s$, we adopt a latent Gaussian model \citep{rue2017bayesian} for the log spatial intensity function $(\log(\lambda_{s,a}))_{a \in \mathcal{A}}$. Our model includes a spatially smoothed term to encourage correlations between neighboring grid points, and coefficients to incorporate the effect of various functions of $\mathcal{C}_s$ as covariates. Each shoe $s \in \mathcal{S}$ is also assigned its own intercept term as a random effect to capture how overall accidental counts vary broadly across shoes.

Rather than directly state our recommended model for $(\log(\lambda_{s,a}))_{a \in \mathcal{A}}$ in isolation, it is helpful to first establish a more general notation describing a class of models. This notation will make it easier to concisely formulate our model, as well as to compare it to possible variants, including those already proposed in the literature.

Consider the following general model for the log intensity:
%x_{s,a}^{\mathbf{i}} &:= (\mathcal{C}_{s,ij})^{i_{1}}
%    (\mathcal{C}_{s,(i-1)j})^{i_{2}}
%    (\mathcal{C}_{s,(i+1)j})^{i_{3}}
%    (\mathcal{C}_{s,i(j-1)})^{i_{4}}
%    (\mathcal{C}_{s,i(j+1)})^{i_{5}}
%    (I_{s,ij})^{i_{6}} \text{ for  } \mathbf{i} \in \mathcal{I}; \notag\\
\begin{align}
\label{model_spatial_varying_coefficients}
\log(\lambda_{s,a})&:=\beta^{\text{smooth}}_{a}+ \beta^{\text{shoe}}_{s}+\sum_{\mathbf{i}\in\mathcal{I}} x_{s,a}^{\mathbf{i}}\beta_{f}^{\mathbf{i}} + \sum_{\mathbf{i}\in\mathcal{I}^*}  x_{s,a}^{\mathbf{i}}\beta_{sv,a}^{\mathbf{i}} \text{, where}\\
x_{s,a}^{\mathbf{i}} &:= (\mathcal{C}_{s,a})^{i_{1}}
    (\mathcal{C}_{s,a - h})^{i_{2}}
    (\mathcal{C}_{s,a + h})^{i_{3}}
    (\mathcal{C}_{s,a - v})^{i_{4}}
    (\mathcal{C}_{s,a + v})^{i_{5}}
    (I_{s,a})^{i_{6}} \text{ for  } \mathbf{i} \in \mathcal{I} \notag
\end{align}
for some $\mathcal{I} \subseteq \left\{0,1\right\}^{6}$ and
$\mathcal{I}^* \subseteq \mathcal{I}$. Here, 
\begin{itemize}
%\item -$\beta_0$ is the common intercept, 
\item $(\beta_a^{\text{smooth}})_{a \in \mathcal{A}}$ denotes a spatially smoothed trend that is common to all shoes,
\item $(\beta_s^{\text{shoe}})_{s \in \mathcal{S}}$ denotes the shoe-specific random intercepts,
\item $(x_{s,a}^{\mathbf{i}})_{s \in \mathcal{S}, a \in \mathcal{A}, \mathbf{i} \in \mathcal{I}}$ denotes an interaction term involving the contact surface in the neighborhood of $a$ and the image gradient at $a$;  $h = (1,0)$, $v = (0,1)$,
\item $(\beta_{f}^{\mathbf{i}})_{\mathbf{i} \in \mathcal{I}}$ denotes fixed effect regression coefficients for interaction terms,
\item $\beta_{f}^{\mathbf{0}}$ is the overall intercept term when $\mathbf{i}= (0,0,0,0,0,0)$,
\item $(\beta_{sv, a}^{\mathbf{i}})_{a \in \mathcal{A}, \mathbf{i} \in \mathcal{I}^*}$ denotes spatially varying regression coefficients for interaction terms.
\end{itemize}

Note that the choices of $\mathcal{I}$ and $\mathcal{I}^*$ control which contact surface interaction terms are excluded, which are considered as fixed effects, and which are considered as random effects. Note that $\mathcal{I}^* \subset \mathcal{I}$ is typically a proper subset for practical reasons; allowing all effects to vary in space would be both computationally and statistically intractable. Next, we will consider some concrete choices of $\mathcal{I}$ and $\mathcal{I}^*$, representing our proposed model, some variants on it, and models that already exist in the literature. These models are summarized in Table~\ref{tab:5_comparizon_1}.

\subsection{Specific Models}
\label{specific_models}
Let us begin with the model we ultimately recommend, referred to as ``Our Method'' in Table~\ref{tab:5_comparizon_1}. 
 %\backslash \left\{0 \right\}^6
 \textbf{Our method}: our recommended configuration is $\mathcal{I} = \left\{0,1 \right\}^6 $, including fixed effect terms for all 64 interactions (up to order 6) involving the contact surface at a gridpoint, the image gradient at that gridpoint, and the contact surface at neighboring gridpoints. Furthermore, we suggest $\mathcal{I}^* = \left\{(1,0,0,0,0,0), (0,0,0,0,0,1), (1,0,0,0,0,1) \right\}$, amounting to three spatially varying coefficients: one for the contact surface at $a$, one for the image gradient at $a$, and one for the interaction of these two terms. We demonstrate in Section~\ref{model_evaluation} that this set-up exhibits superior predictive performance while maintaining a notion of interpretability. We hold off our discussion of the inference of its parameters until Section~\ref{inference_procedure}.

 \textbf{Existing methods}: Next, it is helpful to consider contrasting these choices with existing contributions from the literature. To facilitate a direct comparison, we will focus on a subset of competitor models that can be framed using the general formulation outlined in Section~\ref{general_formulation}; namely, we consider those approaches  \citep{stone2006footwear, yekutieli2012expert, spencer2020bayesian} that employ a Euclidean strategy for contact surface registration, and can be framed as latent Gaussian models. This ensures that the models can be tractably scaled up to the WVU database using the tools we describe in Section~\ref{inference_procedure}, thus facilitating a direct performance comparison in Section~\ref{model_evaluation}.

For competitors, let us first consider the simple uniform model of \cite{stone2006footwear}. It corresponds to a log intensity function that is uniform in space, equivalent to a prior on $\beta_a^{\text{smooth}}$ that enforces equality for all $a \in \mathcal{A}$. The contact surface is not considered, corresponding to $\mathcal{I} = (0,0,0,0,0,0),\; \mathcal{I}^* = \emptyset$. Similarly, the spatial smoothing model of \cite{yekutieli2012expert} is also independent of the contact surface with $\mathcal{I} = (0,0,0,0,0,0),\; \mathcal{I}^* = \emptyset$. However, $\beta_a^{\text{smooth}}$ can vary smoothly with $a$ in space. 
%\neil{Update this one too}
%\neil{Update this to reflect the new notation}
\cite{spencer2020bayesian} consider multiple variants on their model that exhibit similar performance; our comparison will focus on the ``without scores and kernel'' variant of their final model, as it is the most expressive of their models that can still be framed as a latent Gaussian model and thus tractably fit to the WVU database. All of their models incorporate the binary contact surface $\tilde{\mathcal{C}}_s$ in place of $\mathcal{C}_s$, estimating a model equivalent to using $\mathcal{I} = (\left\{0,1 \right\}^5 \times \left\{0 \right\})$. That is, all interactions of the contact surface are considered, but the image gradient is not. Finally, $\mathcal{I}^* = \emptyset$, as spatially varying coefficients are not considered.

\textbf{Model Variants}: Finally, to justify the specifics of our recommended configuration, it is worth considering additional variants on our model. We summarize four variants in Table~\ref{tab:5_comparizon_1}. Unlike the work of \cite{spencer2020bayesian}, they all use the continuous version of the contact surface. 
%\backslash \left\{0 \right\}^6
\begin{itemize}
\item Variant A is configured according to $\mathcal{I} = (\left\{0,1 \right\}^5 \times \left\{0 \right\})$ and $\mathcal{I}^* = \left\{\mathbf{i} \in \mathcal{I}: \sum_{j=1}^5\mathbf{i}_j \leq 2 \right\}$. 

\item Variants B, C, and D are configured according to $\mathcal{I} = \left\{0,1 \right\}^6$. For Variant B, $\mathcal{I}^* = 
\emptyset$. For Variant C, $\mathcal{I}^* = 
\left\{(1,0,0,0,0,0)\right\}$. For Variant D, $\mathcal{I}^* = 
\left\{(1,0,0,0,0,0), (0,0,0,0,0,1)\right\}$.
\end{itemize}
\begin{table}[ht!]
    \centering
    \resizebox{\textwidth}{!}{%
        \begin{tabular}{|c|c|c|*{2}{c}|*{5}{c}|c|c|}
            \hline
            \multirow{2}{*}{Method} & \multirow{2}{*}{Intercept} & \multirow{2}{*}{Contact format} & \multicolumn{2}{c|}{Contact}
            & \multicolumn{5}{c|}{Contact Interactions Order} 
            & \multirow{2}{*}{\shortstack{\footnotesize Image \\ \footnotesize gradient (IG) }} 
            & \multirow{2}{*}{\shortstack{\footnotesize IG*Loc}}\\ 
            \cline{4-10} 
            &  &  & Loc & Nbd & 2 & 3 & 4 & 5 & 6 & &  \\ 
            \hline
            Our method / M(Final) & SV & Continuous & SV & F & F & F & F & F & F & SV  & SV \\
            \hline
            M(variant A) & SV & Continuous & SV & SV & SV & F & F & F & \(\times\) & \(\times\)  & \(\times\) \\
            \hline
            M(variant B) & SV & Continuous & F & F & F & F & F & F & F & \(\times\)  & \(\times\) \\
            \hline
            M(variant C) & SV & Continuous & SV & F & F & F & F & F & F & \(\times\)  & \(\times\) \\
            \hline
            M(variant D) & SV & Continuous & SV & F & F & F & F & F & F & SV  & \(\times\) \\
            \hline\hline
            M(b); \cite{spencer2020bayesian} & SV & Binary & F & F & F & F & F & F & \(\times\) & \(\times\)  & \(\times\) \\
            \hline
            M(a); \cite{yekutieli2012expert} & SV & \(\times\) & \(\times\) & \(\times\) & \(\times\) & \(\times\) & \(\times\) & \(\times\) & \(\times\) & \(\times\) & \(\times\)\\
            \hline
            Uniform; \cite{stone2006footwear} & F & \(\times\) & \(\times\) & \(\times\) & \(\times\) & \(\times\) & \(\times\) & \(\times\) & \(\times\) & \(\times\) & \(\times\)\\
            \hline
        \end{tabular}
    } % <-- closes resizebox, still inside table
    \caption{Description of different methods with fixed effects and spatially varying effects \\ 
    \footnotesize{Note: Loc: the pixel intensity, SV: spatially varying, F: fixed, \(\times\): not used in the model, *: interaction}}
    \label{tab:5_comparizon_1}
\end{table}

\section{Inference Procedure}
\label{inference_procedure}
Here, we present an inference procedure that is applicable to models formulated as in Section~\ref{general_formulation}, including those in Table~\ref{tab:5_comparizon_1}. For concreteness, our description will focus on the configuration of our proposed model. However, it is straightforward to adapt to other choices of $\mathcal{I}$ and $\mathcal{I}^*$.

\subsection{Bayesian Inference of Parameters} \label{sec:ParameterInference}
Recall the parameters from the model proposed in  Section~\ref{general_formulation}. Here, we reproduce them in vector form: $\beta^{\text{smooth}} \in \mathbb{R}^{|\mathcal{A}|}$, $\beta^{\text{shoe}} \in \mathbb{R}^{|\mathcal{S}|}$, $\beta_f \in \mathbb{R}^{|\mathcal{I}|}$, and $\beta_{\text{sv}}^{\mathbf{i}} \in \mathbb{R}^{|\mathcal{A}|}$ for all $\mathbf{i} \in \mathcal{I}^*$. For convenience, let us define a compact notation concatenating all of the vector parameters into a single vector $\thetavec \in\mathbb{R}^{ |\mathcal{S}| + |\mathcal{I}| + (|\mathcal{I}^*| + 1)|\mathcal{A}|}:$
\begin{align}
\thetavec := \left(\beta^{\text{shoe}}, \; \beta_f,\; \beta^{\text{smooth}}, \;(\beta_{\text{sv}}^{\mathbf{i}})_{\mathbf{i} \in \mathcal{I}^*}  \right). \label{eqn:thetadef}
\end{align}
The likelihood can be written as: 
\begin{align}
%\label{likelihood}
\prod_{s=1}^{|\mathcal{S}|}\mathbb{P}(\bY_{s}=\by_{s}|\mathcal{C}_{s},\thetavec)
 &=\prod_{s=1}^{|\mathcal{S}|}\prod_{a\in\mathcal{A}}  \mathbb{P}(Y_{s,a}=y_{s,a}|\mathcal{C}_{s},\thetavec)
 = \prod_{s=1}^{|\mathcal{S}|}\prod_{a\in\mathcal{A}} \frac{e^{-\lambda_{s,a}} \lambda_{s,a}^{y_{s,a}}}{y_{s,a}!}.
\end{align}
%\neil{Do we have an intercept $\beta_0$? Can we frame that as the mean of the shoe-specific intercepts?}
%\neil{I saw your slack response. Please go through and update the manuscript accordingly. Make sure that all sections (in previous ones) appropriately reflect this change.}
Further, recall that our available data consist of accidental count-contact surface pairs $(Y_s, \mathcal{C}_s) \in \mathbb{N}^{|\mathcal{A}|} \times [0,1]^{|\mathcal{A}|}$ for each $s \in \mathcal{S}$, where $|\mathcal{S}| = 1300$. Our task is to use this available data for inference of the model parameters. For our purposes, we are most interested in estimating the parameters $\beta^{\text{smooth}}$, $\beta_f$, and $\beta_{\text{sv}}$. The shoe-specific intercept terms $\beta^{\text{shoe}}$ are nuisance parameters, included to account for variability in total accidental counts across shoes.

Faced with many random effects and a desire to promote spatial smoothness, it is practical to approach this estimation problem from a Bayesian viewpoint. Specifically, by computing the posterior distribution of $\thetavec \mid (Y_s, \mathcal{C}_s)_{s \in \mathcal{S}}$, we may derive estimates for the components of $\theta$. 
To achieve this, we adopt the following hierarchical prior setup. We employ the following independent priors for each of $\beta_{f},\beta^{\text{shoe}},\beta_{\text{sv}}, \beta^{\text{smooth}}$.
%\neil{We don't really assume the priors are independent, we define them as independent. This should be mentioned when the priors are originally explained instead of right here. And the priors for $\beta^{\text{shoe}}$ are only independent conditional on the hyperparameter $\tau_{\text{s}}$}.

\begin{figure}[h]
    \centering
    \includegraphics[width=.7\textwidth]{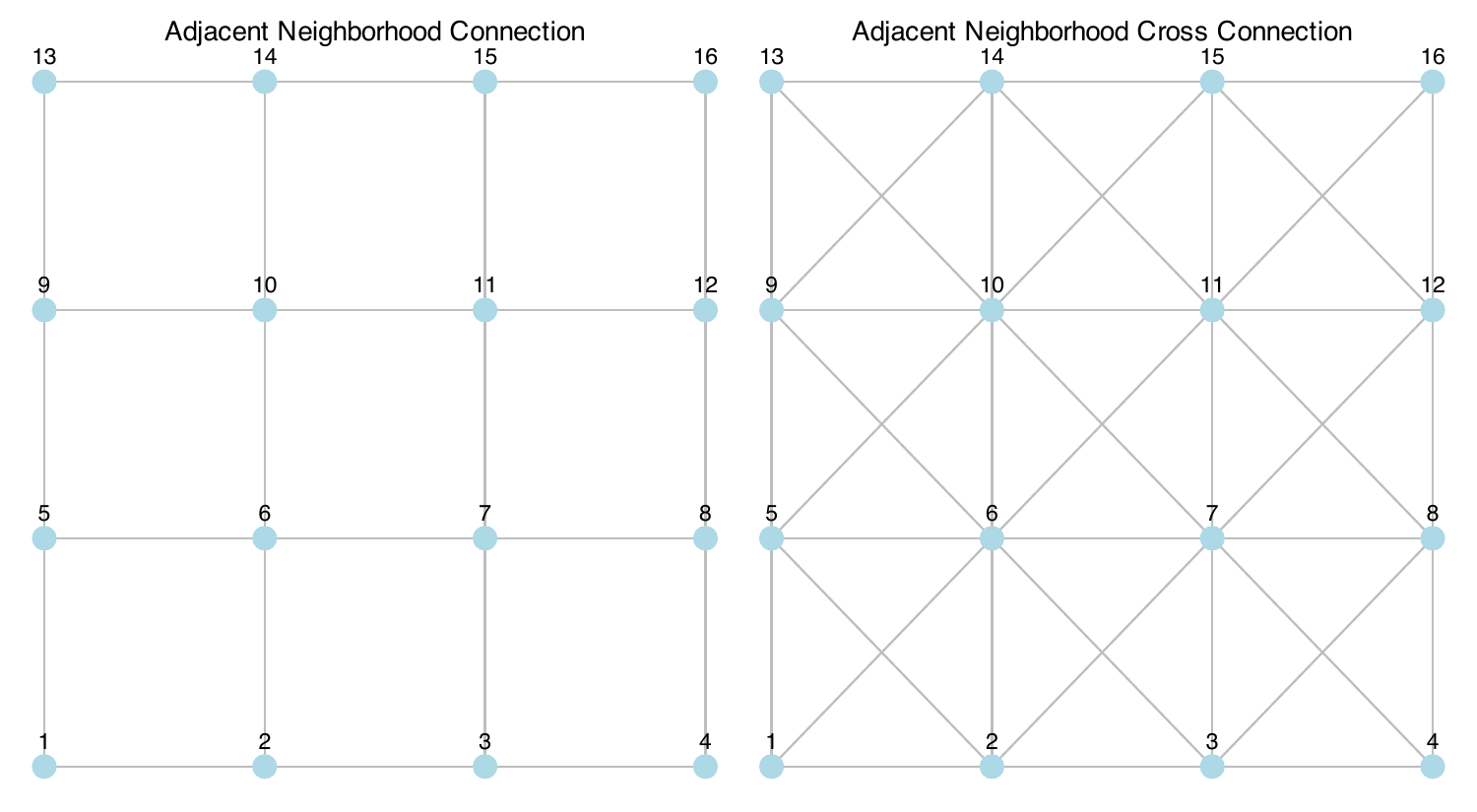} % Change filename if needed
    \caption{Adjacent neighborhood connections if the number of nodes were 16. The left panel neighborhood connection does not involve the diagonal connection. The right panel considers the diagonal connection. We have considered the neighborhood structure as mentioned in the \textit{right panel} for all the nodes in the grid $[30, 68] \times [5, 95]$.
    %\neil{Which model comparison? Comparing what?}. \neil{You mention this is of the ``first few nodes'' this is 16 nodes. Their numbering does not match any indexing we use in the paper }
    } 
    \label{fig:adj_nbd}
\end{figure}

%\neil{Alokesh, could you please concretely write down the priors we are considering for each of the parameters in the model, as well as define any hyperpriors? I know you did something like that below, but it would be better to explicitly state the prior we use on each component. For example, I have gotten one started below already}
%\neil{What prior is employed for each entry in $\beta^{\text{smooth}}$? What prior is employed for each entry $\beta_f$? What prior is employed for each entry in $\beta_{\text{sv}}$? You should write each like you do below for $\beta_{\text{shoe}}$}
%\neil{What is the parametrization of this log gamma distribution? For example, what is its pdf? You do not provide enough information.}
For each $s \in \mathcal{S}$, we consider Gaussian priors on the shoe-specific intercepts as:
\begin{align*}
\beta^{\text{shoe}}_s \mid \tau_{\text{s}}  &\mathop{\sim}\limits^{\text{iid}} \text{Normal}\left(0,\tau^{-1}_{s} \right) \text{ where,  } \tau_{\text{s}} \sim \text{Exponential} (5\times 10^{-5})
\end{align*}
with $\text{Exponential}(\lambda)$ denoting an exponential distribution with rate parameter $\lambda$, and $\text{Normal}(\mu ,\sigma^2)$ denoting a normal distribution with mean $\mu$ and variance $\sigma^2$. Similarly, we assign independent Gaussian priors on each of the fixed effects:
\begin{align*}
\beta_f^{\mathbf{i}} &\mathop{\sim}\limits^{\text{iid}} \text{Normal}\left(0, \;1000 \right)  \text{ for each } \mathbf{i} \in \mathcal{I}.
\end{align*}
For our spatially varying parameters, we use constrained multivariate normal priors:
\begin{align*}
    \beta^{\text{smooth}} 
&\sim \text{Normal}\!\bigl(\mathbf{0}, \; \tau_{\text{sm}}^{-1}\boldsymbol{Q}^{-1}\bigr) \text { with support } \left\{ \beta^{\text{smooth}} \in \mathbb{R}^{|\mathcal{A}|}:\mathbf{1}^{\top}\beta^{\text{smooth}} = 0 \right\}, \\ 
    \beta_{\text{sv}}^{\mathbf{i}}
&\sim \text{Normal}\!\bigl(\mathbf{0}, \; \tau_{\textbf{i}}^{-1}\boldsymbol{Q}^{-1}\bigr) \text { with support } \left\{\beta^{\mathbf{i}}_{\text{sv}} \in \mathbb{R}^{|\mathcal{A}|}: \mathbf{1}^{\top}\beta^{\mathbf{i}}_{\text{sv}} = 0  \right\} \text{ for each } \mathbf{i} \in \mathcal{I}^*,
\end{align*}
where $\mathbf{1} \in \mathbb{R}^{|\mathcal{A}|}$ is a vector consisting of ones. The  precision matrix $Q \in \mathbb{R}^{|\mathcal{A}| \times |\mathcal{A}|}$ is given by
$$
\boldsymbol{Q}_{i j}= \begin{cases} n_{k}& \text { if } i=j =k,\\ -1 & \text { if } a_i \sim a_j, \text { i.e. } \max\!\left(
\left| a_{i1} - a_{j1} \right|,
\left| a_{i2} - a_{j2} \right|
\right) = 1,\\ 0 & \text { otherwise, }\end{cases}
$$
where $a_i = (a_{i1}, a_{i2})$ denotes the $(x,y)$ coordinates of the $i$th entry in $\mathcal{A}$, and $n_k = -\sum_{j \neq k}Q_{jk}$. This Besag dependence structure \citep{besag1974spatial} promotes smoothing over the regular lattice of spatial coordinates; each gridpoint $a \in \mathcal{A}$'s neighborhood is defined as those gridpoints $a' \in \mathcal{A}$ that share at least a boundary point (i.e. queen adjacency). This neighborhood graph structure is depicted in the left panel of Figure~\ref{fig:adj_nbd}.

Note that the precision matrix $Q$ corresponds to the graph Laplacian for this adjacency structure, implying that it is doubly-centered and therefore not full-rank. To compensate, we included the sum-to-zero constraints on the vectors of coefficients. This support restriction is a common convention \citep{keefe2018formal} to ensure the prior is proper.

The $\tau_{\text{sm}}$ and $(\tau_{\mathbf{i}})_{\mathbf{i} \in \mathcal{I}}$ hyperparameters control the level of prior spatial smoothing over $\beta^{\text{smooth}}$ and $\beta_{\text{sv}}$, respectively. For these, we employ independent hyperpriors:
\begin{align*}
\tau_{\text{sm}} &\sim \text{Exponential}(5 \times 10^{-4}),\\
\tau_{\mathbf{i}} &\sim \text{Exponential}(5 \times 10^{-4}) \text{ for all } \mathbf{i} \in \mathcal{I}^*.
\end{align*}
Note that if $\mathcal{I}^*$ is large, it may be computationally impractical to independently infer $\tau_{\mathbf{i}}$ for all $\mathbf{i}$. In these cases, we limit hyperparameter inference to those $\tau_{\textbf{i}}$ corresponding to main effects (i.e. $\mathbf{i}^T \mathbf{i} = 1$). 
For higher order interactions, we consider the fixed value of $\tau_{\mathbf{i}} = 100$. However, for notational succinctness, we consider priors on all elements of $\mathcal{I}^*$ in our expressions below.

All together, we have parameters $\thetavec$ whose prior distribution is governed by hyperparameters $\psivec=(\tau_{\text{s}},\tau_{\text{sm}},\tau_{\mathbf{i}}; \mathbf{i} \in \mathcal{I^*})$.  Note that, conditional on the hyperparameters $\psi$, our joint prior on $\thetavec$ is a constrained multivariate Gaussian. Specifically, our full prior takes the form $\pi_{0}(\thetavec, \psivec) = \pi_{0}(\thetavec \mid \psivec) \pi_{0}(\psivec)$ for
\begin{align}
% \pi_{0}(\psi) &\propto e^{-5000(10\tau_{\text{s}} + \tau_{\text{sm}} + \sum_{\textbf{i} \in \mathcal{I}^*}\tau_{\mathbf{i}})}, \label{hyperprior}\\
% \pi_{0}(\psi) &:= \left(\tau_{\text{s}} \tau_{\text{sm}} \prod_{\textbf{i} \in \mathcal{I}^*} \tau_{\mathbf{i}}\right)e^{-5000(10\tau_{\text{s}} + \tau_{\text{sm}} + \sum_{\textbf{i} \in \mathcal{I}^*}\tau_{\mathbf{i}})}, \label{hyperprior}\\
\pi_{0}(\psi) &:\propto e^{-5 \times 10^{-3}(0.1\tau_{\text{s}} + \tau_{\text{sm}} + \sum_{\textbf{i} \in \mathcal{I}^*}\tau_{\mathbf{i}})}, \label{hyperprior}\\
\pi_{0}(\thetavec \mid \psivec)&:= \sqrt{\frac{|\Sigma(\psivec)|_*}{(2\pi)^{d}}}\exp\left(- \frac{\thetavec^\top \Sigma(\psivec) \thetavec}{2}\right), \text{ for} \notag\\
d &:= |\mathcal{S}| + |\mathcal{I}| + (|\mathcal{I}^*| + 1)(|\mathcal{A}|-1) \text{, and} \label{eqn:d} \\
\Sigma(\psivec) &:=  \left(\tau_{\text{s}} \mathbb{I}_{|\mathcal{S}|}\right) \oplus \left(1000  \mathbb{I}_{|\mathcal{I}|}\right)  \oplus \left(\tau_{\text{sm}} \boldsymbol{Q} \right) \bigoplus_{\mathbf{i} \in \mathcal{I}^*} \tau_{\mathbf{i}} \boldsymbol{Q},
\label{precmat}
\end{align}
where $\mathbb{I}_n$ denotes the $n \times n$ identity matrix, $\oplus$ denotes block diagonal concatenation, and the support of $\pi_0(\thetavec\mid \psi)$ is restricted to $\thetavec \in \left\{\mathbb{R}^{d+1 + |\mathcal{I}^*|}:  \mathbf{1}^\top \beta^{\text{smooth}} = 0, \;\mathbf{1}^\top \beta_{\text{sv}}^{\mathbf{i}} = 0 \text{ for all $\mathbf{i} \in \mathcal{I}^*$} \right\}$, and $|\Sigma|_*$ denotes the generalized determinant of $\Sigma$---the product of the $d$ nonzero eigenvalues. 

Combining the prior and likelihood, our target posterior thus takes the form:
\begin{align}
\pi\left(\thetavec, \psivec \mid \left\{\Yvec_{s}=\yvec_{s}\right\}_{s=1}^{\mathcal{S}},\left\{\mathcal{C}_{s}\right\}_{s=1}^{\mathcal{S}} \right)& \propto  \prod_{s=1}^{|\mathcal{S}|}\prod_{a\in\mathcal{A}} \frac{e^{-\lambda_{s,a}} \lambda_{s,a}^{y_{s,a}}}{y_{s,a}!} \sqrt{\frac{|\Sigma(\psivec)|_*}{(2\pi)^{d}}} \exp\left(- \frac{\thetavec^\top \Sigma(\psivec) \thetavec}{2}\right) \pi_{0}(\psi), \label{posterior}
\end{align}
where $\pi_{0}(\psi)$ is given as in (\ref{hyperprior}), $d$  is given as in (\ref{eqn:d}), $\Sigma(\psivec)$ is given as in \eqref{precmat}, and the support is restricted according to the sum-to-zero constraints described above. Recognizing that this set-up does not yield a closed form solution for the normalization constant, we instead derive a computationally tractable approximate inference strategy for $\thetavec$.

\subsection{Inference Strategy}

For our suggested model M(Final) in Table~\ref{tab:5_comparizon_1}, the full posterior distribution described by (\ref{posterior}) involves $d + 1 + |\mathcal{I}^*| = 14261$ parameters, as well as four hyperparameters. Our available data consists of counts and covariate values for a relatively large number, $|\mathcal{S}| \times |\mathcal{A}| = 1300 \times 3549 \approx 4.6 \times 10^6$, of individual gridpoints. Consequently, reliable inference of $\thetavec$ via Markov chain Monte Carlo is not computationally feasible within a reasonable time frame given our available resources. Instead, we have formulated our model such that the posterior in (\ref{posterior}) adheres to a specific structure---that of a latent Gaussian model (LGM). LGMs admit a fast computation via a technique known as integrated nested Laplace approximations or INLA \citep{rue2009approximate}.

\subsubsection{Review of Latent Gaussian Models and INLA}

\textbf{LGMs}: Latent Gaussian Models (LGMs) are a broad class of hierarchical statistical models in which the observed data $\mathbf{y}=(y_1,\ldots,y_m)^\top$ are assumed to be conditionally independent given a Gaussian random field $\mathbf{x}$ and hyperparameters $\psivec$. Specifically,
$\mathbf{x}\mid \psivec \sim\text{Normal}(\bf{0},\mathbf{Q}(\psivec_1)^{-1})$, where the precision matrix $\mathbf{Q}(\psivec_1)$ depends on hyperparameters $\psivec_1$, and each $y_{i}$ is independently distributed according to a distribution $p(y_i \mid \eta_i(\mathbf{x}), \psivec_2)$ with additional hyperparameters $\psivec_2$ such that $\psivec = (\psivec_1, \psivec_2)$. Here, each $\eta_i(\cdot)$ denotes an observation-specific function of the latent field $\mathbf{x}$.

Models following this set-up can thus be expressed as:
\begin{align*}
\psivec &\sim p(\psivec), \; \text{where } \psivec:=(\psivec_1, \psivec_{2});\\
\mathbf{x}\mid \psivec_1 &\sim \text{Normal}(\bf{0},\mathbf{Q}(\psivec_1)^{-1});\\
p(\mathbf{y}\mid \mathbf{x}, \psivec_2) &= \prod_{i=1}^m p(y_i\mid \eta_{i}(\mathbf{x}),\psivec_2).
\end{align*}
The prior distribution $p(\psivec)$ is typically specified independently of $\mathbf{x}$. The resulting joint distribution therefore factorizes as 
$p(\mathbf{y},\mathbf{x},\boldsymbol{\psi})
= p(\boldsymbol{\psi})\,p(\mathbf{x}\mid\boldsymbol{\psi}_1)\prod_{i=1}^m p(y_i\mid \eta_{i}(\mathbf{x}),\boldsymbol{\psi}_2)$. 

Typically, the functions $\eta_i(\mathbf{x}) = \mathbf{b}_i^\top \mathbf{x}$ are observation-specific linear combinations of the latent field; observed covariates can then be incorporated in the weight vector $\mathbf{b}_i$. Popular models adhering to this framework include logistic regression models with random effects (\cite{rue2009approximate}), spatio-temporal models (\cite{simpson2016going}), and survival models (\cite{martino2011approximate}). Posteriors that factorize this way are amenable to fast and accurate approximation via INLA.\\
%\cite{meehan2023spatially}, \cite{gomez2019missing}, \cite{belmont2024spatio}) \neil{I meant something like: logistic regression models with random effects (cite), time series models (cite), survival models (cite), and spatiotemporal point processes ()}
\noindent\textbf{INLA}: Bayesian inference for LGMs typically targets entries of $\mathbf{x}$ and $\psivec$, with $\mathbf{x}$ being of primary interest. Recognizing that the joint posterior $p(\psi, \textbf{x}\mid y)$ is rarely available in closed form, practitioners typically approximate $p(\psi, \textbf{x}\mid y)$ using tools like Markov chain Monte Carlo (MCMC). However, reliable MCMC is often computationally infeasible when $\textbf{x}$ is high-dimensional, or the number of data points $m$ is relatively large. 

Instead, INLA (\cite{rue2009approximate}) is a fast alternative for doing inference under LGMs, producing reliable approximations of the individual marginal posteriors $p(\mathbf{x}_j\mid \mathbf{y})$ for each entry $j$ in the Gaussian latent field $\mathbf{x}$.
This is accomplished by leveraging the Gaussianity of the latent field to implement accurate Laplace approximations to $p(\psivec\mid \mathbf{y})$ and to $p(\mathbf{x}_j\mid \mathbf{y},\psivec)$, followed by numerical integration over $\psivec$.

The resultant marginal posterior approximations $\tilde{p}(\mathbf{x}_j\mid \mathbf{y})$ for each $j$ provide an avenue to estimate $\mathbf{x}_j$ via $\hat{\mathbf{x}}_j$, the approximate posterior mean. Combined, the result is $\hat{\mathbf{x}}$, a natural point estimator of $\mathbf{x}$. Posterior credible intervals or marginal standard deviations can be used to quantify univariate uncertainty.

It is worth noting that INLA provides primarily accurate approximations to marginal posterior distributions and related low-dimensional summaries; the full joint posterior of $\mathbf{x}$ is not available in closed form. Nonetheless, univariate and low-dimensional summaries are typically the primary inferential targets in applications. Additional details on the specifics of INLA (\cite{rue2009approximate}, \cite{rue2017bayesian}) are available in the supplement. A fast implementation of INLA is available within the \texttt{RINLA} package in R (\cite{RINLAsoftware}).

%INLA (\cite{rue2009approximate}) performs Bayesian inference for latent Gaussian models. The model is specified through a likelihood for the observed data conditional on a latent Gaussian field and a set of hyperparameters, where the latent field typically includes regression coefficients and random effects, and the hyperparameters control aspects such as variances or precisions. INLA uses nested Laplace approximations to integrate out the latent Gaussian field and to approximate the posterior distribution of the hyperparameters. Conditional on these hyperparameters, INLA then computes accurate approximations to the marginal posterior distributions of all latent variables. As a result, INLA avoids the computational cost of MCMC while still providing reliable Bayesian inference. INLA computes the marginal posterior distribution of the coefficients with point estimates such as the posterior mean, along with measures of uncertainty.  We provided computational details of INLA and the framework of the latent Gaussian Model in the supplementary material. 

%\neil{Make note that INLA can handle constrained posterior.}

\subsubsection{Framing Our Approach as a LGM}

The posterior of our model, described in (\ref{posterior}) with $\lambda$ defined as in (\ref{model_spatial_varying_coefficients}), satisfies the criteria of a LGM, and is thus amenable to fast marginal inference via INLA. Specifically, the latent Gaussian field is given by $\textbf{x} = \thetavec$ as in (\ref{eqn:thetadef}), the hyperparameters $\psivec=(\tau_{\text{s}},\tau_{\text{sm}},\tau_{\mathbf{i}}; \mathbf{i} \in \mathcal{I^*})$ as in Section~\ref{sec:ParameterInference}. The response $\mathbf{y}$ is given by $(Y_{s,a})_{s \in \mathcal{S}, a \in \mathcal{A}}$, with entries given by accidental counts in each contact surface gridpoint on each shoe. Indexing according to $s$ and $a$, we thus have:
$$
\eta_{s,a}(\thetavec)
=
\beta^{\text{smooth}}_{a}
+
\beta^{\text{shoe}}_{s}
+
\sum_{\mathbf{i}\in\mathcal{I}}
x_{s,a}^{\mathbf{i}}\,
\beta_{f}^{\mathbf{i}}
+
\sum_{\mathbf{i}\in\mathcal{I}^*}
x_{s,a}^{\mathbf{i}}\,
\beta_{sv,a}^{\mathbf{i}}.
$$
Note that this expression is linear in $\thetavec$ in that it can be expressed as $\mathbf{b}_i^{\top}\thetavec$ for some $\mathbf{b}_i \in \mathbb{R}^{d + 1 + |\mathcal{I}^*|}$.

Because our model is posed as a relatively standard application of a 2-dimensional spatial LGM, inference is straightforward to carry out using the tools available in the \texttt{RINLA} package. We rely on \texttt{RINLA} for the relevant computations in Section~\ref{model_fit} and Section~\ref{model_evaluation}. Note that the sum-to-zero constraint on the spatially varying effects is standard in \texttt{RINLA}.

In the next section, we inspect posterior mean estimates of the model parameters derived from the INLA marginal posteriors, focusing on the inferred spatial structure and the spatially varying covariate effects.

%\neil{What happened to the following point? Describe how we obtain point estimators of various parameters from this output (i.e. posterior means), and how the marginals from INLA are sufficient for extracting these values}

%We are going to inspect the posterior means of the model parameters in the next section. 

%\neil{You need to work on writing this smoother}

%Without smoothing, we overfitted where we observed accidentals, assigning almost no probability to seeing accidentals in places we hadn’t already seen \neil{What do you mean here? What is a ``place we haven't already seen''? Since you are saying this is something you did, you should demonstrate it.}.  
%\betavec^{\text{spatial}}_{A,o_{2}}:=\left\{\betavec^{\text{spatial}}_{a,o_{2}},a\in\mathcal{A}\right\}
%\neil{The variable $\tau_{\text{sm}}$ was never defined. Neither was $\tau_{\text{s}}$. These should all be defined in the prior section if we are sgoing to use them} \alok{in the prior I updated}
%\neil{You should mention they are independent of each other}.\alok{I updated that.}

\section{Model fit}
\label{model_fit}

Using \texttt{RINLA} to fit our proposed model to the full dataset yields posterior point estimates for our main parameters of interest; namely, the spatially varying coefficients $\beta^{\text{smooth}}$ and $(\beta^{\mathbf{i}}_{\text{sv}})_{\mathbf{i} \in \mathcal{I}^*}$, and the fixed coefficients $(\beta_f^{\mathbf{i}})_{\mathbf{i} \in \mathcal{I}}$. These specific fitted values describe accidental propensity varies across space, local contact surface intensity, and their interactions. In this section, we explore and interpret our fitted values for these parameters. Because they are straightforward to visualize as a two-dimensional grid, we begin with the spatially varying effects.

\subsection{Spatial Effects}

\begin{figure}[h]
    \centering
 \includegraphics[width=.9\linewidth,scale=0.6]{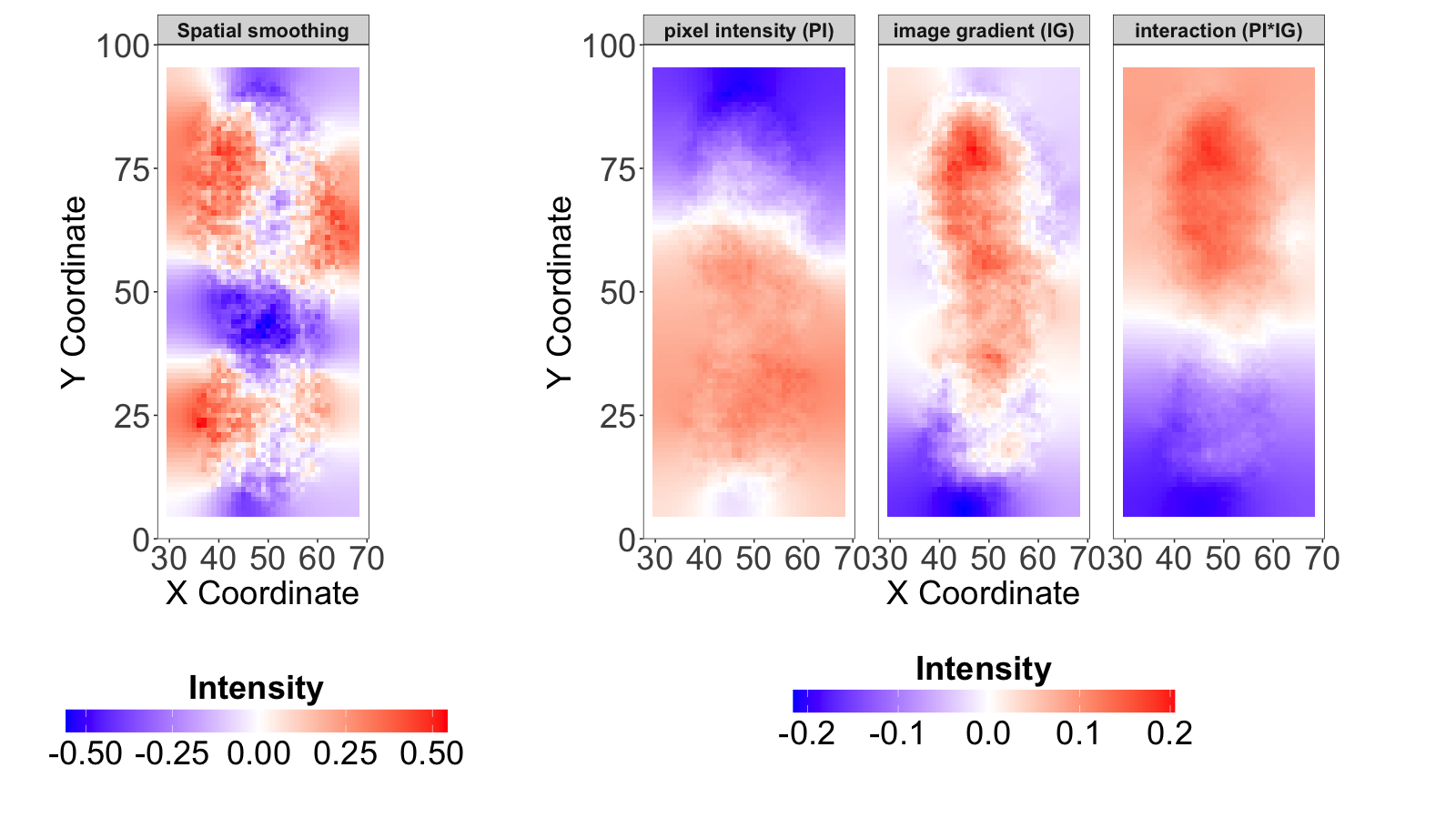}
    \caption{Overall spatial smoothing ($\beta^{\text{smooth}}$) and three raw spatially varying coefficients of pixel intensity, image gradient, and their interaction. The red patches demonstrate high intensity, and the blue patches lower intensity across coefficient values.}
    \label{fig:intercept_and_3_sv}
\end{figure}

\begin{figure}[h]
    \centering
    \includegraphics[width=.9\linewidth,scale=0.6]{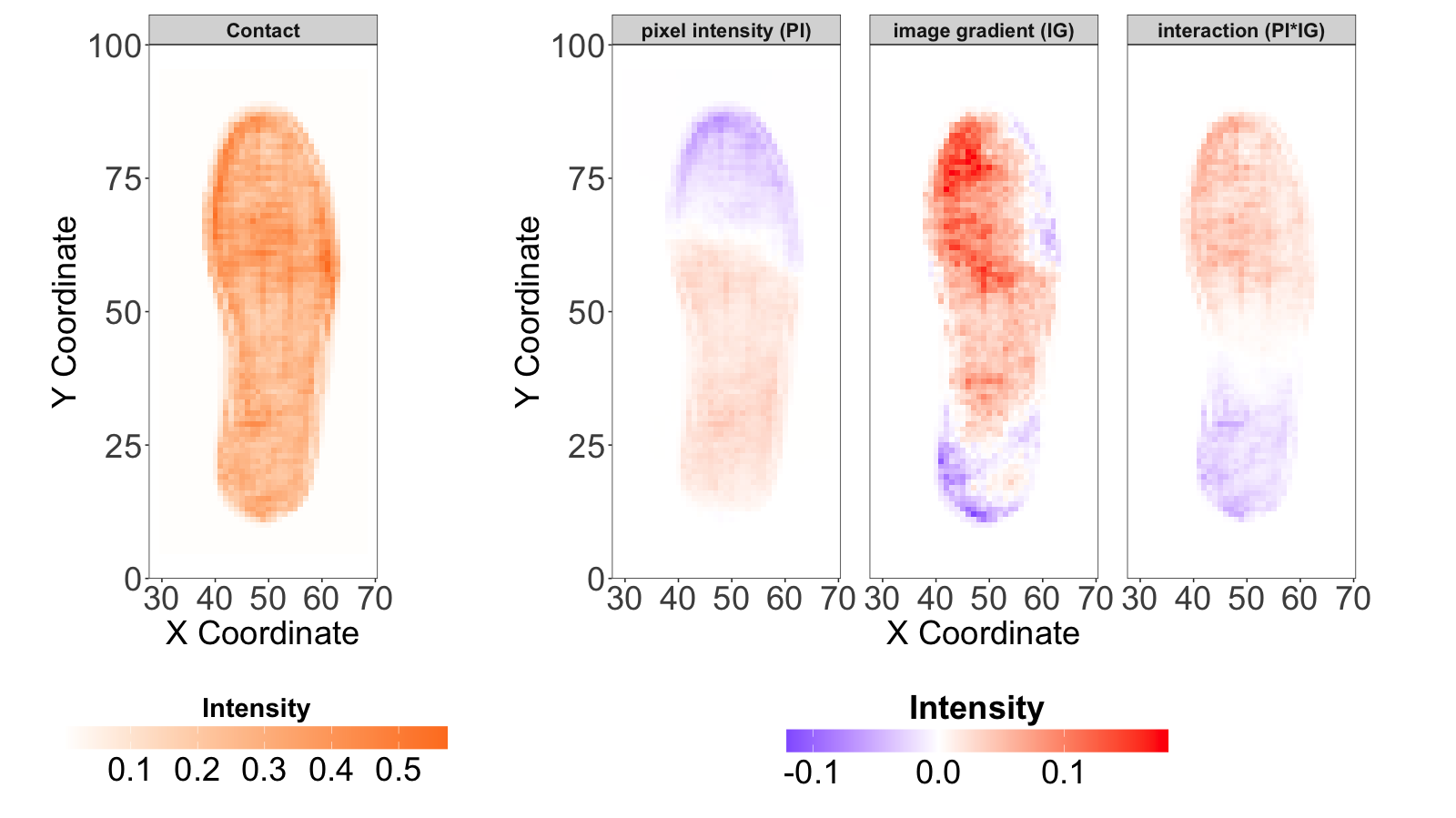}
    \caption{Contact and spatially varying coefficients multiplied with the covariates of an example shoe 202-L. The red patches demonstrate high intensity, and the blue patches lower intensity across coefficient values.}
    \label{fig:contact_and_3_sv_with_x}
\end{figure}

For the four spatially varying effects, we consider both their raw estimated values  (Figure~\ref{fig:intercept_and_3_sv}) and their effect in the context of a specific shoe (Figure~\ref{fig:contact_and_3_sv_with_x}).

The left panel of Figure~\ref{fig:intercept_and_3_sv} shows a heat map of the posterior mean of the spatially varying intercept $\beta^{\text{smooth}}$. Our inference strategy constrains the sum of $\beta^{\text{smooth}}$ to 0, so red regions indicate above typical accidental propensity, whereas blue regions depict lower. We observe that the center portion around $[40,60]\times [30,55]$ has relatively low intensity compared to its surrounding area. This suggests that, even controlling for contact surface, accidentals are less likely to occur at the arch of the shoe. Likewise, the tip of the toe and tip of the heel also exhibit lower accidental propensity. In contrast, the outsole below the sides of the toebox have relatively high propensity for accidentals, as does the inner heel. Note that the range of this effect is approximately 1, meaning the highest propensity regions are prone to roughly $e^{1} \approx 2.7$ as many accidentals as low propensity regions of lowest propensity. 

Next, we consider the right panel of Figure~\ref{fig:intercept_and_3_sv}, depicting three spatially varying coefficients:
\begin{align*}
\beta_{\text{sv}}^{\mathbf{i_1}}\; \text{with} \; \mathbf{i_1}= (1,0,0,0,0,0) &   \; \;\text{(spatially varying effect of pixel intensity),}\; & \\
 \beta_{\text{sv}}^{\mathbf{i_2}} \; \text{with} \;  \mathbf{i_2}= (0,0,0,0,0,1)  & \;\; \text{(spatially varying effect of image gradient),} \text{ and}\\ 
 \beta_{\text{sv}}^{\mathbf{i_3}} \; \text{with} \;  \mathbf{i_3}= (1,0,0,0,0,1) &   \;\; \text{(spatially varying effect of their interaction)}.
 \end{align*}
 Recall from Section~\ref{inference_procedure} that the sum of each of these coefficients is constrained to 0, with red indicating an above typical covariate effect, and blue being below typical. Care must be taken in interpretation: the presence of corresponding fixed effects in the model means that negative values of the spatially varying coefficient does not necessarily imply the covariate has a negative effect. The effect is negative only if the sum of both the fixed and spatially varying effects is negative. 

 Because these three spatially varying coefficients impact accidental propensities through the contact surface covariate, it is challenging to interpret them in isolation. For instance, though shown as strong, the effects shown along the rectangular boundary are unlikely to be impactful in practice because they rarely correspond to nonzero contact surface. Instead, it is helpful to consider the effects in the context of a given shoe as in Figure~\ref{fig:contact_and_3_sv_with_x}.

Figure~\ref{fig:contact_and_3_sv_with_x} depicts the product of the spatially varying coefficients with their corresponding covariates for an example shoe $s'$ (corresponding to 202-L in the WVU database). Namely, Figure~\ref{fig:contact_and_3_sv_with_x} shows how $x_{s',a}^{\mathbf{i}} \beta_{\text{sv}}^{\mathbf{i}}$ varies across $a \in\mathcal{A}$ for $\mathbf{i}= \mathbf{i_1}, \mathbf{i_2}, \text{and } \mathbf{i_3}$. 
 
 With this additional visual aid, we can better see that coefficient for pixel intensity ($\mathbf{i}_1$) varies smoothly across the shoe surface, with larger effects in the lower and central regions and smaller effects toward the upper forefoot. In contrast, the image gradient coefficient shows stronger spatial heterogeneity, with stronger positive effects at the inner toe, and relatively weaker effects at the heel and outer toe. This suggests that edge-related features contribute differently depending on location. The interaction term exhibits a comparatively smoother structure, with stronger effects concentrated in the forefoot and weaker effects toward the heel.

\subsection{Contact Surface Effects}

\begin{figure}[h]
\centering
\includegraphics[width=.8\linewidth,scale=0.8]{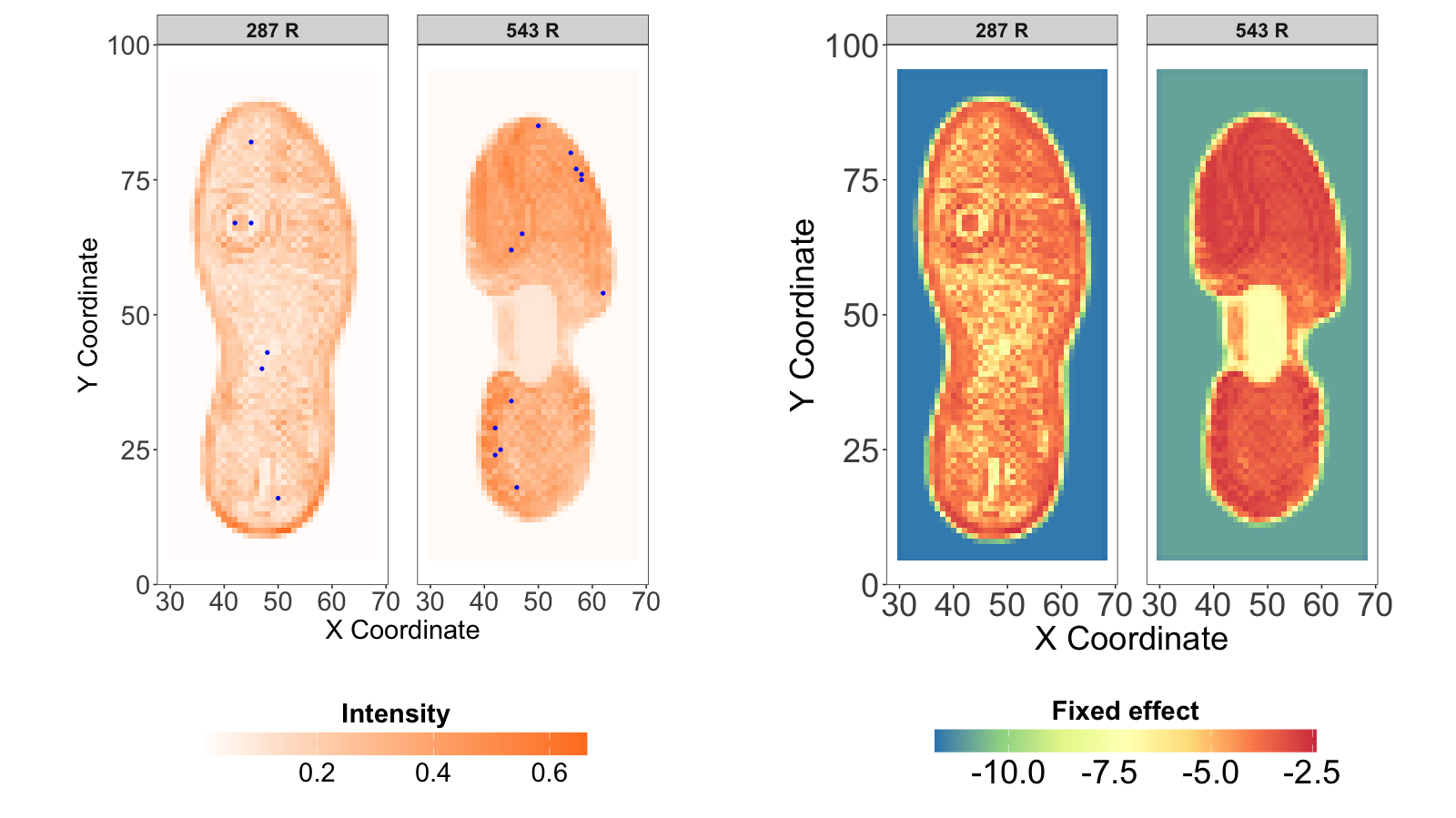}
    \caption{Contact surface and the corresponding cumulative fixed effects $(\sum_{\mathbf{i}\in\mathcal{I}} x_{s,a}^{\mathbf{i}}\beta_{f}^{\mathbf{i}})$  over different spatial locations for shoe 287-R and 543-R.}
    \label{fig:contact_and_fixed_effect}
\end{figure}

Next, we consider our estimates of the fixed effects $(\beta_{f}^{\mathbf{i}})_{\mathbf{i} \in \mathcal{I}}$ modeling the effect of the contact surface on accidental propensities. It is challenging to interpret each effect on its own, as their effects are intertwined through the full set of interactions. Instead, we consider the cumulative fixed effect $(\sum_{\mathbf{i}\in\mathcal{I}} x_{s,a}^{\mathbf{i}}\beta_{f}^{\mathbf{i}})$ below. In Figure~\ref{fig:contact_and_fixed_effect}, we present the contact surface (left panel) and the corresponding cumulative fixed effect (right panel) for two example shoes, 287-R and 543-R. In the right panel, the warmer the color, the higher the contribution to the accidental propensities.

Note that shoe 543-R has a high mid arch, resulting in low contact surface intensity and lower accidental propensity in the arch region. Similarly, there is a circular pattern of lower intensity contact surface on shoe 297-R, translating to lower accidental propensity. In contrast, the inner heel of shoe 543-R is a region of especially high intensity and uniform contact surface, corresponding to especially high accidental propensity. A similar effect is shown at the tip of the heel of 287-R. Finally, for each shoe, the perimeter of low-intensity contact surface neighboring the shoe nonetheless corresponds to moderate accidental propensity, implying that the high-intensity neighboring pixels are driving up the propensity.

\begin{figure}[htbp]
    \centering
    \includegraphics[width=\linewidth]{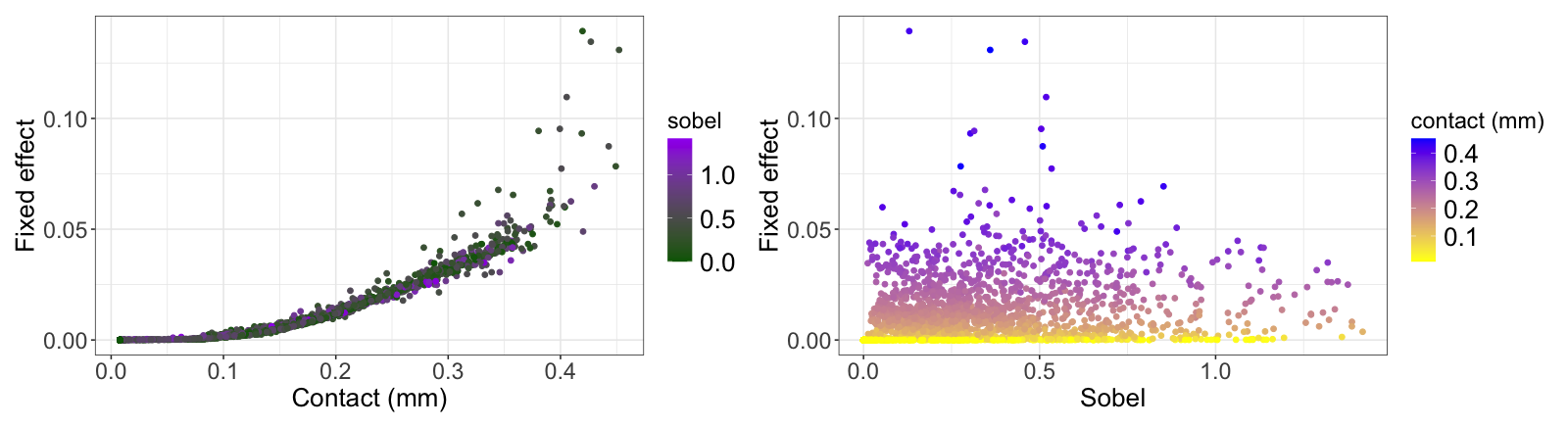}
    \caption{Contact surface and sobel/image gradient vs exponentiated cumulative fixed effect $(\exp(\sum_{\mathbf{i}\in\mathcal{I}} x_{s,a}^{\mathbf{i}}\beta_{f}^{\mathbf{i}}))$ for Shoe 287-R. Contact surface intensity has a major contribution, whereas Sobel has a less significant contribution to the cumulative fixed effect. }
\label{fig:contact_sobel_fixed_effect}
\end{figure}

Overall, the most strikingly apparent observation from comparing the panels in Figure~\ref{fig:contact_and_fixed_effect} is that pixels with more contact surface tend to have higher accidental propensity. To further validate this conclusion, Figure~\ref{fig:contact_sobel_fixed_effect} plots the exponentiated cumulative fixed effect $\exp(\sum_{\mathbf{i}\in\mathcal{I}} x_{s,a}^{\mathbf{i}}\beta_{f}^{\mathbf{i}})$ for Shoe 287-R against two main covariates: the pixel-wise contact surface $x^{\mathbf{i_1}}_{s,a} = \mathcal{C}_{s,a}$ and the image gradient $x^{\mathbf{i_2}}_{s,a} = \mathcal{I}_{s,a}$.

%\neil{Are you sure this is not exponentiated? It must be}
The left panel of Figure~\ref{fig:contact_sobel_fixed_effect} shows a decisive marginal relationship between the two variables, indicating that pixel-wise contact surface value is a dominant driver of the fixed effect component, while the impact of the image gradient is relatively weaker. Note that neighboring pixel intensities tend to be strongly correlated, which can make it difficult to isolate individual effects. Nevertheless, the additional noise around the main trend in the left panel captures the impact of the neighboring contact intensities and the image gradient. 

\begin{figure}
    \centering
    \includegraphics[width=.9\linewidth]{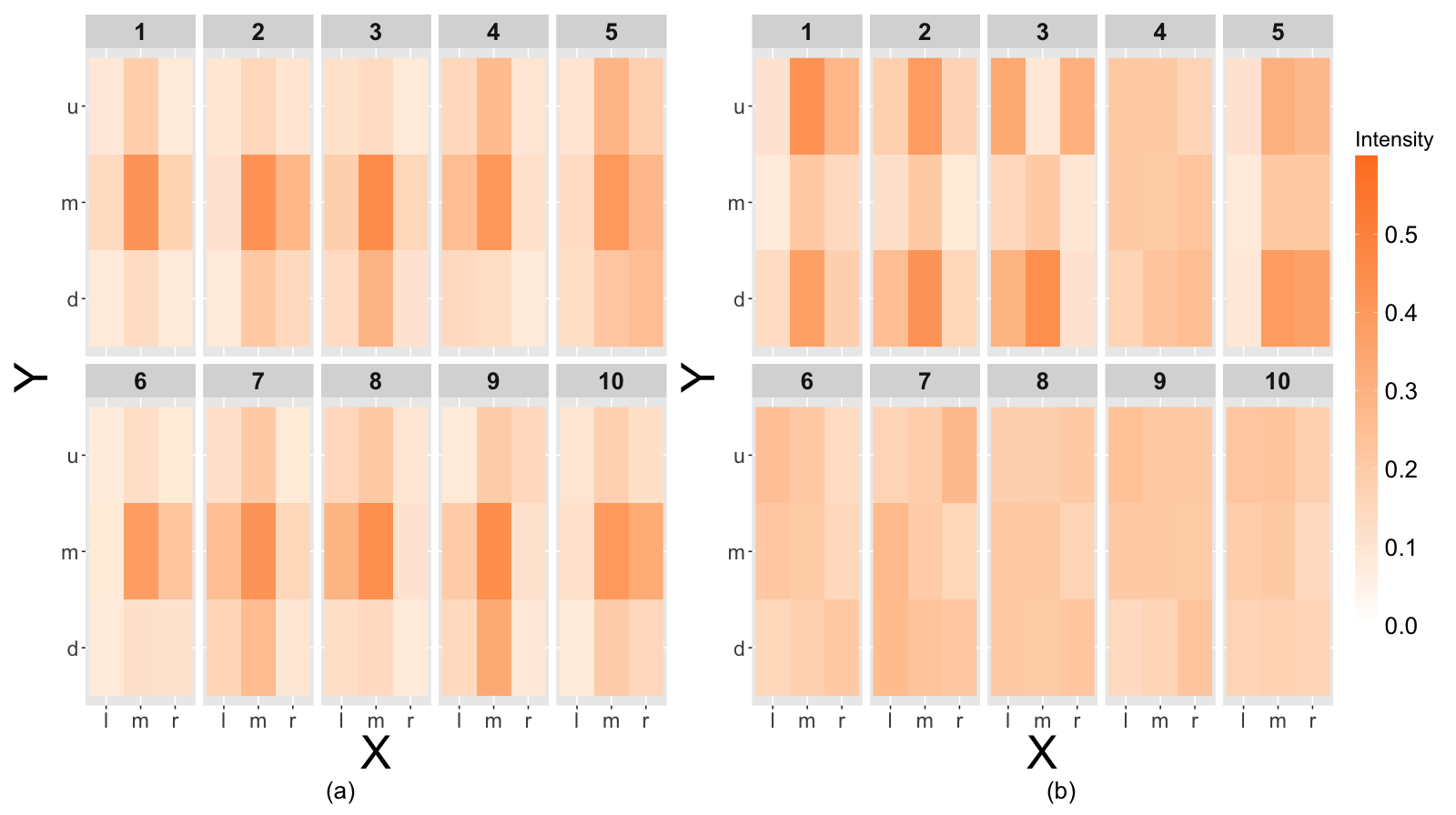}
    \caption{Top 10 (left) combinations and bottom 10 (right) combinations ranked by the fixed effect ($(\sum_{\mathbf{i}\in\mathcal{I}} x_{s,a}^{\mathbf{i}}\beta_{f}^{\mathbf{i}})$ ) of a test shoe 287-R. The lower combinations show the flatter pattern in pixel shoe surfaces. The top combinations have higher pixel intensity in the middle.}
    \label{fig:checkerboard_plot}
\end{figure}

For additional exploration of this relationship, Figure~\ref{fig:checkerboard_plot}
displays the neighboring pixel intensities for the ten highest- and ten lowest-ranking
coordinates \(a\) on shoe 287-R, ranked according to the cumulative fixed effect
\(\sum_{\mathbf{i}\in\mathcal{I}} x_{s,a}^{\mathbf{i}}\beta_{f}^{\mathbf{i}}\), where
only pixels with intensity greater than \(0.2\) are included to exclude regions outside
the shoe print.
The highest-ranking coordinates exhibit pronounced pixel intensity concentrated in the
central region (average intensity .4), indicating stronger and more coherent surface contact.
In contrast, the lowest-ranking coordinates display flatter and more diffuse intensity (average intensity 0.15) patterns, consistent with worn or low-contact areas of the shoe.
Overall, higher pixel intensity is systematically associated
with larger values of \(\sum_{\mathbf{i}\in\mathcal{I}} x_{s,a}^{\mathbf{i}}\beta_{f}^{\mathbf{i}}\),
whereas regions with reduced or uneven contact contribute less to the cumulative fixed
effect. These results provide visual confirmation that increased contact surface is strongly
associated with the magnitude of the cumulative fixed effect.

%For additional exploration of the relationship, consider Figure~\ref{fig:checkerboard_plot}, which displays the neighboring pixel intensities of the 10 highest ranking coordinates $a$ on shoe 287-R in terms of cumulative fixed effects, as well as the 10 lowest ranking coordinates top and bottom ten combinations of pixel intensity, where $\text{pixel intensity}>.2$ and surrounding pixels after ranking all pixels by their fixed effect values. The choice of the cutoff .2 was made to avoid the pixels outside the shoe print. Higher pixel intensity ($\text{contact\_mm}$) tends to correspond to a higher fixed effect, whereas flatter or worn-out regions of the shoe exhibit lower fixed effect values. These conclusions demonstrate that more contact surface is strongly associated with the cumulative fixed effect.

%\neil{Are the white parts excluded?}.
%\neil{We still need to work on the take home message of this plot.}

%\neil{You are referencing things to do with fixed effects before we even presented any of those results.} 
%In Figure~\ref{fig:contact_sobel_interaction_random_effect}, Image gradient and interaction components contribute positively but less strongly, while pixel intensity has a minor negative contribution. Points with high spatial smoothing consistently have higher total random effect, illustrating its dominant role in shaping the spatial variability. When comparing the \textit{range} of the estimated coefficients for a shoeprint 287-R, we observe that the fixed effects span approximately 0 to 0.1 (Figure~\ref{fig:contact_sobel_fixed_effect}), whereas the random effects exhibit a wider range of roughly 0.4 to 1.6 (Figure~\ref{fig:contact_sobel_interaction_random_effect}). The random effects contribute substantially to explaining the variability compared to the cumulative fixed effect. Interestingly, the random effects sometimes exhibit negative slopes, suggesting that the fixed effect, particularly pixel intensity, which can overfit the data, and the random effect serve to adjust and balance this overfitting across spatial locations. This highlights the complementary role of fixed and random effects: while fixed effects capture the general trend, random effects absorb residual spatial heterogeneity, allowing for a more accurate and nuanced representation in the model structure. 

\subsection{Predictive Distribution}

Recall from Section~\ref{description} that, for a given shoe $s$, the distribution of its accidental configuration $\mathbf{Y}_s$ conditional on its contact surface $\mathcal{C}_s$ plays a central role in evaluating accidental-based random match probabilities. Thus far, we have interpreted components of the fitted model by examining the cumulative fixed effects $\sum_{\mathbf{i}\in\mathcal{I}} x_{s,a}^{\mathbf{i}}\beta_{f}^{\mathbf{i}}$ separately from the spatially varying effects. We now combine these components to obtain the predictive distribution of accidental locations, $\mathbf{Y}_s \mid \mathcal{C}_s$. Accurately characterizing this predictive distribution is our primary objective, as it is the object used to compute random match probabilities.

As shown in Section~\ref{general_formulation}, our model defines a spatial Poisson process \citep{Kingman1993Poisson} over the discretized shoe domain $\mathcal{A}$. Consequently, conditional on $\mathcal{C}_s$ and the model parameters $\thetavec$, the accidental locations for shoe $s$ may be viewed as independent draws from $\mathcal{A}$, with probability mass proportional to the intensity $\lambda_{s,a}$. The resulting predictive spatial distribution is therefore characterized by
\[
q(\mathcal{C}_s, \thetavec)
:= \left(q_{a}(\mathcal{C}_s, \thetavec)\right)_{a \in \mathcal{A}},
\]
where the probability $q_{a}(\mathcal{C}_s, \thetavec)$ is defined as:
\begin{align}
\label{eqn:posterior_probability}
q_{a}(\mathcal{C}_s, \thetavec)
&:= \frac{\lambda_{s,a}}{\sum_{a' \in \mathcal{A}} \lambda_{s,a'}} \nonumber\\
&= \frac{\exp\!\left(
\sum_{\mathbf{i}\in\mathcal{I}} x_{s,a}^{\mathbf{i}}\beta_{f}^{\mathbf{i}}
+ \sum_{\mathbf{i}\in\mathcal{I}^*} x_{s,a}^{\mathbf{i}}\beta_{sv,a}^{\mathbf{i}}
+ \beta_{a}^{\text{smooth}}
\right)}
{\sum_{a' \in \mathcal{A}} \exp\!\left(
\sum_{\mathbf{i}\in\mathcal{I}} x_{s,a'}^{\mathbf{i}}\beta_{f}^{\mathbf{i}}
+ \sum_{\mathbf{i}\in\mathcal{I}^*} x_{s,a'}^{\mathbf{i}}\beta_{sv,a'}^{\mathbf{i}}
+ \beta_{a'}^{\text{smooth}}
\right)} .
\end{align}

Importantly, this distribution is independent of the shoe-specific random effect $\beta^{\text{shoe}}_{s}$, since the multiplicative factor $\exp(\beta^{\text{shoe}}_{s})$ cancels in the normalization. Thus, the predictive spatial distribution of an arbitrary accidental on shoe $s$ is fully determined by its contact surface $\mathcal{C}_s$ and the global model parameters $\thetavec$ through $q(\mathcal{C}_s, \thetavec)$.

\begin{figure}[h]
    \centering
    \includegraphics[width=.9\linewidth]{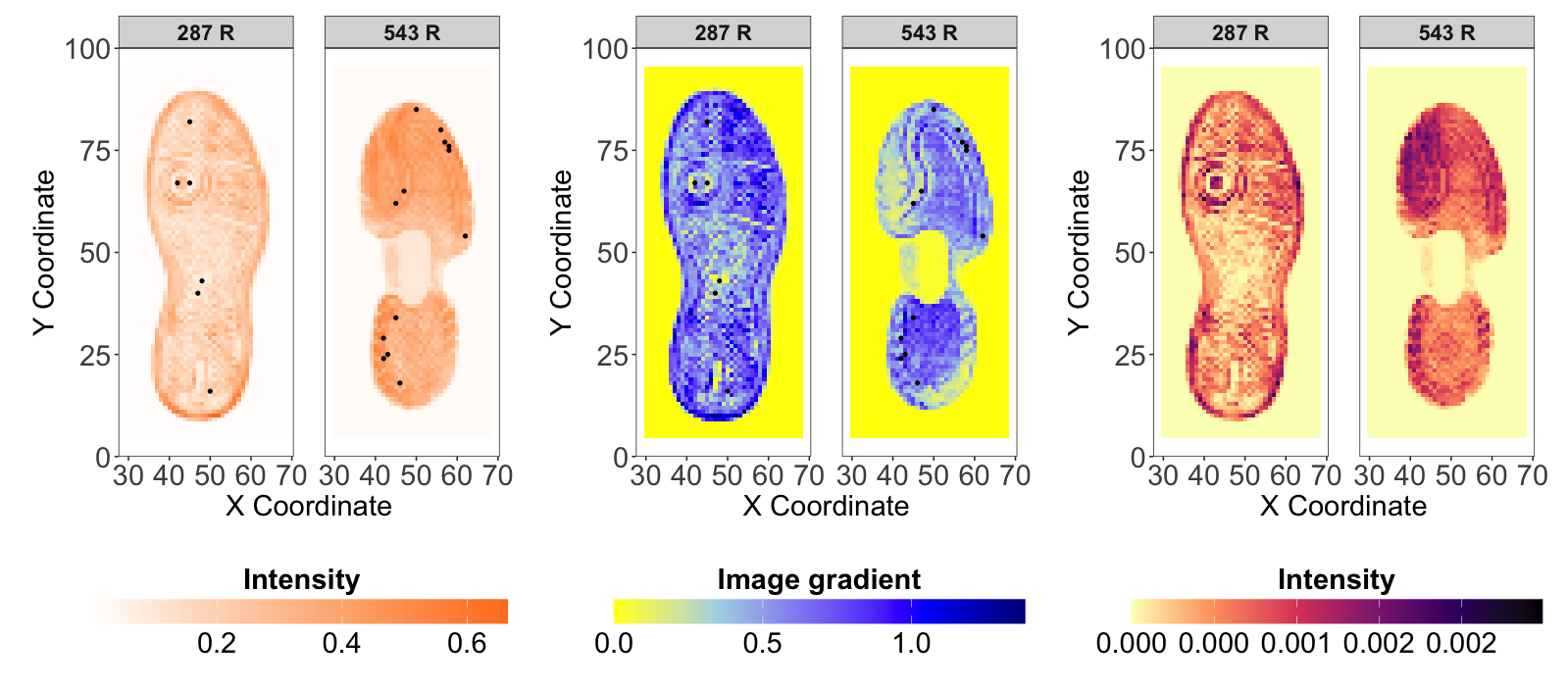}
    \caption{Contact, image gradient, and posterior predictive distribution plots for two different shoeprints 287-R and 543-R. In the third panel, the darker color demonstrates the higher intensity of the posterior probability over the contact surface.}
    \label{fig:contact_sobel_probability_plot}
\end{figure}

Figure~\ref{fig:contact_sobel_probability_plot} illustrates the contact surface $\mathcal{C}_s$, the associated image gradient $I_s$, and the resulting predictive distribution $q(\mathcal{C}_s, \thetavec)$ for two example shoes from the WVU database. The displayed predictive distributions are computed using posterior mean estimates of $\thetavec$ obtained by fitting our final model with the inference strategy described in Section~\ref{inference_procedure}.

%The \textit{posterior probability} $q_{s,a}^{M}$ (Equation~\ref{eqn:posterior_probability}) for each location of two different shoe surfaces (287 R, 543 R), based on $M=\text{M(Final)}$ is given in 
%$\sum_{\mathbf{i}\in\mathcal{I}} x_{s,a}^{\mathbf{i}}\beta_{f}^{\mathbf{i}}+\sum_{\mathbf{i}\in\mathcal{I}^*}  x_{s,a}^{\mathbf{i}}\beta_{\text{sv},a}^{\mathbf{i}}+\beta_{a}^{\text{smooth}}$

%Moreover, partial prints and wearing thread patterns might cause another source of uncertainty. 
%\neil{Please rework this paragraph of interpretation of Figure 9. Right now, I cannot tell exactly what point you are making. Please comment on the inhomogeneity of the predictive intensity across pixels (going from what value to what value, and specify what is driving the primary differences. Also, you've included the observed accidental locations in Figure 9, so you need to make some comment based on them.}

In the Figure~\ref{fig:contact_sobel_probability_plot}, the predictive intensity (third panel) is inhomogeneous across the shoe surface, ranging from $0$ to $.002$, which implies the absolute difference for higher probability regions is approximately $.002$ higher than the low probability regions. These are largely driven by the contact surface ($\mathcal{C}_{s}$) and image gradient, and the parameters estimated from the model M(Final). For example, within the logo (287-R), the pixel intensity and image gradient are higher, resulting in higher contrast with the surrounding area $(X=45, Y=67)$.  We also observe that two exact accidentals appear in the region. In contrast, low image gradient and pixel intensity result in lower predictive intensity. For example, in the middle of the shoe 543-R (around $(X=50, Y=50)$), the contact surface intensity and image gradient are close to 0, resulting in the probability close to 0. In the original image of 543 R, we do not observe any accidentals around that region. Thus, if a shoe print is provided, the above model can capture the probability of the accidental occurrence just based on the contact surface, the derived image gradient intensity, and the coefficients estimated from the fitted model M(Final).

Although visual inspection of individual predictive distributions can provide insight into qualitative aspects of the fitted model, our primary motivation for estimating these distributions is to compute accidental-based random match probabilities. Consequently, the most meaningful assessment of our estimate of $q(\mathcal{C}_s,\thetavec)$ is its predictive performance. Section~\ref{model_evaluation} is devoted to this evaluation. Before turning to that assessment, however, we compare the findings presented here with those of related models in the existing literature.

\subsection{Comparison to Existing Literature}
Some of the insights presented above are consistent with existing findings in the literature, while others appear to be new.

Beginning with $\beta^{\text{smooth}}$, the left panel of Figure~\ref{fig:intercept_and_3_sv} shares several similarities with \citet[Figure~1]{richetelli2022spatial}, which depicts a summary of the aggregated spatial distribution of total accidental counts across the WVU database after polar-coordinate registration. In particular, they also show higher accidental propensity below the inner side of the toebox, at the inner heel, and (to a lesser extent) along the outer side of the toebox. Likewise, both figures show fewer accidentals occurring at the arch of the shoe, as well as near the toe tip and heel tip.

However, the two figures also exhibit key differences. Notably, \citet{richetelli2022spatial} shows a dip in accidental counts near the perimeter of the shoe, as well as substantial spatial variation across bins; some coordinates have more than an order-of-magnitude difference in accidental counts relative to others. In contrast, our Figure~\ref{fig:intercept_and_3_sv} shows no comparable drop along the perimeter, and the overall range of spatial effects is much smaller, with the highest-propensity regions exhibiting roughly $e^{1} \approx 2.7$ times the intensity of the lowest-propensity regions. This reduced variation can be largely attributed to the fact that our model explicitly controls for contact surface. Similarly, the difference along the perimeter can be explained by differences in registration, as the polar-coordinate approach enforces a direct perimeter-to-perimeter correspondence, whereas our Euclidean registration does not.

Another relevant comparison for $\beta^{\text{smooth}}$ is provided by \citet[Supplemental Figure~1]{spencer2020bayesian}, which depicts how the fitted spatial effect varies bin-to-bin under a model that also uses Euclidean registration and fixed effects to control for the contact surface. That model, however, is applied to a different dataset (JESA). The resulting spatial surface shares limited similarities with Figure~\ref{fig:intercept_and_3_sv}; both exhibit increased accidental propensity at the inner heel, and both display a similar overall magnitude of spatial variation (on the order of a factor of two). This level of variation is substantially lower than that observed in models that do not control for contact surface \citep[e.g., Figure~2 of][]{damary2019spatial}, emphasizing the importance of accounting for contact surface when computing random match probabilities.

Turning to the remaining spatially varying effects, there is no existing baseline against which to compare $\beta^{\mathbf{i}_1}_{\text{sv}}$, $\beta^{\mathbf{i}_2}_{\text{sv}}$, and $\beta^{\mathbf{i}_3}_{\text{sv}}$, as our model is the first to incorporate spatially varying effects for the contact surface--related covariates. Nonetheless, our finding that accounting for such effects improves model performance. This result, as well as the specific spatial variation we observe, may be of independent interest to shoe examiners.

Finally, for the fixed effects associated with the contact surface, the closest comparison in the literature is that of \citet{spencer2020bayesian}. While their model employs a binary representation of the contact surface and ours uses a continuous one, the substantive conclusions are largely consistent. In both models, grid points containing, or neighboring, more contact surface tend to exhibit higher accidental propensity, whereas those neighboring little or no contact surface tend to exhibit lower accidental propensity. Moreover, incorporation of the contact surface appears to be a primary driver of improved model performance in both settings.

\section{Model Evaluation and Comparison}
\label{model_evaluation}
% \textcolor{blue}{
% -most interested on multinomial part, -describe metric, -decribe 10 fold strategy, -competitor method from literature and figure, comparison to similar model and figure, table describe different models with average liklihood and 5 percentile lower liklihood (Robust)
% -Drawbacks, Inla not full posterior, continuous contact surface might get overfit, poor performance of predicting the number of accidentals, no replicate data
% }

Recall that the primary use case for our model is the estimation of reliable random match probabilities. Accordingly, our goal is to accurately quantify the relative frequency with which different spatial patterns of accidentals arise on shoe soles, conditional on the contact surface. Specifically, we seek to model $P(Y_s = y_s \mid \mathcal{C}_s)$ in a way that is practically reliable for the accidentals $y_s$ and contact surfaces $\mathcal{C}_s$ encountered in forensic casework. Reliable estimation of this quantity is important for assessing the likelihood of a coincidental match between accidentals observed on a shoe and those present in a crime scene impression. Conversely, unreliable estimates may lead to systematic over or under-estimation of random match probabilities, thereby risking misleading evidentiary conclusions in courtroom settings.

It is thus appropriate to evaluate model performance by assessing how reliably $P(Y_s = y_s \mid \mathcal{C}_s)$ is estimated for shoes in the WVU dataset (see section~\ref{data}). To avoid optimistic bias arising from evaluating the model on the same data used for fitting, we adopt a data-splitting strategy. Specifically, we use 10-fold cross-validation, in which the dataset is randomly partitioned into ten equally sized subsets of shoes; in each fold, nine subsets are used for model fitting and the remaining subset is held out for evaluation.

For each held-out shoe in the validation set, the probability $P(Y_s = y_s \mid \mathcal{C}_s)$ can be decomposed as a product of two components:
\begin{enumerate}
\item a multinomial probability mass function evaluation to quantify the probability of the observed spatial configuration $y_s$, and
\item a Poisson probability mass function evaluation to quantify the probability of observing the total number of accidentals in $y_s$.
\end{enumerate}
Details of this decomposition are provided in Section~1.2 and 1.3 of the Supplement.

Recall that, in typical forensic settings, only a subset of accidentals present on the source shoe are detectable in a crime scene impression. As a result, the total number of accidentals---captured by the Poisson component of $P(Y_s = y_s \mid \mathcal{C}_s)$---is generally not of primary relevance for computing random match probabilities. Instead, the spatial distribution of accidentals, as characterized by the multinomial component, is used to derive our metric for model evaluation.

Specifically, we employ the following per-shoe evaluation metric:
\begin{equation}
\label{eqn:comparison_metric}
m(Y_s = y_s \mid \mathcal{C}_s, \thetavec) = \frac{\sum_{a \in \mathcal{A}} y_{s,a} \log\left(q_{a}(\mathcal{C}_s, \thetavec)\right)}{\sum_{a \in \mathcal{A}} y_{s,a}} - \log(\delta_{\mathcal{A}}),
\end{equation}
where $q_{a}(\mathcal{C}_s, \thetavec)$ is defined as in (\ref{eqn:posterior_probability}), and $\delta_{\mathcal{A}} = \tfrac{783 \times 336}{91 \times 39} \approx 74.13$. 

The quantity in (\ref{eqn:comparison_metric}) corresponds to the natural logarithm of the multinomial component of $P(Y_s = y_s \mid \mathcal{C}_s)$, subject to three modifications. First, the expression is normalized by the total number of accidentals on the shoe, allowing comparisons across shoes with differing accidental counts. Second, the multinomial coefficient is omitted for the same reason. Finally, the constant $\delta_{\mathcal{A}}$ is included to account for the coarseness of the grid $\mathcal{A}$.

In the case of the WVU database, the original images of the laboratory generated shoe impressions were provided at a resolution of $869 \times 869$, corresponding to a $783 \times 336$ region containing the outsole impressions. The adjustment by $\delta_{\mathcal{A}}$ therefore corresponds to converting probabilities defined on the coarsened grid back to the original resolution, implicitly distributing probability mass uniformly across the pixels comprising each grid cell. While we do not explore alternative coarsening schemes in this work, we highlight this adjustment to emphasize a subtle challenge that arises when comparing results across different databases, discretization levels, and registration strategies.

Rather than evaluate our proposed model in isolation, we provide context by comparing our model's data-splitting performance to  related models discussed in Section~\ref{literature_survey}, all expressed within our latent Gaussian model framework as described in Section~\ref{specific_models}.  A summary of these models is available in Table~\ref{tab:5_comparizon_1}. Specifically, we collected the metric in (\ref{eqn:comparison_metric}) for all 1300 WVU database shoes for our model M(Final) along with four competing models from Table~\ref{tab:5_comparizon_1}: the Uniform model as in \cite{stone2006footwear}; M(a), a spatially model without contact surface effects as in \cite{yekutieli2012expert}; M(b), a binary contact surface model as in \cite{spencer2020bayesian}; M(Variant A), a variant of our model with 15 spatially varying coefficients for the contact surface, but no image gradient terms. 

A summary of this comparison is provided in Table~\ref{tab:average_performance_euclidean_5methods}, which reports the average value of the metric in (\ref{eqn:comparison_metric}) across the 130 held-out shoes in each fold. The relative performance of the five models is consistent across folds. Our method achieves the best performance in every fold, followed closely by the M(Variant A) specification of our model. The next best-performing approach is M(b) from \cite{spencer2020bayesian}, after which there is a substantial drop in performance among models that do not incorporate contact surface information. This pattern reinforces the importance of accounting for the contact surface when computing reliable random match probabilities.

\begin{table}[!h]
\centering
\renewcommand{\arraystretch}{} % Adjust row spacing
\resizebox{.7\textwidth}{!}{%
\begin{tabular}{|r|r|r|r|r|r|}
  \hline
  \textbf{Fold} & \makecell{Uniform} & \makecell{Intercept SV\\ M(a) } & \makecell{Binary contact\\ M(b)} & \makecell{15 SV\\ M(variant A)} & \makecell{Our Method/\\ M(Final)} \\
   \hline
1 & -12.480 & -11.920 & -11.451 & -11.344 & \textbf{-11.311} \\ 
  2 & -12.480 & -11.940 & -11.463 & -11.336 & \textbf{-11.314} \\ 
  3 & -12.480 & -11.921 & -11.460 & -11.362 & \textbf{-11.334} \\ 
  4 & -12.480 & -11.887 & -11.475 & -11.327 & \textbf{-11.297} \\ 
  5 & -12.480 & -11.900 & -11.459 & -11.348 & \textbf{-11.326} \\ 
  6 & -12.480 & -11.944 & -11.545 & -11.452 & \textbf{-11.426} \\ 
  7 & -12.480 & -11.882 & -11.521 & -11.388 & \textbf{-11.359} \\ 
  8 & -12.480 & -11.904 & -11.402 & -11.331 & \textbf{-11.287} \\ 
  9 & -12.480 & -11.884 & -11.466 & -11.353 & \textbf{-11.328} \\ 
  10 & -12.480 & -11.913 & -11.500 & -11.395 & \textbf{-11.364} \\ 
   \hline
\end{tabular}
}
\caption{Average model performance as measured by (\ref{eqn:comparison_metric}) across ten folds. Top performer for each fold shown in bold.} %Models formulated as in Table~\ref{tab:5_comparizon_1}
\label{tab:average_performance_euclidean_5methods}
\end{table}

\begin{figure}[h]
    \centering
    \begin{subfigure}[t]{0.48\linewidth}
        \centering
        \includegraphics[width=\linewidth]{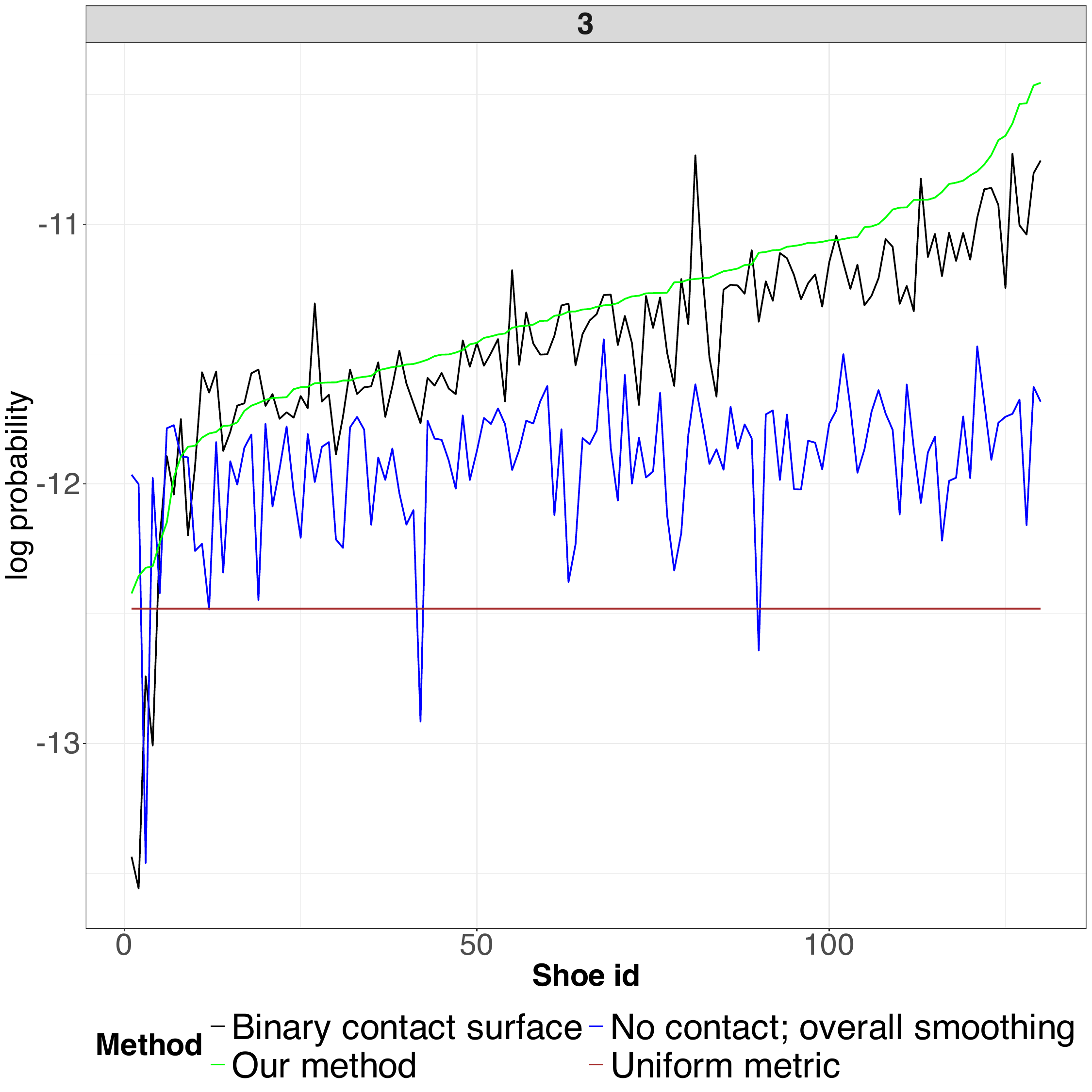}
        \caption{Comparison of methods for Fold 3.}
        \label{fig:multinomial_comparizon_plot}
    \end{subfigure}
    \hfill
    \begin{subfigure}[t]{0.48\linewidth}
        \centering
        \includegraphics[width=\linewidth]{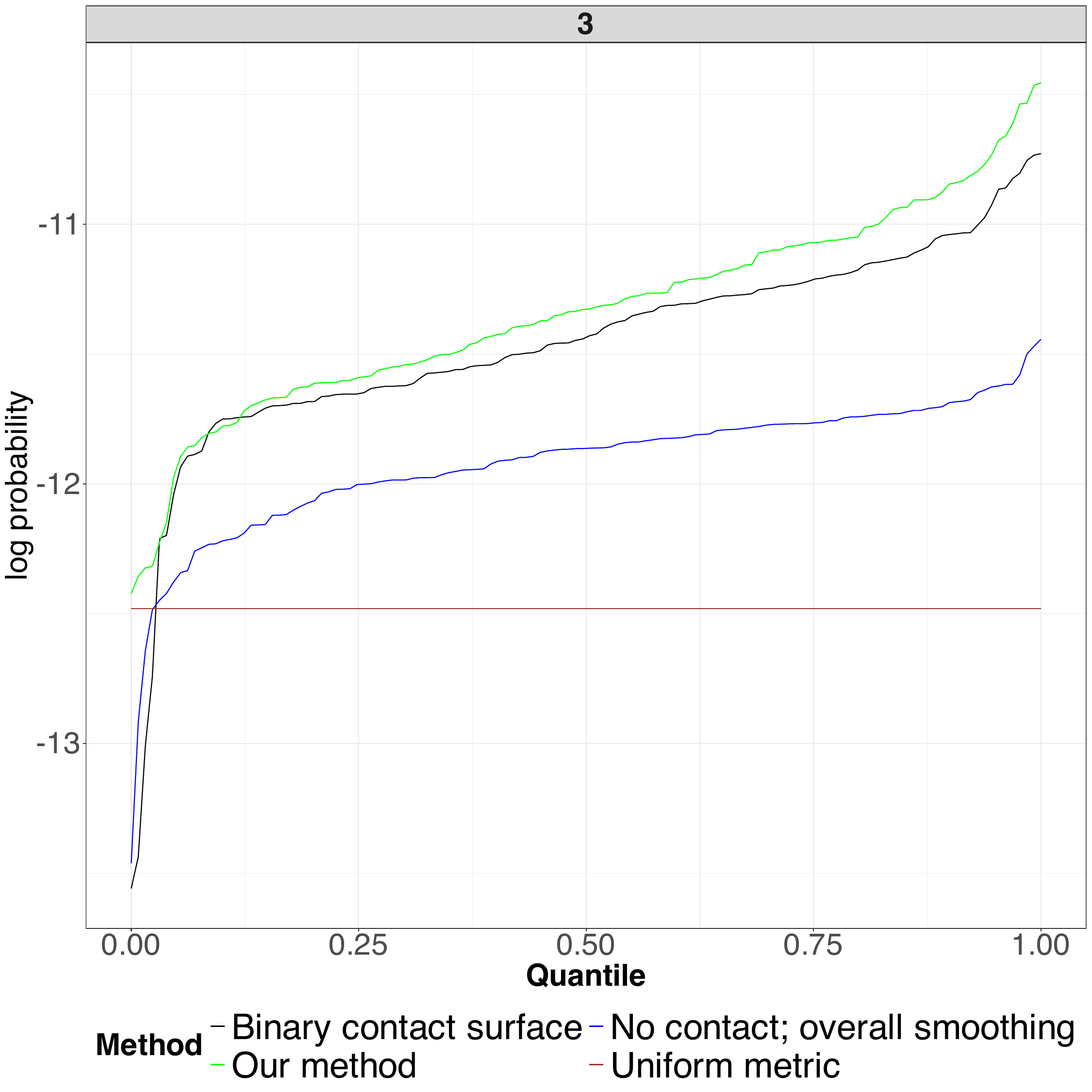}
        \caption{Quantile-based robustness check.}
        \label{fig:robustness_check_all_fold_with_m1}
    \end{subfigure}
    \caption{
    A comparison of four models from Table~\ref{tab:5_comparizon_1} according to (\ref{eqn:comparison_metric}) on the individual shoes in cross-validation Fold 3. In (a), the shoes are ordered identically according to increasing performance of Our method. In (b), the quantiles of performance are compared by ordering each method individually.
       % State-of-the-art comparison across footwear analysis methods. 
       % \footnotesize{(a) On the horizontal axis, shoe IDs are ordered by performance under our spatially varying coefficient model with image gradient. Methods without spatial covariates (e.g., uniform metric \cite{stone2006footwear} and no contact; overall smoothing \cite{yekutieli2012expert}) perform worse than those that incorporate spatial features \cite{spencer2020bayesian}. Our method, combining $2^6$ fixed effects, overall smoothing, and only 3 spatial random effects (pixel intensity, image gradient, and their interaction), performs best. 
       % (b) For fold 3, quantile comparisons ($Q_M(\tau;\mathcal{S}_{3}); \tau\in[0,1]$, Equation~\ref{eqn:quantile}) show our method (green) dominates others even at lower quantiles, highlighting robustness. A variant that excludes the image gradient (light blue) sits between our final method and a purely binary spatial model, indicating the added value of gradient features.}
    }
    \label{fig:combined_comparison_plots}
\end{figure}

While fold-averaged performance summarizes central tendency, it obscures whether differences are driven by a few cases or are sustained across all shoes. To examine this more closely, Figure~\ref{fig:combined_comparison_plots} compares the performance of our model against competing approaches on a shoe-by-shoe basis for the held-out shoes in Fold 3.

In Figure~\ref{fig:multinomial_comparizon_plot}, shoes are ordered along the horizontal axis according to increasing performance under our method. Our model consistently outperforms the uniform baseline and exhibits greater robustness than M(a) and M(b), both of which show a much steeper decline in performance in the lower tail. Moreover, our method outperforms M(a) on more than 90 percent of shoes. Although there is some correlation between the performance of our method and M(b), our approach yields superior performance for the vast majority of shoes. This robustness is further illustrated in Figure~\ref{fig:robustness_check_all_fold_with_m1}, which compares the empirical quantiles of the performance metric across all four methods. Across nearly the entire range of quantiles, our method dominates the competing approaches.

For clarity of visualization, M(Variant A) was omitted from Figure~\ref{fig:combined_comparison_plots}. Its performance is visually very similar to that of our method, though slightly worse on average, as shown in Table~\ref{tab:average_performance_euclidean_5methods}. Nevertheless, Our method outperforming M(Variant A) demonstrates the utility of parsimony and feature engineering: despite having far fewer parameters, Our method leverages important information in the image gradient to improve average performance. 

To further compare our model against closely related alternatives, Table~\ref{tab:average_performance_euclidean_model_selection} reports average performance across the ten folds for the model variants described in Table~\ref{tab:5_comparizon_1}. Performance is very similar across these models, particularly among M(Final), M(Variant C), and M(Variant D). Although M(Final) is the top performer in a plurality of folds, M(Variant C) and M(Variant D) exhibit marginally better average performance across folds. In contrast, M(Variant A) is consistently outperformed by the other variants, further demonstrating the utility of incorporating image gradient information into the model.

\begin{table}[ht!]
\centering
\renewcommand{\arraystretch}{} % Adjust row spacing
\resizebox{.8\textwidth}{!}{%
\begin{tabular}{|r|r|r|r|r|r|}
  \hline
Fold & \makecell{M(Final)/Our Method \\ IG, PI, \\ and interaction SV} & \makecell{IG and PI; SV \\ M(variant D)} & \makecell{PI; SV \\ M(variant C)} & \makecell{Only \\ overall Smoothing\\ M(variant B)} & \makecell{15 SV \\ M(Variant A)}\\ 
  \hline
1  & \textbf{-11.31081} & -11.31179 & -11.31330 & -11.31352 & -11.34449 \\ 
2  & -11.31446 & \textbf{-11.31259} & -11.31331 & -11.31343 & -11.33637 \\ 
3  & -11.33436 & -11.33304 & \textbf{-11.33258} & -11.33311 & -11.36233 \\ 
4  & \textbf{-11.29719} & -11.29756 & -11.29784 & -11.29828 & -11.32656 \\ 
5  & -11.32610 & -11.32652 & \textbf{-11.32515} & -11.32537 & -11.34773 \\ 
6  & \textbf{-11.42573} & -11.42637 & -11.42672 & -11.42743 & -11.45226 \\ 
7  & -11.35894 & -11.35701 & -11.35623 & \textbf{-11.35588} & -11.38762 \\ 
8  & \textbf{-11.28722} & -11.28781 & -11.28896 & -11.28934 & -11.33064 \\ 
9  & -11.32842 & -11.32716 & -11.32740 & \textbf{-11.32714} & -11.35275 \\ 
10 & -11.36413 & -11.36426 & \textbf{-11.36408} & -11.36464 & -11.39468 \\ 
\hline
Average
   & -11.33474 
   & \textbf{-11.33441} 
   & -11.33456 
   & -11.33481 
   & -11.36354 \\ 
\hline
\end{tabular} 
}
\caption{
Average performance of model variants as measured by (\ref{eqn:comparison_metric}) across ten folds. Top performer for each fold shown in bold. Models variants formulated as in Table~\ref{tab:5_comparizon_1}.
}
\label{tab:average_performance_euclidean_model_selection}
\end{table}

Given the very similar predictive performance of M(Final), M(variant D), and M(variant C), selecting a single recommended specification is not straightforward based exclusively on average cross-validation performance. The performance differences among these models are small in magnitude and vary slightly across folds. We therefore supplement the fold-averaged comparison with an additional metric designed to assess typical relative per-accidental performance at the level of individual shoes.

Specifically, we introduce the median performance loss ratio, which compares two any models $M_1$ and $M_2$ according to
$$
\mathcal{R}_{\mathcal{M}}(M_1, M_2)  = 100 \times\operatorname*{median}_{s \in \mathcal{S}} \left( \frac{\exp\!\bigl(m(Y_s=y_s \mid \mathcal{C}_s, M_1)\bigr)}
{\exp\!\bigl(m(Y_s=y_s \mid \mathcal{C}_s, M_2)\bigr)} \right)
$$
where $m(Y_s = y_s \mid \mathcal{C}_s, M)$ is defined in (equation~\ref{eqn:comparison_metric}), using the parameter $\thetavec$ estimated using the same 10-fold cross-validation procedure described above. This quantity measures the typical per-accidental predictive loss incurred by choosing $M_1$ in place of $M_2$. A value of 100 indicates parity in median performance, values below 100 indicate inferior performance of $M_1$ relative to $M_2$, and values above 100 indicate superior performance of $M_1$.

Table~\ref{tab:median_ratio_of_log_multinomial_m_final} reports this metric for each competing model from Table~\ref{tab:5_comparizon_1}, using M(Final) as the reference model $M_2$. The results show that M(Final) dominates all baseline and variant models. In particular, models that exclude contact surface information exhibit dramatic losses in median per-accidental predictive performance, with the binary contact surface of \cite{spencer2020bayesian} and M(Variant A) also having demonstrably worse performance.

\begin{table}[ht!]
\centering
\renewcommand{\arraystretch}{1.1}
\begin{tabular}{lc}
\hline
Method ($M_1$) & $\mathcal{R}_{\mathcal{M}}(M_1, M_2) $ (\%) \\
\hline
Intercept + PI + IG + interaction SV (M(Final)) & $1.00 \times 10^{2}$ \\
\hline
Intercept and PI SV (M(variant C)) & $.998 \times 10^{2}$ \\
Only intercept SV (M(variant B)) & $.998 \times 10^{2}$ \\
Intercept, IG and PI SV (M(variant D)) & $.994 \times 10^{2}$ \\
15 SV (excluding IG) (M(variant A)) & $.648 \times 10^{2}$ \\
\hline
Binary contact surface (M(b)) & $.13 \times 10^{2}$ \\
No contact; overall smoothing (M(a)) & $2.3 \times 10^{-3}$ \\
Uniform metric & $3.7 \times 10^{-9}$ \\
\hline
\end{tabular}
\caption{
median performance loss ratios comparing each method ($M_1$) against the proposed method M(Final).
Values below 100 indicate inferior performance relative to M(Final), while values equal to 100 indicate parity.
}
\label{tab:median_ratio_of_log_multinomial_m_final}
\end{table}

Among the closely related variants---M(variant D), M(variant C), and M(variant B)---median performance ratios are close to 100, indicating broadly comparable predictive behavior. Nevertheless, M(Final) attains the highest median performance while maintaining a small set of spatially varying coefficients related to pixel intensity and image gradients. In contrast, M(variant A), which relies on a substantially larger number of spatially varying coefficients but excludes image gradient information, performs consistently worse across metrics.

Taken together, these results motivate our choice of M(Final) as the preferred specification. Although its average predictive performance differs only marginally from that of M(variant D) and M(variant C), M(Final) offers a balance between predictive accuracy, interpretability, and parsimony. In particular, the inclusion of spatially varying image gradient effects provides additional explanatory insight into how the impact of the contact surface varies in space, as illustrated in Figure~\ref{fig:intercept_and_3_sv}, without incurring the complexity associated with high-dimensional spatial coefficient models. For these reasons, we adopt M(Final) as our recommended model for estimating reliable random match probabilities in forensic footwear analysis.

\section{Conclusion}
\label{conclusion}
% \textcolor{blue}{
% -Framed as GLM, -spatial varying coefficients, -image gradient, -used continuous contact surface, -larger data, -Big wins- state of the art, completely outperformed uniform
% }\neil{This could potentially use a light revision, with a little bit of future work.}

In this work, we developed a Bayesian spatial point process model for the distribution of accidental locations on footwear outsoles conditional on their contact surface. Our primary objective was to support the reliable estimation of accidental-based random match probabilities, which play a key role in interpreting forensic footwear evidence. Building on earlier work by \citet{spencer2020bayesian}, we reformulated this problem within a latent Gaussian model framework, enabling scalable Bayesian inference while simultaneously expanding the class of models that can be practically considered. Novel contributions of this study include considering a continuous-valued version of the contact surface image rather than a thresholded one, allowing the impact of the contact surface to vary spatially, and the inclusion of a novel image gradient terms such that our model specifically accounted for edges. 

We validated this framework using the WVU footwear database \citep{speir2016quantifying}, which is over three times as large as than the JESA database employed by \cite{spencer2020bayesian}. Our recommended specification consistently outperformed existing approaches in the literature as measured by predictive performance on held-out data. Importantly, this superior performance was not confined to a small subset of shoes; our model demonstrated superior robustness in the lower tail, where unreliable estimates pose the greatest forensic risk.

Despite these advances, several limitations remain before such models can be deployed directly in routine forensic casework. Our analysis relied on high-quality laboratory-generated impressions with precisely measured accidental locations. In practice, crime scene impressions are often partial and noisy, with only a subset of accidentals detectable. Understanding how probability models trained on laboratory data translate to these more challenging conditions remains an open problem. Recent work by \citet{smale2024estimate1, smale2024estimate2} represents important progress in this direction, but further study is needed to further bridge this gap.

There are also several promising methodological extensions. Alternative registration schemes, such as the polar-coordinate approaches proposed by \citet{richetelli2022spatial}, could potentially lead to further improvement when incorporated within the latent Gaussian framework. Additional pre-processing steps, including standardization of pixel intensity across shoes, may further improve data pooling. Finally, while our current model assumes linear effects of covariates on accidental intensity, nonlinear extensions using generalized additive models or spline-based approaches \citep{hastie1986generalized, perperoglou2019review}, as well as clustered point process models such as Neyman–Scott processes \cite{wang2024spatiotemporal}, represent natural avenues for future research.

Overall, this work further demonstrates that modern Bayesian spatial modeling tools can substantially improve the estimation of accidental-based random match probabilities in forensic footwear analysis. By combining scalable computation with a large database, the proposed framework provides a principled foundation for future methodological development and, ultimately, for more reliable assessment of forensic footwear evidence in forensic practice.

\section{Acknowledgment}
\label{Acknowledgment}
We sincerely thank Jacqueline A. Speir for providing the forensic shoeprint dataset used in this study, as well as helpful guidance and feedback. We acknowledge the continuous support from the INLA developers during the implementation and experimentation with different models. We also acknowledge the generous support of the University of Connecticut High Performance Computing (HPC) facilities and sincerely thank its maintainers for ensuring a stable and efficient computing environment for our intensive calculations.

\clearpage
\bibliographystyle{imsart-nameyear} % Style BST file
\bibliography{shoeprint}       % Bibliography file (usually '*.bib')

%% or include bibliography directly:
% \begin{thebibliography}{}
% \bibitem[\protect\citeauthoryear{???}{???}]{b1}
% \end{thebibliography}

\clearpage

\section{Appendix}
\label{appendix}

We divide the appendix into the following sections:
\begin{enumerate}
    \item Section \ref{INLA_procedure} describes the detailed procedure that is used to evaluate the posterior parameter inference using integrated nested Laplace approximation (INLA).
    \item Section \ref{multinomial_poisson_liklihood} describes the factorization of likelihood into Poisson and multinomial probability mass functions. 
    %\item To avoid the numerical error we described our procedure for the Multinomial probability mass function in the section \ref{multinomial_PMF_evaluation}.
    %\item Additional discussion on the comparison metric is provided in the section~\ref{Comparison_metric}.
    \item Section~\ref{detailed_description} provides Table~\ref{tab:5_comparizon_2}, which summarizes the various competing models.
    %\item We provide some literature review on the forensic modeling of accidentals in the section~\ref{literature}.
    \item Section~\ref{image_gradient} includes a description of the image gradient/sobel function employed in our model to determine edges.
    \item Section~\ref{additional_comparative_study} reports additional comparison metrics across the competitor models.
    %We summarized our results in the section 7 of our main paper. In addition, we would like to provide additional summary of the results in the Table~\ref{tab:gain_matrix}, \ref{tab:median_ratio_of_log_multinomial}, and \ref{tab:ratio_matrix_sv} for the validation of the performance of our model. Table~\ref{tab:gain_matrix} describes the average performance across different methods, whereas Table~\ref{tab:median_ratio_of_log_multinomial}, and \ref{tab:ratio_matrix_sv} describes the median based comparison among different competitive methods described in the Table 1 in the main document (Table~\ref{tab:5_comparizon_2}).
    \item Section~\ref{figures} includes supplementary Figures.
\end{enumerate}

\textbf{The general set-up of our model: }Recall that the main model was defined as 
\begin{align}
\label{model_spatial_varying_coefficients}
    &Y_{s,a}| \mathcal{C}_s \sim \text{Poisson}\left(\lambda_{s,a}\right);s\in\mathcal{S}, a\in\mathcal{A};\\
    & x_{s,a}^{\mathbf{i}} = (\mathcal{C}_{s,ij})^{i_{1}}
    (\mathcal{C}_{s,(i-1)j})^{i_{2}}
    (\mathcal{C}_{s,(i+1)j})^{i_{3}}
    (\mathcal{C}_{s,i(j-1)})^{i_{4}}
    (\mathcal{C}_{s,i(j+1)})^{i_{5}}
    (I_{s,ij})^{i_{6}};\quad \mathbf{i} \in \{0,1\}^{6};\\
    \label{link_function}
&\log(\lambda_{s,a})=\beta^{\text{smooth}}_{a}+ \beta^{\text{shoe}}_{s}+\sum_{\mathbf{i}\in\mathcal{I}, \mathcal{I}^{*} \subseteq \mathcal{I}} x_{s,a}^{\mathbf{i}}
    (\beta_{f}^{\mathbf{i}}+\beta_{\text{sv},a}^{\mathbf{i}} \mathbf{1}(\mathbf{i}\in \mathcal{I}^{*}));\quad \mathcal{I} \subseteq \{0,1\}^{6}.
\end{align}

Consequently, the likelihood can be written as: 
\begin{align}
\label{likelihood}
%\label{posterior}
\prod_{s=1}^{|\mathcal{S}|}\mathbb{P}(\bY_{s}=\by_{s}|\mathcal{C}_{s},\thetavec)
 &=\prod_{s=1}^{|\mathcal{S}|}\prod_{a\in\mathcal{A}}  \mathbb{P}(Y_{s,a}=y_{s,a}|\mathcal{C}_{s},\thetavec)= \prod_{s=1}^{|\mathcal{S}|}\prod_{a\in\mathcal{A}} \frac{e^{-\lambda_{s,a}} \lambda_{s,a}^{y_{s,a}}}{y_{s,a}!}.
\end{align}
Following the notation of \cite{rue2009approximate}, the latent Gaussian field is defined as
\[
\textbf{x}
=
\Big(
(\beta^{\text{smooth}}_{a})_{a\in\mathcal{A}},
\;
(\beta^{\text{shoe}}_{s})_{s\in\mathcal{S}},
\;
(\beta^{\mathbf{i}}_{f})_{\mathbf{i}\in\mathcal{I}},
\;
(\beta^{\mathbf{i}}_{\text{sv},a})_{a\in\mathcal{A},\,\mathbf{i}\in\mathcal{I}^*}
\Big)^{\top}.
\]
The linear predictor for observation $(s,a)$ is given by
\[
\eta_{s,a}
=
\beta^{\text{smooth}}_{a}
+
\beta^{\text{shoe}}_{s}
+
\sum_{\mathbf{i}\in\mathcal{I}}
x_{s,a}^{\mathbf{i}}\,
\beta_{f}^{\mathbf{i}}
+
\sum_{\mathbf{i}\in\mathcal{I}^*}
x_{s,a}^{\mathbf{i}}\,
\beta_{\text{sv},a}^{\mathbf{i}}\text{ ;}
\]
where $\log(\lambda_{s,a})=\eta_{s,a}$.
Let 
$
\boldsymbol{\psi}
=
\left(
\tau,\,
\tau_{sm},\,
\{\tau_i\}_{i\in\mathcal{I}^*}
\right)
$
denote the collection of hyperparameters. Then, up to proportionality, the conditional density is the following:
\begin{align}
\pi(\textbf{x}\mid \boldsymbol{y}, \boldsymbol{\psi})
\;\propto\;&
\prod_{s\in\mathcal{S}}
\prod_{a\in\mathcal{A}}
\exp\!\left(
y_{s,a}\,\eta_{s,a}
-
e^{\eta_{s,a}}
\right)
\times
\exp\!\left(
-\frac{1}{2}
\textbf{x}^{\top}
\boldsymbol{Q}(\boldsymbol{\psi})
\textbf{x}
\right),
\label{posterior_lgm}
\end{align}
where the precision matrix $\boldsymbol{Q}(\boldsymbol{\psi})$ is block diagonal with blocks
\[
\boldsymbol{Q}(\boldsymbol{\psi})
=
\operatorname{diag}\!\left(
\tau_{sm}\boldsymbol{Q}_{\text{sm}},
\;
\tau \boldsymbol{I},
\;
\boldsymbol{Q}(\boldsymbol{\beta}_{f}),
\;
\{\tau_i \boldsymbol{Q}_i\}_{i\in\mathcal{I}^*}
\right).
\]
\subsection{INLA procedure}
\label{INLA_procedure}
% We need to compute the full posterior and/or marginal posterior expectations. Markov Chain Monte Carlo (MCMC) is a powerful statistical technique used for sampling from complex probability distributions. However, it has its challenges and potential issues, for example, convergence issues due to poor choice of initial values, autocorrelation, burn-in period identification, etc. Moreover, MCMC can be computationally intensive, especially for large datasets or high-dimensional parameter spaces. 

Integrated Nested Laplace Approximation (INLA) is a statistical method that provides a computationally efficient alternative to traditional Markov Chain Monte Carlo (MCMC) methods (for example, STAN, \cite{carpenter2017stan}) for Bayesian inference for Latent Gaussian models. INLA is employed when exact Bayesian inference is computationally challenging, especially in cases involving high-dimensional and hierarchical models. It combines Laplace approximation with numerical integration techniques to provide fast and accurate approximations of posterior distributions. 

%%We are implementing INLA for computational advantage and estimation of the parameters. 
% For computational advantage in INLA, one small error term $\epsilon \sim \mathcal{N}(0,\tau^2_{\epsilon})$ is added in link function \ref{link_function}. It can relax the assumption of independence of the prior. If $\tau^2_{\epsilon}=0$, we can shift to the independence assumption of prior, which we assumed for calculating the posterior in \eqref{posterior}. 
% By choosing  INLA will not attempt to estimate the previous posterior distribution mentioned in \eqref{posterior} but 
Following the notations from our paper, let $\yvec=\left(\yvec_s\right)_{s\in\mathcal{S}}$. We are interested in the marginals of the parameters $\thetavec$ for individual component $\theta_{i}$ and hyperparameters $\psivec$ for individual component $\psi_{k}$. So; to calculate marginal posteriors, which are the following: 
\begin{align}
   p(\theta_{i}|\yvec) &= \int p(\theta_{i}|\yvec,\psivec) p(\psivec|\yvec) \,d\psivec; \\ 
   p(\psi_{k}|\yvec)&= \int p(\psivec|\yvec) \,d\psivec_{-k}. 
\end{align}

To compute the above we need to calculate $p(\psivec|\yvec)$ and $p(\theta_{i}|\yvec,\psivec)$. To calculate $p(\psivec|\yvec)$ we observe that we can approximate it with the Gaussian approximation of $p(\thetavec|\psivec,\yvec)$ as $\tilde{p}(\thetavec|\psivec,\yvec)$ at the estimated mode of $p(\thetavec|\yvec,\psivec)$ which is $\thetavec^{*}(\psivec):=\argmax_{\thetavec} p(\thetavec|\psivec,\yvec)$. 

One can express 
\begin{align}
\log(p(\thetavec|\yvec,\psivec))&=\log(p(\yvec|\thetavec,\psivec))+\log(p(\thetavec|\psivec))+\text{constant}.
\end{align}
Now one should compute the Taylor series expansion of $f(\thetavec)= \log(p(\yvec|\thetavec,\psivec))$ around $\thetavec^{*}(\psivec)$ and approximate with Gaussian distribution with precision matrix as $Q(\thetavec)+c(\thetavec)$ where $c(\thetavec)$ contains negative second derivative at $\thetavec^{*}(\psivec)$ with respect to $\thetavec$ and the mean parameter is $\thetavec^{*}(\psivec)$. Observe that,
\begin{align}
   p(\psivec|\yvec)= \frac{p(\thetavec,\psivec|\yvec)}{p(\thetavec|\psivec,\yvec)}\propto
   \frac{p(\yvec|\thetavec)p(\thetavec|\psi)p(\psi)}{p(\thetavec|\psivec,\yvec)}
   \approx \frac{p(\yvec|\thetavec)p(\thetavec|\psi)p(\psi)}{\tilde{p}(\thetavec|\psivec,\yvec)}\Big|_{\thetavec=\thetavec^{*}(\psivec)}
   := \tilde{p}(\psivec|\yvec).
\end{align}

Let $\thetavec_{-i}$ is the vector of the parameters excluding the $i$th component. The other integration is also calculated similarly, as follows. We also approximate $p(\thetavec_{i}|\theta_{i},\psivec,\yvec)$ by $\tilde{p}(\thetavec_{-i}|\theta_{i},\psivec,\yvec)$ at the estimated mode $\thetavec^{*}_{-i}(\theta_{i},\psivec)$.
\begin{align}
   p(\theta_{i}|\yvec,\psivec)= \frac{p((\theta_{i},\thetavec_{-i})|\yvec,\psivec)}{p(\thetavec_{-i}|\theta_{i},\psivec,\yvec)}\propto
   \frac{p((\theta_{i},\thetavec_{-i})|\yvec,\psivec)}{\tilde{p}(\thetavec_{-i}|\theta_{i},\psivec,\yvec)}\Big|_{\thetavec_{-i}=\thetavec^{*}_{-i} (\theta_{i},\psivec)}
   :=  \tilde{p}(\theta_{i}|\yvec,\psivec).
\end{align}

Finally, we perform a numerical integration to compute $p(\theta_{i}|\yvec)$.
\begin{equation}
\begin{split}
   p(\theta_{i}|\yvec)\approx \sum_{k=1}^{K} \tilde{p}(\theta_{i}|\yvec,\psi_{k}) \tilde{p}(\psi_{k}|\yvec) \Delta_{k}, \\ 
\end{split}
\end{equation}
where $K$ is the number of hyperparameters with their corresponding weights $\Delta_{k}$ for $k=1,\cdots, K$. A complete description can be found in \cite{rue2009approximate}, \cite{blangiardo2013spatial}, \cite{rue2017bayesian}, etc.

In practice, INLA is most computationally efficient when the number of hyperparameters is relatively small (typically fewer than about 10).

\subsection{Multinomial Poisson likelihood}
\label{multinomial_poisson_liklihood}

Recall that the linear predictor can be written as
\[
\eta_{s,a}
=
\beta^{\text{smooth}}_{a}
+
\beta^{\text{shoe}}_{s}
+
\sum_{\mathbf{i}\in\mathcal{I}}
x_{s,a}^{\mathbf{i}}\,
\beta_{f}^{\mathbf{i}}
+
\sum_{\mathbf{i}\in\mathcal{I}^*}
x_{s,a}^{\mathbf{i}}\,
\beta_{\text{sv},a}^{\mathbf{i}}.
\]
This can be decomposed as $\eta_{s,a}=\eta_{s,a}^{1}+ \beta^{\text{shoe}}_{s}$, where 
\[\eta_{s,a}^{1}=\beta^{\text{smooth}}_{a}
+
\sum_{\mathbf{i}\in\mathcal{I}}
x_{s,a}^{\mathbf{i}}\,
\beta_{f}^{\mathbf{i}}
+
\sum_{\mathbf{i}\in\mathcal{I}^*}
x_{s,a}^{\mathbf{i}}\,
\beta_{\text{sv},a}^{\mathbf{i}}.\]
Now, $\mathbb{P}(\bY_{s}=\by_{s} |\mathcal{C}_{s})$ can be written as:
\begin{align*}
       & \mathbb{P}(\bY_{s}=\by_{s} |\mathcal{C}_{s})
        \\
       & = \prod_{a\in\mathcal{A}}\frac{\exp(-\lambda_{s,a})(\lambda_{s,a})^{y_{s,a}}}{y_{s,a}!}  \\
        &=  \frac{\exp(-\sum_{a\in\mathcal{A}}\exp(\eta_{s,a}^{1}+\beta^{\text{shoe}}_{s}))  (\prod_{a\in\mathcal{A}} \exp(\eta_{s,a}^{1}+\beta^{\text{shoe}}_{s})^{y_{s,a}}) }{\prod_{a\in\mathcal{A}}y_{s,a}!}\\
 &= \frac{\exp(-\sum_{a\in\mathcal{A}}\exp(\eta_{s,a}^{1}+\beta^{\text{shoe}}_{s}))}{\prod_{a\in\mathcal{A}}y_{s,a}!} \times
 \\
 & \times (\sum_{a\in\mathcal{A}}\exp(\eta_{s,a}^{1}+\beta^{\text{shoe}}_{s}))^{N_{s}} \prod_{a\in\mathcal{A}}\frac{\exp(\eta_{s,a}^{1}+\beta^{\text{shoe}}_{s})^{y_{s,a}}}{(\sum_{a\in\mathcal{A}}\exp(\eta_{s,a}^{1}+\beta^{\text{shoe}}_{s}))^{y_{s,a}}}\\
 &= \frac{\exp(-\sum_{a\in\mathcal{A}}\exp(\eta_{s,a}^{1}+\beta^{\text{shoe}}_{s})) (\sum_{a\in\mathcal{A}}\exp(\eta_{s,a}^{1}+\beta^{\text{shoe}}_{s}))^{N_{s}}}{N_{s}!} \times \\ 
 &  \times \frac{N_{s}!}{\prod_{a\in\mathcal{A}}y_{s,a}!} \prod_{a\in\mathcal{A}}\frac{\exp(\eta_{s,a}^{1}+\beta^{\text{shoe}}_{s})^{y_{s,a}}}{(\sum_{a\in\mathcal{A}}\exp(\eta_{s,a}^{1}+\beta^{\text{shoe}}_{s}))^{y_{s,a}}},
\end{align*} 
where $N_s :=\sum_{a\in\mathcal{A}}y_{s,a}$. Note that this function can be seen as a product of two parts. The first part is the probability mass function, evaluated at $N_{s}$, of a Poisson distribution with parameter 
\begin{align}
\sum_{a\in\mathcal{A}}\exp(\eta_{s,a}^{1}+\beta^{\text{shoe}}_{s})=\exp(\beta^{\text{shoe}}_{s})\sum_{a\in\mathcal{A}}\exp(\eta_{s,a}^{1}).   
\end{align}
The second part is the probability mass function, evaluated at $(y_{sa})_{a \in \mathcal{A}}$, of a multinomial distribution with parameters 
\[
\left(N_{s},\left\{\pi_{a} \right\}:= \left\{\frac{\exp(\eta_{s,a}^{1}+\beta^{\text{shoe}}_{s})}{(\sum_{a\in\mathcal{A}}\exp(\eta_{s,a}^{1}+\beta^{\text{shoe}}_{s}))}\right\};\;a\in\mathcal{A}\right).
\]
Note that the probabilities for the multinomial component are independent of $\beta^{\text{shoe}}_{s}$; the numerator and denominator of $\frac{\exp(\eta_{s,a}^{1}+\beta^{\text{shoe}}_{s})^{y_{s,a}}}{(\sum_{a\in\mathcal{A}}\exp(\eta_{s,a}^{1}+\beta^{\text{shoe}}_{s}))^{y_{s,a}}}$; $\exp(\beta^{\text{shoe}}_{s})^{y_{s,a}}$ will be canceled out. Hence, we can write it as:
\begin{align}
%\begin{split}
 \frac{\exp(\eta_{s,a}^{1}+\beta^{\text{shoe}}_{s})}{(\sum_{a\in\mathcal{A}}\exp(\eta_{s,a}^{1}+\beta^{\text{shoe}}_{s}))}
 &=\frac{\exp(\eta_{s,a}^{1}) \exp{(\beta^{\text{shoe}}_{s})}}{(\sum_{a\in\mathcal{A}}\exp(\eta_{s,a}^{1}))
 \exp{(\beta^{\text{shoe}}_{s}})}\\
 &=\frac{\exp(\eta_{s,a}^{1}) }{(\sum_{a\in\mathcal{A}}\exp(\eta_{s,a}^{1}))}.
%\end{split}
\end{align}

The multinomial part of the likelihood is thus free of $\beta^{\text{shoe}}_{s}$. 

\subsection{Detailed description of different models}
\label{detailed_description}
A detailed description of the methods can also be found in Table~\ref{tab:5_comparizon_2}.
\begin{table}[ht!]
    \centering
    \adjustbox{width=\textwidth,center}{
    \begin{tabular}{|p{1.8cm}|p{1.37cm}|p{4.5cm}|p{.7cm}|p{.8cm}|p{3.1cm}|p{1 cm}|p{1.3cm}|}
        \hline
        Model &  \begin{tabular}[c]{@{}l@{}}Binary($B$)\\None($\times$),\\Continuous\\($C$)\end{tabular} &Description & 32 (fixed) & sobel fixed & Spatially varying & \begin{tabular}[c]{@{}l@{}}sobel\\ spatially \\varying \end{tabular} & \begin{tabular}[c]{@{}l@{}}Overall \\ smoothing \end{tabular} \\
        \hline
       Uniform   & $\times$ &uniform metric   & $\times$   &  $\times$  & $\times$   & $\times$   &  $\times$   \\
        \hline
        M(a)   & $\times$ & spatially varying intercept   &  $\times$ & $\times$  & $\times$  & $\times$  & \checkmark  \\
        \hline
        M(b)  & $B$ & spatially varying intercept and 32 fixed effects ($\mathcal{N}^s_{a}$ with interaction) & \checkmark  & $\times$  & $\times$ & $\times$ & \checkmark  \\
        \hline
        \hline
        \begin{tabular}[c]{@{}l@{}}M(variant A) \end{tabular} & $C$ & spatially varying intercept \textit{with 15 spatially varying (upto 2nd order interactions of $\mathcal{N}^s_{a}$ )}, and 32 fixed effects ($\mathcal{N}^s_{a}$ with interaction)& \checkmark & $\times$  & 15 (upto 2nd order) & $\times$  &  \checkmark \\
        \hline
        \hline
        \begin{tabular}[c]{@{}l@{}}M(variant B) \end{tabular} & $C$ & spatially varying intercept, 64 fixed effect (\textbf{sobel} and $\mathcal{N}^s_{a}$ with interaction)  & \checkmark  & \checkmark  & $\times$ & $\times$  & \checkmark  \\
        \hline
        \begin{tabular}[c]{@{}l@{}}M(variant C) \end{tabular} & $C$ & spatially varying intercept \textit{and pixel intensity}, 64 fixed effect (\textbf{sobel} and $\mathcal{N}^s_{a}$ with interaction)  & \checkmark  & \checkmark  & 1 (pixel intensity)  & \checkmark  & \checkmark  \\
        \hline
        \begin{tabular}[c]{@{}l@{}}M(variant D) \end{tabular}  & $C$ & spatially varying intercept, \textit{pixel intensity and \textbf{sobel}}; 64 fixed effect (\textbf{sobel} and $\mathcal{N}^s_{a}$ with interaction) & \checkmark  & \checkmark  & \begin{tabular}[c]{@{}l@{}} 2 \\(pixel intensity \\+ sobel ) \end{tabular} & \checkmark  & \checkmark  \\
        \hline
        \begin{tabular}[c]{@{}l@{}}M(Final) \end{tabular} & $C$ & spatially varying intercept, \textit{pixel intensity, \textbf{sobel} and interaction between sobel and pixel intensity}; 64 fixed effect (\textbf{sobel} and $\mathcal{N}^s_{a}$ with interaction)  & \checkmark  & \checkmark  & \begin{tabular}[c]{@{}l@{}} 2 \\(pixel intensity +\\pixel intensity * sobel )\end{tabular}  & \checkmark  & \checkmark  \\
        \hline
    \end{tabular}
    }
    \caption{Description of Competitor models.The neighborhood is defined as $\mathcal{N}^s_{a}:=(\mathcal{C}_{s,a},
    \mathcal{C}_{s,a - h},
    \mathcal{C}_{s,a + h},
    \mathcal{C}_{s,a - v},
    \mathcal{C}_{s,a + v})$ and the sobel $(I_{s,a})$ in our main paper.}
    \label{tab:5_comparizon_2}
\end{table}

\section{Image gradient}
\label{image_gradient}
The Sobel operator or image gradient is a discrete differentiation operator used to compute an approximation of the gradient of the image intensity function. A high-pass filter enhances rapid intensity changes in an image, which correspond to edges and fine details, while suppressing smooth areas with gradual transitions or low frequencies.
The Sobel operator uses two convolution kernels, one for detecting changes in the horizontal direction (\( G_x \)) and another for detecting changes in the vertical direction (\( G_y \)). These kernels are defined as:

\[
G_x = 
\begin{bmatrix}
-1 & 0 & 1 \\
-2 & 0 & 2 \\
-1 & 0 & 1
\end{bmatrix}; \qquad
G_y = G_x ^{T} .
\]

Let the horizontal filter produce (\(G_{x,a}: \mathcal{\bar{N}}^s_{a} \times G_x \rightarrow \mathbb{R}\)) and the vertical filter produce (\(G_{y,a}: \mathcal{\bar{N}}^s_{a} \times G_y \rightarrow \mathbb{R} \)) for all coordinates $a\in\mathcal{A}$. 

For computational efficiency, particularly when processing large images or multiple frames, we implement the convolution operations using Fast Fourier Transform (FFT). The convolution theorem states that convolution in the spatial domain is equivalent to element-wise multiplication in the frequency domain: 
$$\text{conv}(f,g) = \mathcal{F}^{-1}[\mathcal{F}(f) \cdot \mathcal{F}(g)],$$
where $\mathcal{F}$ denotes the FFT operation. For an image of size $n_{x} \times n_{y}$ pixels, this approach reduces the computational complexity from $O(n_{x} n_{y} \cdot k^2)$ for direct spatial convolution with a $k \times k$ kernel to $O(n_{x} n_{y} \log (n_{x} n_{y}))$ for FFT-based convolution (see \cite{vetterli1984simple}, \cite{pau2010ebimage}, e.t.c.). Implementing FFT-based convolution is particularly advantageous for high-resolution images. The horizontal and vertical gradient components are calculated using FFT-based convolution as:
$$
G_{x,a} = \mathcal{F}^{-1}[\mathcal{F}(\mathcal{\bar{N}}^s_{a}) \cdot \mathcal{F}(G_x)],\quad
G_{y,a} = \mathcal{F}^{-1}[\mathcal{F}(\mathcal{\bar{N}}^s_{a}) \cdot \mathcal{F}(G_y)],
$$
where $\mathcal{\bar{N}}^s_{a}$ represents the neighborhood around pixel $a$, and the convolution operations are efficiently computed using FFT for all coordinates $a\in\mathcal{A}$. 

The gradient magnitude at each pixel is then computed as:
\[
G_{a} = \sqrt{G_{x,a}^2 + G_{y,a}^2} \in \mathbb{R}; \quad \forall a\in\mathcal{A}.
\]

\section{Additional comparative study from other methods}
\label{additional_comparative_study}

\textbf{CCC}: The Concordance Correlation Coefficient (CCC \cite{lawrence1992assay}) is a widely used metric for assessing the agreement between two methods of continuous measurement.
In the context of two of our competitor models $M_1$ and $M_2$, the CCC evaluates the agreement between
\[
p_{M_1} := \left\{ m(Y_s = y_s \mid \mathcal{C}_s, M_1) : s \in \mathcal{S} \right\}
\quad \text{and} \quad
p_{M_2} := \left\{ m(Y_s = y_s \mid \mathcal{C}_s, M_2) : s \in \mathcal{S} \right\},
\]
and is defined as
\[
\mathrm{CCC}_{M_1}^{M_2}
=
\rho_{p_{M_1}}^{p_{M_2}}
\times
\frac{2}{\nu + \nu^{-1} + u^2},
\]
where $\rho_{p_{M_1}}^{p_{M_2}}$ denotes the Pearson correlation coefficient between
the vectors $p_{M_1}$ and $p_{M_2}$,
\[
\nu = \frac{s_{p_{M_1}}}{s_{p_{M_2}}}
\]
is the ratio of their sample standard deviations, and
\[
u = \frac{\mu_{p_{M_1}} - \mu_{p_{M_2}}}
{\sqrt{s_{p_{M_1}}\, s_{p_{M_2}}}}
\]
is the standardized difference between their sample means.
Here, $\mu_{p_{M_j}}$ and $s_{p_{M_j}}$ denote the sample mean and standard deviation
of the components in the vector $p_{M_j}$, respectively, for $j \in \{1,2\}$. 

\begin{table}[ht!]
\centering
\renewcommand{\arraystretch}{1}
\begin{tabular}{rllll}
  \hline
Methods  & \makecell[r]{Our method\\ (M(Final))} 
  & \makecell[r]{15 SV\\ (M(variant A))} 
  & \makecell[r]{Binary contact\\surface (M(b))} 
  & \makecell[r]{No contact; overall\\ smoothing (M(a))} \\
  \hline
  \makecell[r]{M(Final)} & 1.00 & 0.96 [0.96, 0.97] & 0.76 [0.73, 0.78] & 0.13 [0.11, 0.15] \\
  \makecell[r]{M(variant A)} & 0.96 [0.96, 0.97] & 1.00 & 0.76 [0.74, 0.79] & 0.15 [0.12, 0.17] \\
  \makecell[r]{M(b)} & 0.76 [0.73, 0.78] & 0.76 [0.74, 0.79] & 1.00 & 0.17 [0.15, 0.20] \\
  \makecell[r]{M(a)} & 0.13 [0.11, 0.15] & 0.15 [0.12, 0.17] & 0.17 [0.15, 0.20] & 1.00 \\
  \hline
\end{tabular}
\caption{Concordance correlation coefficients with 95\% confidence intervals between model configurations. \cite{lawrence1992assay}, used CCC package from Desktools.}
\label{tab:ccc}
\end{table}

Note that CCC values close to one indicating strong agreement between two methods $p_{M_1} := \left\{ m(Y_s = y_s \mid \mathcal{C}_s, M_1) : s \in \mathcal{S} \right\}
$ and $p_{M_2} := \left\{ m(Y_s = y_s \mid \mathcal{C}_s, M_2) : s \in \mathcal{S} \right\}
$ Table~\ref{tab:ccc} presents pairwise CCC with 95\% confidence intervals between different model configurations. Our method shows the highest agreement with the 15 SV model (CCC = 0.96), indicating similar predictive performance. Moderate agreement is observed with the binary contact surface model (CCC = 0.76), while the no-contact smoothing baseline shows minimal concordance with all other methods (CCC < 0.17), highlighting the added value of contact-based features.

\textbf{Average percentage gain: }To summarize the results across different folds (Table 1 in the main paper), we present another metric:
\begin{equation}
\label{eqn:avg_across_folds}
g(M_{2},M_{1}):=\exp\left(\frac{1}{10}\sum_{i=1}^{10} (P_{\mathcal{S}_{i}}^{M_2}-P_{\mathcal{S}_{i}}^{M_1})\right) \times 100,
\end{equation}
for two methods $M_1$ and $M_2$ in the Table~\ref{tab:gain_matrix}. If $g(M_{2},M_{1})> 100$, then we have positive gain for choosing the method $M_2$ over $M_1$, and similarly $g(M_{2},M_{1})< 100$ means $M_1$ is performing better than $M_2$.
\begin{table}[ht!]
\centering
\renewcommand{\arraystretch}{1} % Adjust row spacing
\resizebox{\textwidth}{!}{%
\begin{tabular}{rlllll}
  \hline
 Method& \begin{tabular}[c]{@{}l@{}}Uniform\end{tabular} 
 & \begin{tabular}[c]{@{}l@{}}M(a)(Only overall\\ smoothing)\end{tabular} 
 & \begin{tabular}[c]{@{}l@{}}M(b)\\ (binary)\end{tabular} 
 & \begin{tabular}[c]{@{}l@{}}M(variant A)\\ (Our variant)\end{tabular}
 & \begin{tabular}[c]{@{}l@{}}Our Method\\ M(Final)\end{tabular} \\ 
  \hline
Uniform                        & 100.00 & 176.92 & 273.41 & 305.38 & 311.30 \\ 
M(a)  & 56.52  & 100.00 & 154.54 & 172.62 & 175.96 \\ 
M(b)                   & 36.58  & 64.71  & 100.00 & 111.69 & 113.86 \\ 
M(variant A)               & 32.75  & 57.93  & 89.53  & 100.00 & 101.94 \\ 
Our method /M(Final)                    & 32.12  & 56.83  & 87.83  & 98.10  & 100.00 \\ 
  \hline
\end{tabular}%
}
\caption{
Average percentage gain of the method in the row method ($M_1$) compared to the column method ($M_2$), 
computed as $g(M_{2},M_{1})$ (Equation~\ref{eqn:avg_across_folds}). Above 100\% means positive gain, and below 100\% means negative gain. Our method account 75.96\% higher accidentals than overall smoothing, 13.86\% from the binary method, and 1.94\% from spatially varying, respectively.
}
\label{tab:gain_matrix}
\end{table}

Recall that, median percentage ratio, we consider the vector of all the log probabilities for method $M$ denoted as $\left\{ P(Y_s=y_s \mid \mathcal{C}_s, M) : s\in\mathcal{S} \right\}$. \\

\textbf{Median performance loss ratio: }
We defined the median performance loss ratio, which compares two models $M_1$ and $M_2$ according to
\begin{equation}
\label{eqn:median_ratio}
 \mathcal{R}_{\mathcal{M}}(M_1, M_2)  = 100 \times\operatorname*{median}_{s \in \mathcal{S}} \left( \frac{\exp\!\bigl(m(Y_s=y_s \mid \mathcal{C}_s, M_1)\bigr)}
{\exp\!\bigl(m(Y_s=y_s \mid \mathcal{C}_s, M_2)\bigr)} \right).   
\end{equation}
% For two methods $M_1$ and $M_2$, we compute the shoe-specific ratios:
% \[
% R_{s}^{M_{1},M_{2}}=
% \frac{\exp\!\bigl(m(Y_s=y_s \mid \mathcal{C}_s, M_1)\bigr)}
% {\exp\!\bigl(m(Y_s=y_s \mid \mathcal{C}_s, M_2)\bigr)},
% \qquad s\in\mathcal{S},
% \]
% and summarize the resulting vector $(R_{s}^{M_{1},M_{2}})_{s\in\mathcal{S}}$ by its median (Table~\ref{tab:median_ratio_of_log_multinomial} and Table~\ref{tab:ratio_matrix_sv}), i.e. 
% \begin{equation}
%     \label{eqn:median_ratio}
%     \widetilde{R}^{M_{1}}_{M_{2}}=
% \operatorname{median}_{s\in\mathcal{S}} R_{s}^{M_{1},M_{2}}.
% \end{equation}
%To quantify relative performance, we computed the median percentage ratios between methods using exponentiated log-probabilities on a per-shoe basis. 
For each pair of methods, we evaluated the ratio of probabilities for each observation and reported the median of these ratios, scaled by 100 for interpretability. A value much greater than 100 indicates that the method in the row significantly outperforms the method in the column. Table~\ref{tab:median_ratio_of_log_multinomial} presents the median percentage ratios between models. Our proposed method (M(Final)) consistently outperformed all alternative configurations across median percentage ratios. In particular, it showed substantial gains over the no-contact baseline and the uniform model, highlighting the effectiveness of incorporating spatially-informed, multi-level modeling in capturing complex signal structures.

\begin{table}[ht!]
\centering
\renewcommand{\arraystretch}{1} % Adjust row spacing
\resizebox{\textwidth}{!}{%
\begin{tabular}{rlllll}
  \hline
 Methods& \begin{tabular}[c]{@{}l@{}}No contact; overall\\ smoothing (M(a))\end{tabular} 
 & \begin{tabular}[c]{@{}l@{}}Binary contact\\ surface (M(b))\end{tabular} 
 & \begin{tabular}[c]{@{}l@{}}Spatially varying \\(M(variant A))\end{tabular} 
 & \begin{tabular}[c]{@{}l@{}}Our method \\(M(Final)) \end{tabular}
 & \begin{tabular}[c]{@{}l@{}}Uniform\\ metric\end{tabular} \\ 
  \hline
M(a) & $1.0 \times 10^{2}$ & $2.3 \times 10^{-2}$ & $3.8 \times 10^{-3}$ & $2.3 \times 10^{-3}$ & $6.1 \times 10^{6}$ \\ 
M(b) & $4.4 \times 10^{5}$ & $1.0 \times 10^{2}$ & $1.9 \times 10^{1}$ & $1.3 \times 10^{1}$ & $1.5 \times 10^{11}$ \\ 
M(variant A) & $2.6 \times 10^{6}$ & $5.2 \times 10^{2}$ & $1.0 \times 10^{2}$ & $6.5 \times 10^{1}$ & $1.7 \times 10^{12}$ \\ 
M(Final) & $4.4 \times 10^{6}$ & $7.7 \times 10^{2}$ & $1.5 \times 10^{2}$ & $1.0 \times 10^{2}$ & $2.7 \times 10^{12}$ \\ 
Uniform metric & $1.6 \times 10^{-3}$ & $6.5 \times 10^{-8}$ & $6.0 \times 10^{-9}$ & $3.7 \times 10^{-9}$ & $1.0 \times 10^{2}$ \\
  \hline
\end{tabular}%
}
\caption{
Median performance loss ratio between methods ($\mathcal{R}_{\mathcal{M}}(M_1, M_2)$, Equation~\ref{eqn:median_ratio}), where the row method ($M_1$) is compared to the column method ($M_2$). We computed the median percentage ratios between methods using exponentiated log-probabilities on a per-shoe basis. For each pair of methods, we evaluated the ratio of probabilities for each observation and reported the median of these ratios, scaled by 100 for interpretability. A value much greater than 100 indicates that the method in the row significantly outperforms the method in the column.
}
\label{tab:median_ratio_of_log_multinomial}
\end{table}

\begin{table}[ht!]
\centering
\renewcommand{\arraystretch}{1} % Adjust row spacing
\resizebox{\textwidth}{!}{%
\begin{tabular}{rlllll}
  \hline
  \begin{tabular}[c]{@{}l@{}}Methods\\(including $2^6$ \\ fixed effects)\end{tabular}
 & \begin{tabular}[c]{@{}l@{}}Our method \\ (M(Final)) \end{tabular}
 & \begin{tabular}[c]{@{}l@{}}Intercept, IG and PI;\\ SV (M(variant D))\end{tabular}
 & \begin{tabular}[c]{@{}l@{}}intercept, PI;\\ SV (M(variant C))\end{tabular}
 & \begin{tabular}[c]{@{}l@{}}Overall smoothing \\(only intercept SV)\\(M(variant B))\end{tabular} 
 & \begin{tabular}[c]{@{}l@{}}intercept, 15 SV \\(M(variant A)) \end{tabular}\\ 
  \hline
M(Final) & $1.00 \times 10^{2}$ & $1.01 \times 10^{2}$ & $1.00 \times 10^{2}$ & $1.00 \times 10^{2}$ & $1.54 \times 10^{2}$ \\ 
M(variant D) & $9.94 \times 10^{1}$ & $1.00 \times 10^{2}$ & $1.00 \times 10^{2}$ & $1.01 \times 10^{2}$ & $1.53 \times 10^{2}$ \\ 
M(variant C) & $9.98 \times 10^{1}$ & $9.98 \times 10^{1}$ & $1.00 \times 10^{2}$ & $1.00 \times 10^{2}$ & $1.50 \times 10^{2}$ \\ 
M(variant B) & $9.98 \times 10^{1}$ & $9.94 \times 10^{1}$ & $9.97 \times 10^{1}$ & $1.00 \times 10^{2}$ & $1.48 \times 10^{2}$ \\ 
M(variant A) & $6.48 \times 10^{1}$ & $6.55 \times 10^{1}$ & $6.65 \times 10^{1}$ & $6.73 \times 10^{1}$ & $1.00 \times 10^{2}$ \\

  \hline
\end{tabular}%
}
\caption{Median performance loss ratio ($\mathcal{R}_{\mathcal{M}}(M_1, M_2)$, Equation~\ref{eqn:median_ratio}) between spatial varying methods, where the row method ($M_1$) is compared to the column method ($M_2$). For M(Final)/Our Method (3 SV; pixel intensity, image gradient, and their interaction, with intercept SV); M(variant D)/Only image gradient and pixel intensity as SV with intercept SV; M(variant C) intercept and PI as SV; M(variant B)/Only intercept SV (overall smoothing); and M(variant A)/15 SV (first two order of interactions with 5 covariates excluding image gradient) with intercept SV across ten folds. The first 4 methods have image gradient as a fixed effect ($2^6$), and the fifth model has no image gradient in fixed effect.}
\label{tab:ratio_matrix_sv}
\end{table}

Table~\ref{tab:ratio_matrix_sv} presents the median percentage ratios between these variants, which are different spatially varying (SV) modeling strategies. The first four methods—Our method (M(Final)); Intercept, image gradient (IG) and pixel intensity (PI) as spatially varying (M(variant D)); Intercept, pixel intensity as spatially varying (M(variant C)); and Overall smoothing with only intercept spatially varying (M(variant B))—show highly comparable performance, with median ratios differing by only a few percent. Importantly, all these approaches include image gradient as a fixed effect along with $2^6$ fixed effects, underscoring its consistent value in the model. In contrast, the Intercept, 15 SV model, which omits image gradient from the fixed effects and includes a broader but potentially noisier set of SV terms, demonstrates notably lower performance. This highlights the relevance of targeted covariate inclusion—specifically, image gradient—in improving model fit and predictive quality. Thus, it justifies that our proposed model performed superior among additional variants of our models (M(Variant A), M(Variant B), M(Variant C), and M(Variant D)).

\section{Additional plots}
\label{figures}
\begin{figure}[ht!]
        \centering
        \includegraphics[scale=1.0,width=.8\linewidth]{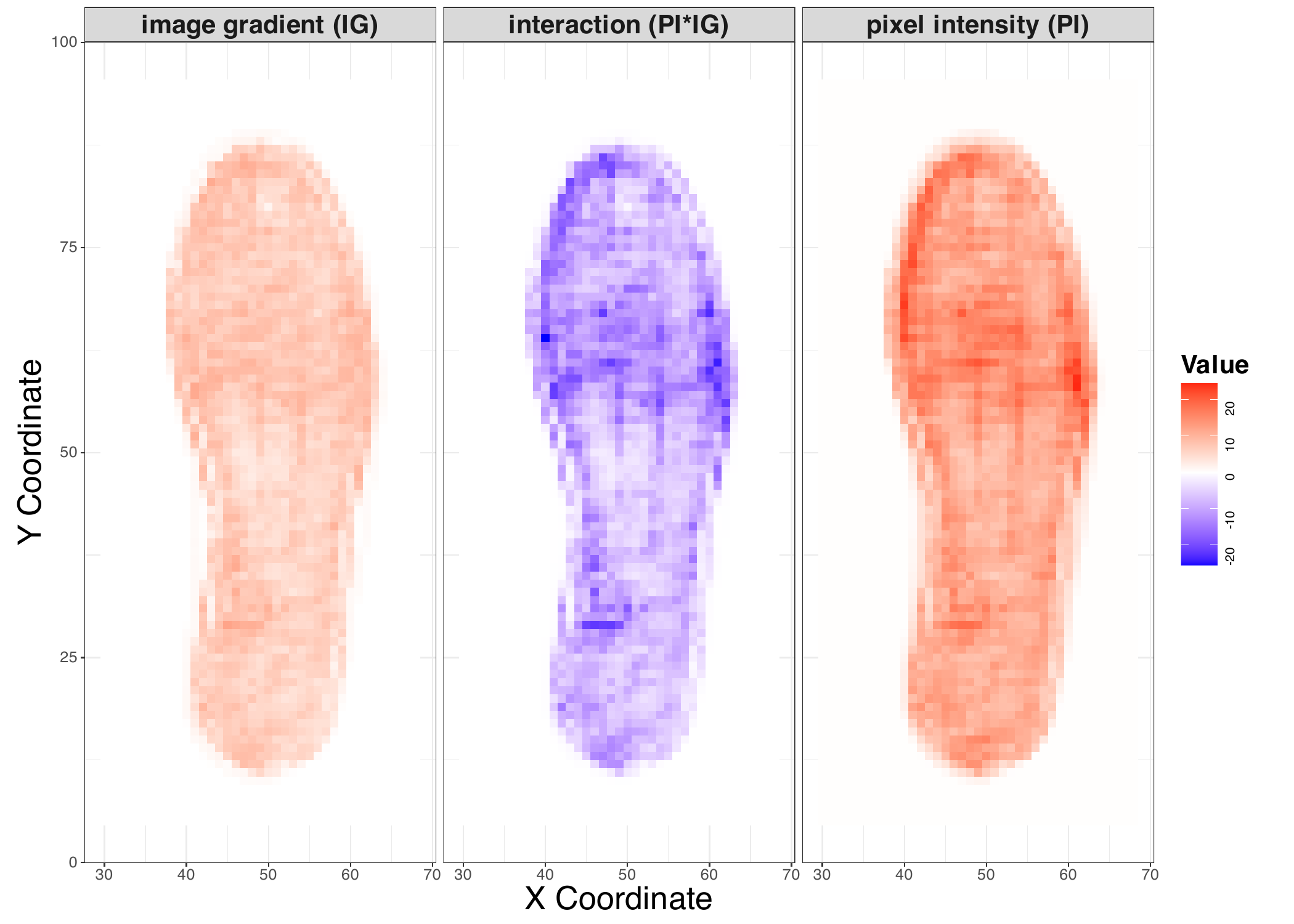}
        \caption{Heatmap for 3 spatially varying coefficients combined with fixed effect for M(Final).}
        \label{fig:heatmap_random_effect_4}
    \end{figure}

Our final recommended model has three spatially varying coefficients:
\begin{align*}
\beta_{\text{sv}}^{\mathbf{i_1}}\; \text{with} \; \mathbf{i_1}= (1,0,0,0,0,0) &   \; \;\text{(spatially varying effect of pixel intensity),}\; & \\
 \beta_{\text{sv}}^{\mathbf{i_2}} \; \text{with} \;  \mathbf{i_2}= (0,0,0,0,0,1)  & \;\; \text{(spatially varying effect of image gradient),} \text{ and}\\ 
 \beta_{\text{sv}}^{\mathbf{i_3}} \; \text{with} \;  \mathbf{i_3}= (1,0,0,0,0,1) &   \;\; \text{(spatially varying effect of their interaction)}.
 \end{align*}
Figure 5 of the main text illustrates the estimated values of these coefficients. Recall that our model also includes fixed effects $\beta_{f}^{\mathbf{i_1}}$, $\beta_{f}^{\mathbf{i_2}}$, and $\beta_{f}^{\mathbf{i_3}}$ corresponding to these variables. For additional context, Figure~\ref{fig:heatmap_random_effect_4} reports the values of these coefficients summed with their corresponding fixed effects for shoe 202-L in the WVU database. That is, it reports $x_{s',a}^{\mathbf{i}} (\beta_{\text{sv}}^{\mathbf{i}}+ \beta_{f}^{\mathbf{i}})$, for $\mathbf{i}=\mathbf{i_2},\mathbf{i_3}, \text{ and } \mathbf{i_1}$). 

\begin{figure}[ht!]
        \centering
\includegraphics[scale=3.0,width=1\linewidth]{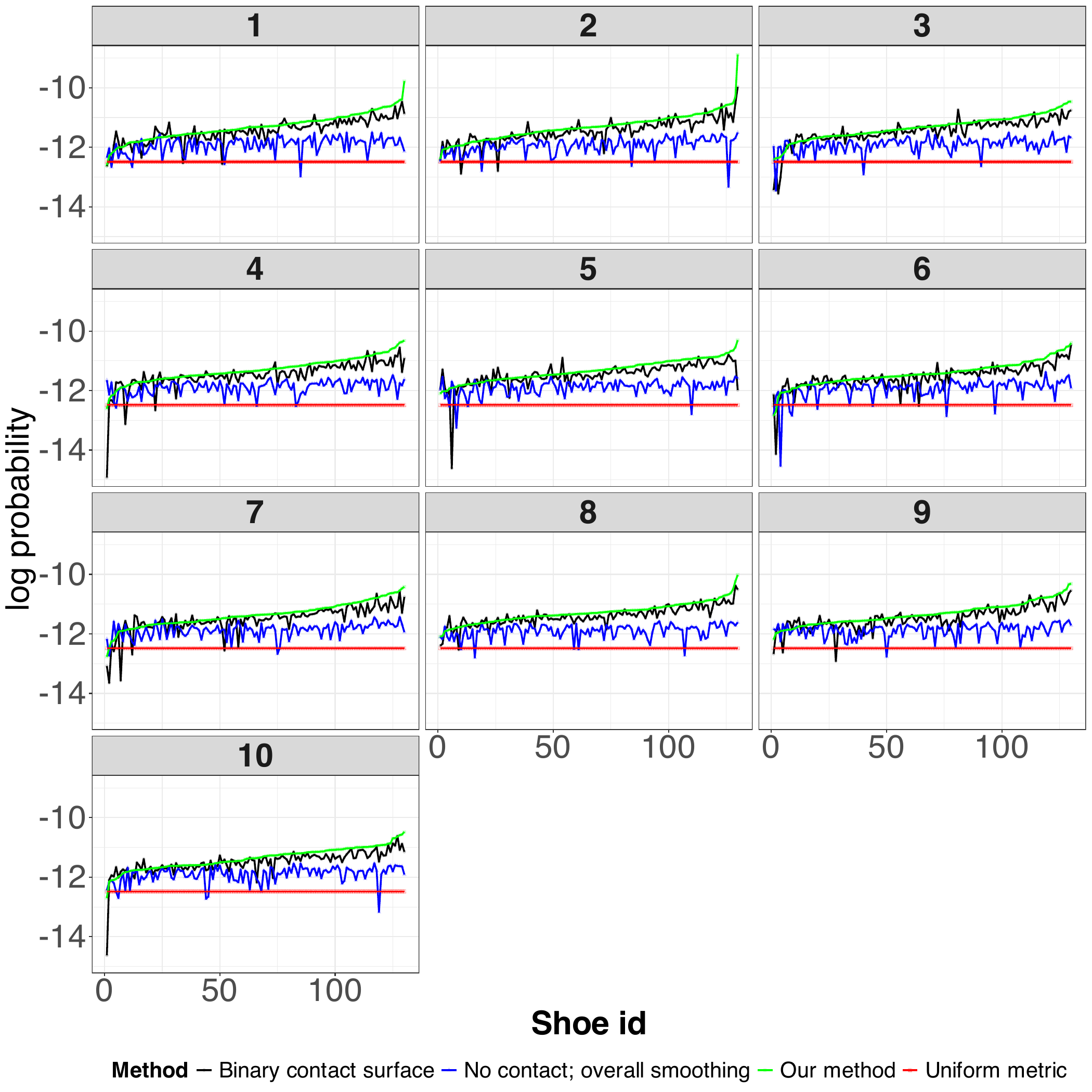} % Replace with your image or plot
        \caption{Comparison of methods for all 10 folds.}
%\footnotesize {Note: In the horizontal axis, the shoe id is arranged based on the performance of our method, which consists the spatial varying coefficients method with image gradient. Uniform metric (\cite{stone2006footwear}) and No contact; overall smoothing (\cite{yekutieli2012expert}) are two methods that do not incorporate any spatial covariates. Based on the binary contact surface with 32 covariates (\cite{spencer2020bayesian}) and spatial smoothing the performance was significantly better than methods that do not incorporate any spatial covariate. Finally, our proposed method performed best among all existing methods that incorporate spatial varying coefficients ($2^6$ fixed effects, overall smoothing, and only 3 random effects i.e. pixel intensity, image gradient, and their interaction) with continuous surfaces. All the methods are arranged based on the performance across our method.}
\label{fig:multinomial_comparizon_plot_10_folds}
\end{figure}

\begin{figure}[ht!]
        \centering
\includegraphics[scale=3.0,width=1\linewidth]{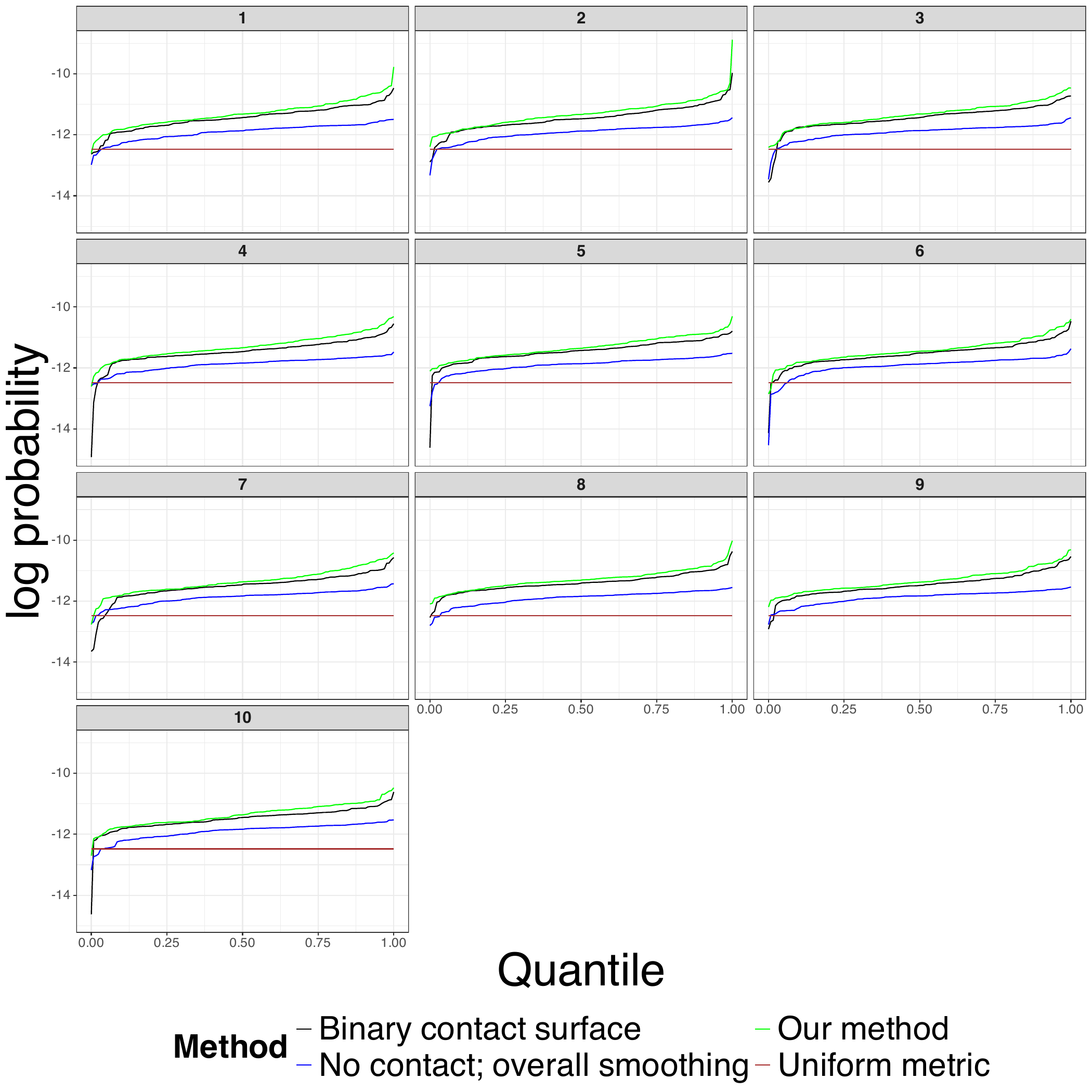} % Replace with your image or plot
        \caption{Quantile Robustness check for all ten folds.}
        %\footnotesize {Note: In this plot, the robustness of our method is checked. Although two other existing methods (overall smoothing and binary contact surface, blue and black colored) are crossing below the uniform metric (red), our method (green) turns out to be above the other three methods even for lower quantiles across all the folds.} }
        \label{fig:robustness_check}
    \end{figure}

Figure 10 in the main text provides a comparison of four competitor models for the third held-out fold in the cross-validation strategy. For completeness, Figures~\ref{fig:multinomial_comparizon_plot_10_folds} and \ref{fig:robustness_check} below report the corresponding results for all 10 folds in a large faceted plot.

\end{document}